\documentclass[journal]{IEEEtran}
\usepackage{natbib}

\usepackage{moreverb,url}

\usepackage[colorlinks,bookmarksopen,bookmarksnumbered,citecolor=red,urlcolor=red]{hyperref}

\usepackage{amsmath,amssymb,amsfonts}
\usepackage{textcomp}
\usepackage{xcolor}

\newcommand\BibTeX{{\rmfamily B\kern-.05em \textsc{i\kern-.025em b}\kern-.08em
T\kern-.1667em\lower.7ex\hbox{E}\kern-.125emX}}

\usepackage{subcaption}
\usepackage{xcolor}
\usepackage{ifthen}
\usepackage{makecell}
\usepackage{array}
\usepackage{MnSymbol}
\usepackage{mh-macros}
\usepackage{graphicx}
\usepackage{tikz}

\usepackage{color,soul}

\newif\ifshowimages
\showimagestrue 

\newboolean{revVer}
\setboolean{revVer}{false} 
\newboolean{revVer1}
\setboolean{revVer1}{false} 
\newboolean{revVer2}
\setboolean{revVer2}{true} 
\newboolean{empText}
\setboolean{empText}{true}

\ifthenelse{\boolean{revVer}}
{
\newcommand{\hlRev}[1] {\textcolor{black}{#1}}
}
{
\newcommand{\hlRev}[1] {{#1}}
}

\ifthenelse{\boolean{revVer}}{\newcommand{\bardh}[1]{{\color{purple} #1}}}
{\newcommand{\bardh}[1]{}}

\ifthenelse{\boolean{empText}}
{

}
{

}
\ifthenelse{\boolean{revVer1}}
{
\newcommand{\revision}[1] {\textcolor{blue}{#1}}
}
{
\newcommand{\revision}[1] {{#1}}
}

\ifthenelse{\boolean{revVer2}}
{
\newcommand{\revisionTwo}[1] {\textcolor{black}{#1}}
}
{
\newcommand{\revisionTwo}[1] {{#1}}
}

\usepackage[ruled]{algorithm}
\usepackage{algpseudocode}
\usepackage{mathtools}
\usepackage{color}

\usepackage{tikz}
\usetikzlibrary{calc,matrix,patterns,shadings,backgrounds,fit}
\usepackage{ifthen}
\definecolor{NewText}{RGB}{100, 128, 64}
\definecolor{NewText2}{RGB}{100, 128, 64}
\newcolumntype{?}[1]{!{\vrule width #1}}
\usepackage[colorinlistoftodos]{todonotes}

\begin{document}

\title{Formalizing and Evaluating Requirements of Perception Systems for Automated Vehicles using Spatio-Temporal Perception Logic
}

\author{Mohammad Hekmatnejad, Bardh Hoxha, Jyotirmoy V. Deshmukh, Yezhou Yang, and Georgios Fainekos
\thanks{Mohammad Hekmatnejad was with the School of Computing and Augmented Intelligence (formerly CIDSE), Arizona State University, USA; e-mail: {mhekmatn@asu.edu}.}
\thanks{Bardh Hoxha is with the Toyota Research Institute of North America, USA.}
\thanks{Jyotirmoy V. Deshmukh is with the University of Southern California, USA.}
\thanks{Yezhou Yang is with Arizona State University, USA.}
\thanks{Georgios Fainekos is with the Toyota Research Institute of North America, USA; the work was performed while he was with the Arizona State University, USA.}
}

\maketitle

\begin{abstract}
Automated vehicles (AV) heavily depend on robust perception systems.
Current methods for evaluating vision systems focus mainly on frame-by-frame performance.
Such evaluation methods appear to be inadequate in assessing the performance of a perception subsystem when used within an AV.
In this paper, we present a logic -- referred to as  Spatio-Temporal Perception Logic (STPL) -- which utilizes both spatial and temporal modalities.
STPL enables reasoning over perception data using spatial and temporal operators.
One major advantage of STPL is that it facilitates basic sanity checks on the functional performance of the perception system, even without ground-truth data in some cases.
We identify a fragment of STPL which is efficiently monitorable offline in polynomial time. 
Finally, we present a range of specifications for AV perception systems to highlight the types of requirements that can be expressed and analyzed through offline monitoring with STPL.
\end{abstract}


\section{Introduction}
The safe operation of automated vehicles (AV), advanced driver assist systems (ADAS), and mobile robots in general, fundamentally depends on the correctness and robustness of the perception system (see discussion in Sec. 3 by \cite{schwarting2018planning}).
Faulty perception systems and, consequently, inaccurate situational awareness can lead to dangerous situations
(\cite{Lee2018arstechnica,Templeton2020forbes-tesla}).
For example, in the accident in Tempe in 2018 involving an Uber vehicle (\cite{Lee2018arstechnica}), different sensors had detected the pedestrian, but the perception system was not robust enough to assign a single consistent object class to the pedestrian.
Hence, the AV was not able to predict the future path of the pedestrian.
This accident highlights the importance of identifying the right questions to ask when evaluating perception systems.

\ifshowimages
\begin{figure}
  \vspace{0.7em}
\centering
\begin{subfigure}{.95\linewidth}
  \captionsetup{justification=centering}
  \centering
  \includegraphics[width=1\linewidth]{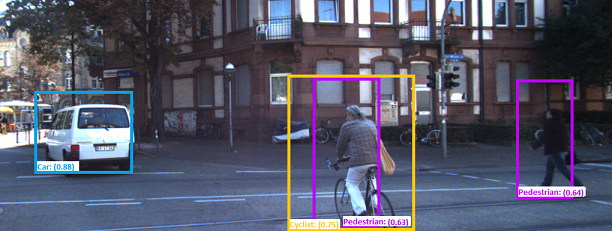}
  \caption{\scriptsize{Frame $f=0$ and $\tau(0)=0$}}\label{fig:kitti-1}
\end{subfigure} 
\begin{subfigure}{.95\linewidth}
  \captionsetup{justification=centering}
  \centering
  \includegraphics[width=1\linewidth]{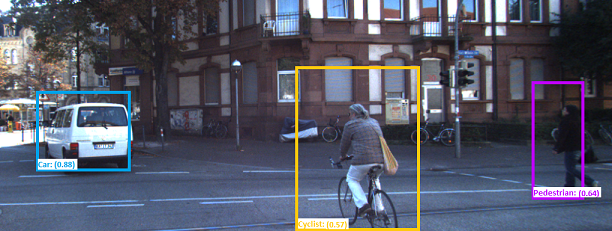}
  \caption{\scriptsize{Frame $f=1$ and $\tau(1)=0.04$}}\label{fig:kitti-2}
\end{subfigure} 
\begin{subfigure}{.95\linewidth}
  \captionsetup{justification=centering}
  \centering
  \includegraphics[width=1\linewidth]{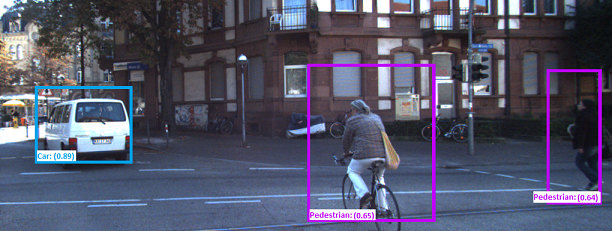}
  \caption{\scriptsize{Frame $f=2$ and $\tau(2)=0.08$}}\label{fig:kitti-3}
\end{subfigure} 
\begin{subfigure}{.95\linewidth}
  \captionsetup{justification=centering}
  \centering
  \includegraphics[width=1\linewidth]{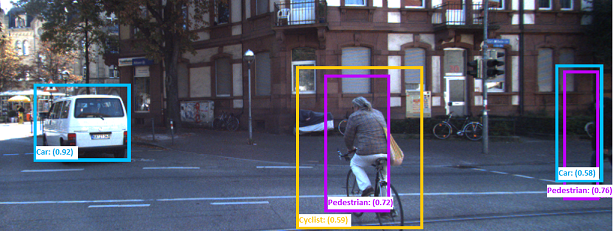}
  \caption{\scriptsize{Frame $f=3$ and $\tau(3)=0.12$}}\label{fig:kitti-4}
\end{subfigure} 
\begin{subfigure}{0.95\linewidth}
  \captionsetup{justification=centering}
  \centering
  \includegraphics[width=1\linewidth]{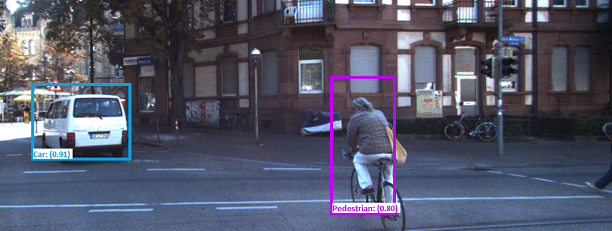}
  \caption{\scriptsize{Frame $f=4$ and $\tau(4)=0.16$}}\label{fig:kitti-5}
\end{subfigure} 
\begin{subfigure}{0.95\linewidth}
  \captionsetup{justification=centering}
  \centering
  \includegraphics[width=1\linewidth]{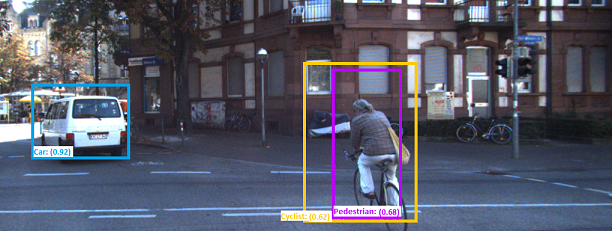}
  \caption{\scriptsize{Frame $f=5$ and $\tau(5)=0.2$}}\label{fig:kitti-6}
\end{subfigure} 
\caption{A series of image frames taken from KITTI (\cite{geiger2013vision}) augmented with the data about classified data-objects using SqueezeDet (\cite{wu2017squeezedet}). Here, $\tau$ is a function that maps each frame $f$ to its captured time. }
\label{fig:kitti:case-study}
\end{figure}
\fi

In this paper, we argue that in order to improve AV/ADAS safety, we need to be able to formally express what should be the assumptions, performance, and guarantees provided by a perception system.
For instance, the aforementioned accident highlights the need for predicting the right object class quickly and robustly.
An informal requirement (in natural language) expressing such a perception system performance expectation could be:
\begin{requirement} \label{lbl:req-intro-exm-1}
{\it Whenever a new object is detected, then it is assigned a class within 1 sec, after which the object does not change class until it disappears.}
\label{req:uber} 
\end{requirement}

Such requirements are not only needed for deployed safety critical systems, but also for existing training and evaluation data sets.
For example, a sequence of six image frames from the KITTI dataset is shown in Figure \ref{fig:kitti:case-study}.
All detected objects with bounding boxes are labeled with class names such as \textit{Car, Pedestrian}, and \textit{Cyclist} in those frames.
Nonetheless, in Fig. \ref{fig:kitti:case-study}(a), a \textit{cyclist} that was detected in frame $f=0$ and existed in frame $f=1$ changed its class to \textit{pedestrian} in frame $f=2$, which is a violation of Req. \ref{lbl:req-intro-exm-1}. 
If such a requirement is too strict for the perception system, then it could be relaxed by requiring that the object is properly classified in at least 4 out of 5 frames (or any other desired frequency).
The first challenge for an automated testing \& verification framework is to be able to formally (i.e., mathematically) represent such high-level requirements.

Formalizing the expectations on the performance of the perception system has several benefits.
First and foremost, such formal requirements can help us evaluate AV/ADAS perception systems beyond what existing metrics used in vision and object tracking can achieve (\cite{YurtseverLCT2020access,Richter2017iccv}). 
In particular, evaluation metrics used for image classification algorithms are typically operating on a frame-by-frame basis ignoring cross-frame relations and dependencies.
Although tracking algorithms can detect misclassifications, they cannot assess the frequency and severity of them,  or more complex spatio-temporal failure conditions.

Second, formal specifications enable automated search-based testing (e.g., see \cite{AbbasKRM2017asme,TuncaliEtAl2020tiv,dreossi2019verifai,DreossiDS2019jar,corso2020survey,GladischEtAl2019ase,decastrointerpretable}) and monitoring (e.g., see \cite{RizaldiEtAl2017ifm,HekmatnejadEtAl2019memocode}) for AV/ADAS. 
\revision{Requirements are necessary in testing in order to express the conditions (assumptions) under which system level violations are meaningful.
Similarly, logical requirements are necessary in monitoring since without formal requirements, it is very hard to concisely capture the rules of the road. 
We envision that a formal specification language which specifically targets perception systems will enable similar innovations in testing and monitoring AV/ADAS.}

Third, formal specifications on the perception system can also function as a requirements language between original equipment manufacturers (OEM) and suppliers.
As a simple example for the need of a requirements language consider the most basic question: should the image classifier correctly classify all objects within the sensing range of the lidar, or all objects of at least $x$ pixels in diameter?
Again, {\it without requirements, any system behavior is acceptable and, most importantly, not testable/verifiable!}

\revision{
Finally, using a formal specification language, we can search offline perception datasets (see \cite{KimEtAl2022iccps,AndersonEtAl2023rv,Yadav2019}) to find scenarios that violate the requirements in order to reproduce them in simulation, e.g., \cite{BashettyBAF2020icra}, or to assess the real-to-sim gap in testing, e.g., \cite{FremontEtAl2020itsc}.
}

\revision{
Even though there is a large body of literature on the application of formal languages and methods to AV/ADAS (see the survey by \cite{MehdipourATB2023automatica}), the prior works have certain limitations when perception systems are specifically targeted.
}
Works like \cite{TuncaliEtAl2020tiv} and \cite{DreossiDS2019jar}
demonstrated the importance of testing system level requirements for an AV when the AV contains machine learning enabled perception components.
Both works use Signal Temporal Logic (STL) (\cite{MalerN04formats}) for expressing requirements on the performance of the AV/ADAS.
However, STL cannot directly express requirements on the performance of the perception system itself.

This limitation was identified by \cite{dokhanchi2018evaluating} who developed a new logic -- Timed Quality Temporal Logic (TQTL) -- which can directly express requirements for image classification algorithms like SqueezeDet by \cite{wu2017squeezedet}. 
Namely, TQTL was designed to enable reasoning over objects classified in video frames and their attributes.
For example, TQTL can formalize Req. \ref{req:uber} that an object should not change classes.
However, TQTL does not support spatial or topological reasoning over the objects 
\revision{
or regions
}
in a scene.
This is a major limitation since TQTL cannot express requirements such as occlusion, overlap, and other spatial relations for basic sanity checks, e.g., that objects do not ``teleport" from frame to frame, 
\revision{
or that ``every 3D bounding box contains at least 1 lidar point".
}

\revision{
In this paper, we introduce the Spatio-Temporal Perception Logic (STPL) to address the aforementioned limitations.
We combine TQTL with the spatial logic ${\mathcal S}4_u$%
\footnote{For a historical introduction to ${\mathcal S}4_u$ see \cite{BenthemB2007_ch5_handbook} and \cite{kontchakov2007spatial}.}
to produce a more expressive logic specifically focused on perception systems.
STPL supports quantification among objects or regions in an image, and time variables for referring to specific points in time.
It also enables both 2D and 3D spatial reasoning by expressing relations between objects across time.
For example with STPL, we can express requirements on the rate that bounding boxes should overlap: 
}

\begin{requirement}\label{lbl:req-intro-exm-2} 
{\it The frames per second of the camera is high enough so that for all detected cars, their bounding boxes self-overlap for at least 3 frames and for at least 10\% of the area.}\label{req:realtime} 
\end{requirement}
\noindent
which is satisfiable for the image frames in Fig. \ref{fig:kitti:case-study}.

\revision{
Beyond expressing requirements on the functional performance of perception systems,
STPL can be used to compare different machine learning algorithms on the same perception data.
}
For example, STPL could assess how rain removal algorithms (see the work by  \cite{sun2019convolutional}) improve the performance of the object detection algorithms over video streams. 
\revision{
Along these lines, \cite{MallickGBD2023rv} proposed to use TQTL for differential testing between two different models for pedestrian detection in extreme weather conditions.
}
As another example, STPL could evaluate the impact of a point cloud clustering algorithm on a prediction algorithm (see the works by \cite{campbell2016traffic} and \cite{qin2016spatial}).

\revision{
Since STPL is a very expressive formal language (TQTL+${\mathcal S}4_u$), it is not a surprise that both the offline and online monitoring problems can be computationally expensive.
The main challenge in designing a requirements language for perception systems is to find the right balance between expressivity and computability.
}

\revision{
STPL borrows time variables and freeze time quantifiers from the Timed Propositional Temporal Logic (TPTL) by \cite{AlurH94acm}.
}
Unfortunately, the time complexity of monitoring TPTL requirements is PSPACE-hard (see \cite{MarkeyR2006tcs}).
\revision{
Even though the syntax and semantics of STPL that we introduce in Section \ref{sec:STPL} allow arbitrary use of time variables in the formulas, in practice, in our implementation, we focus on TPTL fragments which are  computable in polynomial time (see \cite{dokhanchi2016efficient,ElgyuttFH2018formats,GhorbelP2022hscc}).
}

\revision{
Another source of computational complexity in STPL is the use of object/region quantification and set operations, i.e., intersection, complementation, and union.
Even our limited use of quantifiers ($\exists$, $\forall$) over finite sets (i.e., objects in a frame) introduces a combinatorial component to our monitoring algorithms.
This is a well known problem in first-order temporal logics that deal with data (e.g., see \cite{BasinKMZ2015acm} and \cite{HavelundPU2020fmsd}).
In addition, the computational complexity of checking whether sets, which correspond to objects/regions, satisfy an arbitrary ${\mathcal S}4_u$ formula depends on the set representation and the dimensionality of the space (e.g., polytopes \cite{Baotic2009automatika}, orthogonal polyhedra \cite{bournez1999orthogonal}, quadtrees or octrees \cite{aluru2018quadtrees}, etc).
}

\revision{
In this paper, we introduce an offline monitoring algorithm for STPL in Section \ref{lbl:apx:algorithm}, and We study its computational complexity in Section \ref{sec:complexity}.
In Section \ref{sec:poly-frag}, we identify a fragment which can remain efficiently computable.
Irrespective of the computational complexity of the full logic, in practice, our experimental results in Section \ref{sec:exp} indicate that we can monitor many complex  specifications within practically relevant times. 
}

\revision{
All in all, to the best of our knowledge, this is the first time that a formal language is specifically designed to address the problem of functional correctness of perception systems within the context of autonomous/automated mobile systems. 
Our goal is to design a logic that can express functional correctness requirements on perception systems while understanding what is the computational complexity of monitoring such requirements in practice.
As a bi-product, we develop a logical framework that can be used to compare different perception algorithms beyond the standard metrics.
}

\revision{
In summary, the {\bf contributions} of this paper are:
\begin{enumerate}
    \item We introduce a new logic -- Spatio-Temporal Perception Logic (STPL) -- that combines quantification and functions over perception data, reasoning over timing of events, and set based operations for spatial reasoning (Section \ref{sec:STPL}).
    \item We present examples of STPL specifications of increasing complexity as a tutorial (Section \ref{sec:case-study}), and we demonstrate that STPL can find label inconsistencies in publicly available AV datasets (see Figure \ref{fig:kitti-occ-example}).
    \item We introduce an offline monitoring algorithm (Section \ref{lbl:apx:algorithm}), and we propose to use different set representations for efficient computations (Section \ref{sec:complexity}).
    \item We study the time complexity of the offline monitoring problem for STPL formulas, and we propose fragments for which the monitoring problem is computable in polynomial time (Section \ref{sec:complexity}).
    \item We experimentally study offline monitoring runtimes for a range of specifications and present the trade offs in using the different STPL operators (Section \ref{sec:exp}).
    \item We have released a publicly available open-source toolbox (\cite{stpl_tools}) for offline monitoring of STPL specifications over perception datasets.
    The toolbox allows for easy integration of new user-defined functions over time and object variables, and new data structures for set representation.
\end{enumerate}
}

\revisionTwo{
One common challenge with any formal language is its accessibility to users with limited experience in formal requirements.
In order to remove such barriers, \cite{AndersonHF2022re} developed PyFoReL which is a domain specific language (DSL) for STPL inspired by Python and integrated with Visual Studio Code.
In addition, our publicly available toolbox contains all the examples and case studies presented in this paper which can be used as a tutorial by users without prior experience in formal languages.
One can further envision integration with Large Language Models for direct use of natural language for requirement elicitation similar to the work by \cite{PanCB2023icra,FuggittiC2023aaai,CoslerEtAl2023cav} for signal or linear temporal logic. 
}

\revision{
Finally, even though the examples in the paper focus on perception datasets from AV/ADAS applications (i.e., \cite{nuscenes2020website,Geiger2013IJRR}), STPL can be useful in other applications as well. 
}
One such example could be in robot manipulation problems where mission requirements are already described in temporal logic (e.g., see \cite{he2018automated} and \cite{he2015towards}).

\section{Preliminaries}

We start by introducing definitions that are used throughout the paper for representing classified data generated by perception subsystems.
In the following, an ego car (or the Vehicle Under Test -- VUT) is a vehicle equipped with perception systems and other automated decision making software components, but not necessarily a fully automated vehicle (e.g., SAE automation level 4 or 5).

\subsection{Data-object Stream}\label{def:data-obj-stream}
A \textit{data-object stream} $\Data$ is a sequence of data-objects representing objects in the ego vehicle's environment.
\revision{Such data objects could be the output of the perception system, or ground truth data, or even a combination of these.
The ground truth can be useful to analyze the precision of the output of a perception system.
That is, given both the output of the perception system and the ground truth, it is possible to write requirements on how much, how frequently, and under what conditions the perception system is allowed to deviate from the ground truth.
In all the examples and case studies that follow, we assume that we are either given the output of the perception system, or the ground truth -- but not both.
In fact, we treat ground truth data as the output of a perception system in order to demonstrate our framework.
Interestingly, in this way, we also discover inconsistencies in the labels of training data sets for AV/ADAS (see Fig. \ref{fig:kitti-occ-example}).
}

We refer to each dataset in the sequence as a frame.
In each frame, objects are classified and stored in a data-object. 
Data-objects are abstract because in the absence of a standard classification format, the output of different perception systems are not necessarily compatible with each other.
To simplify the presentation, we will also refer to the order of each frame in the data stream as a frame number.
For each frame $i$, $\Data(i)$ is the set of data-objects for frame $i$, where $i$ is the order of the frame in the data-object stream $\Data$.
We assume that a data stream is provided by a perception system, in which a function $\Retrieve$ retrieves data-attributes for an identified object.
A \textit{data-attribute} is a property of an object such as class, probability, geometric characteristics, etc.
Some examples could be:
\begin{itemize}
    \item {by $\Retrieve(\Data (i), id).Class$, we refer to the class of an object identified by $id$ in the $i$'th frame,}
    \item by $\Retrieve(\Data(i),id).Prob$, we refer to the probability that an object identified by $id$ in the $i$'th frame is correctly classified, 
    \item by $\Retrieve(\Data(i),id).PC$, we refer to the point clouds associated with an object identified by $id$ in the $i$'th frame.
\end{itemize}
The function $\sorder$ returns the set of identifiers for a given set of data-objects. 
By $\sorder(\Data(i))$, we refer to all the identifiers that are assigned to the data-objects in the $i$'th frame. 
Another important attribute of each frame is its time stamp, i.e., when this frame was captured in physical time.
We represent this attribute for the frame $i$ by $\tau(i)$.
\begin{exmp}
Assume that data-object stream $\Data$ is represented in Figure \ref{fig:kitti:case-study}, then the below equalities hold:
\begin{itemize}
    \item $\sorder(\Data(1))=\{1,2,3\}$
    \item \revisionTwo{ $\Retrieve(\Data(1),2).Class=cyclist$ }
    \item \revisionTwo{ $\Retrieve(\Data(1),2).Prob=0.57$ }
    \item $\tau(1)=0.04$
\end{itemize}
\end{exmp}

\revision{
Finally, we remark that in offline monitoring, which is the application that we consider in this paper, the data stream $\Data$ is finite. 
We use the notation $|\Data|$ to indicate the total number of frames in the data stream.
}

\subsection{Topological Spaces}

Throughout the paper, we will be using $\nat$ to refer to the set of natural numbers and $\reals$ to real numbers.
In the following, for completeness, we present some definitions and notation related to topological spaces and the $S4_u$ logic which are adopted from \cite{gabelaia2005combining} and \cite{kontchakov2007spatial}. 
Later, we build our $MTL \times S4_u$ definition on this notation.

\begin{defn}[Topological Space]\label{lbl:topo-space}
    A topological space is a pair $\Ts=\langle \Suniverse,\Its \rangle$ in which $\Suniverse$ is the universe of the space, and $\Its$ is the interior operator on $\Suniverse$.
    The universe is a nonempty set, and $\Its$ satisfies the standard Kuratowski axioms \revision{where $X,Y \subseteq \Suniverse$}:
    \begin{align*}
        \Its (X \cap Y) = \Its X \cap \Its Y, 
        \;\;
        \Its X \revision{ = } \;\Its\Its X,
        \;\;
        \Its X \subseteq X,
        \;\;
        \mbox{and} 
        \;\;
        \Its \Suniverse = \Suniverse.
    \end{align*}
\end{defn}
We denote $\Css$ (closure) as the dual operator to $\Its$, such that 
$\Css X = \Suniverse - \Its (\Suniverse - X)$ for every $X \subseteq \Suniverse$.
Below we list some remarks related to the above definitions:
\begin{itemize}
    \item $\Its X$ is the \textit{interior} of a set $X$,
    \item $\Css X$ is the \textit{closure} of a set $X$,
    \item $X$ is called \textit{open} if $X = \Its X$,
    \item $X$ is called \textit{closed} if $X = \Css X$,
    \item If $X$ is an open set then its complement $\overline{X} = \Suniverse - X$ is a closed set and vice versa,
    \item For any set $X \subseteq \Suniverse$, its \textit{boundary} is defined as $\Css X - \Its X$ ($X$ and its complement have the same boundary),
\end{itemize}

\subsection{Image Space Notation and Definitions}\label{sec:image-space}
\begin{figure}
    \centering
    \begin{subfigure}{1\linewidth}
    \includegraphics[width=1\linewidth]{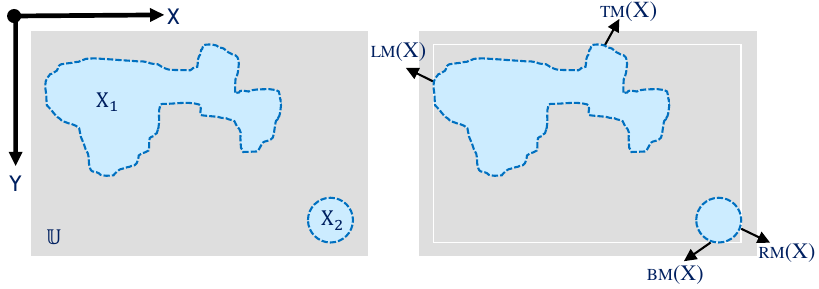}
    \end{subfigure}
    \caption{
    Let $X = X_1 \cup X_2$ with $\Css X = X$.
    Left: the whole rectangle represents the universe (colored gray) $\Suniverse$, the regions $X_1$ and $X_2$ without the dashed-lines (colored light blue) represent the interior $\Its X$ of the set $X$, the dashed lines (colored darker blue) represent the boundary $\Css X - \Its X$ of the set $X$, and the remaining region represents $\Suniverse - X$.
    Right: \revision{the tightest bounding box around set $X$ that includes the four unique elements of the set as in Def. \ref{def:spatial-order}.}
    }
    \label{fig:exm:geo-shape}
\end{figure}

Even though our work is general and applies to arbitrary finite dimensional spaces, our main focus is on 2D and 3D spaces as encountered in perception systems in robotics.
In the following, the discussion focuses on 2D image spaces with the usual pixel coordinate systems.
In Figure \ref{fig:exm:geo-shape} left, a closed set is illustrated as the light blue region, while its boundary is a dashed-line in darker blue.

\begin{defn}[Totally-ordered Topological Space]\label{lbl:po-topo-space}
A Totally-ordered (TO) topological space $\revision{\TOTS}$ is a topological space $\langle \Suniverse,\Its \rangle$ that is equipped with a total ordering relation $\preceq \;\subseteq \revisionTwo{\Suniverse} \times \revisionTwo{\Suniverse}$.
That is, $\forall p_1,p_2 \in \revisionTwo{\Suniverse}$ either $p_1 \preceq p_2$ or $p_2 \preceq p_1$.
\end{defn}

\begin{defn}[Total order in 2D spaces]\label{lbl:def-TO-2D}
In a two-dimensional (2D) TO topological space $\revision{\TOTS}$, $p=(x,y) \in \revisionTwo{\Suniverse}$ denotes its coordinates in the $x-y$ Cartesian coordinate system, where $x,y \in \reals_{\ge 0}$.
For $p_1=(x_1,y_1)$, $p_2=(x_2,y_2)$ the ordering relation is defined as:
\begin{gather*}
(x_1,y_1) \preceq_{\TOIm} (x_2,y_2) \iff 
y_1 < y_2 \vee ( y_1 = y_2 \wedge x_1 \le x_2).
\end{gather*}

\end{defn}

Note that even though Def. \ref{lbl:def-TO-2D} considers the coordinates over the reals, i.e., $\reals^2_{\geq 0}$, in practice, the image space is defined over the integers, i.e., $\nat^2$ (pixel coordinates).
The topological space in Figure \ref{fig:exm:geo-shape} consists of the universe which is all the pixels that belong to the whole 2D gray-rectangle and the closure operator that for any set of regions includes their edges. 
Assume that the left-upper corner of the image is the origin of the $x-y$ Cartesian coordinate system, and the $x-axis$ and the $y-axis$ are along the width and height of the image, respectively. 
This is the standard coordinate system in image spaces.
Then, any pixel that belongs to the universe has an order with respect to the other pixels. 
For example, 
consider $X = X_1 \cup X_2$ in Fig. \ref{fig:exm:geo-shape}, 
then all the pixels that belong to $X_2$ have higher orders than those in $X_1$.

\begin{defn}[Spatial Ordering Functions]\label{def:spatial-order}
Given a closed set \revisionTwo{$\subTOTS \subseteq \Suniverse$} from a 2D TO topological space $\TOTS = \langle \Suniverse,\Its \rangle$,
we define the top-most, bottom-most, left-most, 
\revisionTwo{
right-most, and center-point functions
} 
by $\TM,\BM,\LM,\RM.\CT : \revision{2^\Suniverse \rightarrow \Suniverse}$, respectively such that
\begin{align*}
\revisionTwo{
    \TM(\subTOTS)  =
} & 
\; \revisionTwo{
(x_p,y_p) \in \subTOTS \mbox{ s.t. } 
}  
\\
&
\revisionTwo{
    \forall s=(x_s,y_s)\in \Suniverse,
    y_p < y_s \vee (y_p = y_s \wedge x_p \ge x_s)
}
	\displaybreak[2] 
    \\
	\revisionTwo{
	    \BM(\subTOTS) = 
	}
    &
    \revisionTwo{
    \; (x_p,y_p) \in \subTOTS \mbox{ s.t. }   
	}
	\\
    &
    \revisionTwo{
    \forall s=(x_s,y_s)\in \Suniverse,  
    y_p > y_s \vee (y_p = y_s \wedge x_p \le x_s)
	}
	\displaybreak[2] 
    \\
    \revisionTwo{
 	\LM(\subTOTS) = 
	 }
	& 
   \revisionTwo{
	\; (x_p,y_p)  \in \subTOTS \mbox{ s.t. }   
	} \\
    & 
    \revisionTwo{
    \forall s=(x_s,y_s)\in \Suniverse,  
    x_p < x_s \vee (x_p = x_s \wedge y_p \le y_s)
	}
	\displaybreak[2] 
    \\    
   \revisionTwo{\RM(\subTOTS) =}
    & 
   \revisionTwo{\; (x_p,y_p)  \in \subTOTS \mbox{ s.t. } } \\
   & 
   \revisionTwo{
   	\forall s=(x_s,y_s)\in \Suniverse,  
   x_p > x_s \vee (x_p = x_s \wedge y_p \ge y_s)} \\
\end{align*}
\revisionTwo{$and, \CT(\subTOTS)$ is the center point of the rectangle defined by $\LM(\subTOTS)$, $\RM(\subTOTS)$, $\BM(\subTOTS)$, and $\TM(\subTOTS)$ points.}

\end{defn}

In Def. \ref{def:spatial-order}, the notation $2^{A}$ for a set $A$ denotes the set of all subsets of $A$, i.e., the powerset of $A$.
In Figure \ref{fig:exm:geo-shape}, we have identified the left-most, right-most, bottom-most, and right-most elements of the set $X$ using the $\LM(X)$, $\TM(X)$, $\RM(X),$ and $\BM(X)$ function definitions.
The right rectangle in Figure \ref{fig:exm:geo-shape} is the tightest bounding box for the set $X$, and it can be derived using only those four points.


\subsection{Spatio-Temporal Logic MTL$\times S4_u$}

Next, we introduce the logic MTL$\times S4_u$ 
\revision{
over discrete-time semantics.
MTL$\times S4_u$
}
is a combination of Metric Temporal Logic (MTL) (\cite{Koymans90}) with the  $S4_u$ logic of topological spaces.
From another perspective, it is an extension to the logic PTL$\times S4_u$ (see \cite{gabelaia2005combining}) by adding time/frame intervals into the spatio-temporal operators.
Even though MTL$\times S4_u$ is a new logic introduced in this paper, we present it in the preliminaries section  
\revision{
in order to gradually introduce notation and concepts needed for STPL.
In this paper, we use the Backus-Naur form (BNF) (see  \cite{perugini2021programming}) which is standard for providing the grammar when we define the syntax of formal and programming languages.
} 

\revision{
In the following, we provide the formal definitions for MTL$\times S4_u$.
Henceforth, the symbol $p$ refers to a \textit{spatial proposition}, i.e., it represents a set in the topological space.
In addition, the symbol $\Stau$ represents spatial expressions that evaluate to subsets of the topological space.
We refer to $\Stau$ as a \textit{spatial term}.
}

\begin{defn}[Syntax of MTL$\times S4_u$ as a temporal logic of topological spaces]
\label{lbl:mtl_s4u_syntax-ext}
    Let $\Pi$ be a finite set of spatial propositions over $\Ts=$ $\langle\Suniverse,\Its \rangle$,
    then a formula $\varphi$ of $MTL \times S4_u$ can be defined as follows:
    \begin{align*}
        \Stau &::= p \;|\; \overline{\Stau} \;|\; \Stau \sqcap \Stau \;|\; \Itl\;{\Stau} \;|\; \Stau \;\Un^{s}_{\Ic}\; \Stau 
        \;|\;  \Next^{s}_{\Ic}\; \Stau  
        \\
        \varphi &::= \Sexists\; \Stau \;|\; \neg \varphi \;|\; \varphi \wedge \varphi \;|\; \varphi \;\Un_{\Ic} \; \varphi \;|\;  \Next_{\Ic} \; \varphi
    \end{align*}
    where $p \in \Pi$ is a spatial proposition.
\end{defn}

In the above definition, the grammar for MTL$\times S4_u$ consists of two sets of production rules $\Stau$ and $\varphi$.
\revisionTwo{
The spatial production rule $\Stau$ in Def. \ref{lbl:mtl_s4u_syntax-ext} contains a mix of spatial ($\bar{\;}, \sqcap, \Itl$) and spatio-temporal ($\Un^{s}_{\Ic}, \Next^{s}_{\Ic}$)  operators.
}
Here,
$\overline{\Stau}$ is the spatial complement of the spatial term $\Stau$,
$\sqcap$ is the spatial intersection operator, and 
$\Itl$ is the spatial interior operator. 
\revision{
The $\Un^{s}_{\Ic}$ and $\Next^{s}_{\Ic}$ operators are the spatio-temporal until and the spatio-temporal next, respectively.
}
Here, the subscript $\Ic$ denotes a non-empty interval of $\reals_{\geq 0}$ and captures any timing constraints for the operators.
\revisionTwo{
When timing constraints are not needed, then we can set $\Ic=[0,+\infty)$, or remove the subscript $\Ic$ from the notation.
Intuitively, the spatio-temporal until and next operators introduce spatial operations over time.
For instance, the expression $p_1 \Uc^{s}_{[1,3]} p_2$ computes the union over all sets resulting by the intersection of each occurrence of set $p_2$ at some time in the interval $[1,3]$ with all the sets $p_1$ up to that time (see Def. \ref{lbl:ext-temporal-topo-model} for more details). 
}
We refer to the spatio-temporal operators 
$\Uc^{s}_{\Ic}$ and 
$\Next^{s}_{\Ic}$, 
as \textit{Spatio-Temporal Evolving} (STE) operators.
Also, we call a formula STE if it has STE operators in it.
Similarly, we call a spatio-temporal formula \textit{Spatial Purely Evolving} (SPE) if there is no STE operator in the formula. 
For more information refer to the (OC) and (PC) definitions by \cite{gabelaia2005combining}.

\revision{
In the production rule $\varphi$ in Def. \ref{lbl:mtl_s4u_syntax-ext}, the \textit{spatially exists} $\Sexists$ checks if the spatial term following it evaluates to a non-empty set.
That is, $\Sexists \;\Stau$ checks if there exist some points in the set represented by $\Stau$.
In $\varphi$, except for the \textit{spatially exists} $\Sexists$, the other operators are the same as in MTL.}
That is $\neg$, $\wedge$, $\Un_{\Ic}$ and $\Next_{\Ic}$ are the negation, conjunction, timed until, and next time operators, respectively (see the review by \cite{BartocciCBMT2010sacs}).
\revisionTwo{
As an example of a simple formula in MTL$\times S4_u$, the expression $\Sexists p_1 \Un_{[0.5, 2.5]} \Sexists p_2$ is true if the set represented by $p_2$ is nonempty at some point in time between $[0.5, 2.5]$ and until then the set represented by $p_1$ is non-empty.
}

In the following, we define the bounded discrete-time semantics for MTL$\times S4_u$. 
The definition uses the spatial semantics of $S4_u$ while extending the temporal fragment (PTL) with time constraints over finite traces as in MTL. 
The semantics are defined over a data-object stream $\Data$. 
However, for consistency with PTL$\times S4_u$, 
we will assume the existence of a spatio-temporal valuation function 
$\sval: \Pi \times \mathbb{N} \rightarrow 2^{\Suniverse}$
that associates with every proposition $p$ and time frame $i$ a subset of the topological space. 
\revisionTwo{
In the definition of STPL in Section \ref{sec:STPL}, the sets $\sval(p,i)$ will correspond to identified objects in the environment, i.e., bounding boxes, bounding rectangles, or even regions in semantic segmentation. 
In this section, the the sets $\sval(p,i)$ are just arbitrary subsets of the universe.
}

\begin{defn}[Quantified Topological Temporal Model and Valuation]
\label{lbl:ext-temporal-topo-model}
    A Quantified Topological Temporal Model (QTTM) is a tuple of the form 
    \revision{$\SVAL=$} $\langle \Ts, \sval, \Data,  \tau \rangle$, 
    where 
    $\Ts = \langle \Suniverse,\Its \rangle$ is a TO topological space,
    $\sval$ is a spatio-temporal valuation function,
    $\Data$ is a data-object stream \revision{of size $|\mathcal{D}|$}, 
    $\tau: \mathbb{N} \rightarrow \mathbb{R^+}$ maps frame numbers to their physical times, and 
    $\Ic$ is any non-empty interval of  $\;\mathbb{R}_{\ge 0}$ over time.

    \revisionTwo{Given a model $\SVAL$,} 
    the valuation function 
    $\Sval(p, \Data, i, \tau)$ 
    represents a subset of the topological space $\Ts$ that is occupied by 
    spatial \revision{proposition} $p \in \Pi$ 
    in the $i$'th frame (e.g., $\tau(i)$ is the captured time).
    The valuation \revision{$\Sval$} is inductively extended to any formulas that can be produced by the production rule $\Stau$ in the Def. \ref{lbl:mtl_s4u_syntax-ext} as follows:
\begin{align*}
    & \Sval( p, \Data, i, \tau) :=
    \sval(p,i) 
    \displaybreak[2] 
    \\
	& \Sval( \Stau_1 \;\Un^{s}_{\Ic}\; \Stau_2, \Data,  i, \tau) := 
	\bigcup_{i' \in \{ j \in \mathbb{N} \; | \; \tau(j) \in (\tau(i)+\Ic) \} } 
	\\
    &\;\;\;\;\;\;\;\;\;\;\;\;\;\;\;\;\;\;\;\;\;\;
    \left ( \Sval(\Stau_2, \Data,  i', \tau) \cap
    \bigcap_{i \leq i'' <i'} \Sval(\Stau_1, \Data,  i'', \tau)
    \right )
	\displaybreak[2] 
	\\
	& \Sval( \bigcirc^{s}_{\Ic}\; \Stau, \Data,  i, \tau) := 
	\\
    &\left \{
    \begin{tabular}{ll}
    $\Sval( \Stau, \Data,  i+1, \tau)$ & if  $i+1 \revision{< |\mathcal{D}|} , \tau(i+1)\in (\tau(i) + \Ic)$\\ 
    $\Sempty$ & otherwise (i.e., an empty set) \\
    \end{tabular}
    \right .
    \displaybreak[2] 
    \\
    & \Sval( \Stau_1 \sqcap \Stau_2, \Data,  i, \tau) := 
    \Sval( \Stau_1, \Data,  i, \tau ) \cap \; \Sval( \Stau_2, \Data,  i, \tau) 
	\displaybreak[2] 
    \\
    & \Sval( \overline{\Stau}, \Data,  i, \tau) :=
    \overline{\Sval(\Stau, \Data, i, \tau)}
	\displaybreak[2] 
    \\
    & \Sval( \Itl\; \Stau, \Data,  i, \tau) := 
    \;  \Its \;\Sval( \Stau, \Data,  i, \tau)       
	\displaybreak[2] 
\end{align*}  
where $t+\Ic = \{ t'' \; | \; \exists t' \in \Ic \, .\, t'' = t + t' \}$. 
\end{defn}

\revisionTwo{
The valuation function $\Sval$ definition is straightforward for the spatial operations, i.e., $\bar{\;}, \sqcap, \Itl$; $\Sval$ is just applying the corresponding set theoretic operations,  i.e.,  $\bar{\;}, \cap, \Itl$.
The more interesting cases are the spatio-temporal ($\Un^{s}_{\Ic}, \Next^{s}_{\Ic}$)  operators.
The spatial-next operator $\Next^{s}_{\Ic} \Stau$ first checks if the next sample $(i+1)$ satisfies the timing constraints, i.e., $\tau(i+1)\in (\tau(i) + \Ic)$, and if so, it returns the set that $\Stau$ represents at time $(i+1)$.
The spatial until is a little bit more involved and it is better explained through derived operators. 
In the following, we define some of the commonly used derived operators:
}
\begin{itemize}
    \item \revisionTwo{The {\it spatially for all} operator $\Sforall$ checks if the spatial expression $\Stau$ represents a set which is the same as the universe:}
    $\Sforall\; \Stau \equiv \neg \;\Sexists\; \overline{\Stau},$
    
    \item \revisionTwo{The {\it spatial union} operator:} $\Stau_1 \sqcup \Stau_2 \equiv \overline{\overline{\Stau_1} \sqcap \overline{\Stau_2}},$
    
    \item \revisionTwo{The {\it spatial closure} operator:} $\Csl\; \Stau \equiv \overline{\Itl\;\overline{\Stau}},$
    
    \item \revisionTwo{The {\it spatial eventually} operator:} $\Diamond^{s}_{\Ic}\; \Stau \equiv \Suniverse \;\Un^{s}_{\Ic}\; \Stau$,  

    \item \revisionTwo{The {\it spatial always} operator:} $\Box^{s}_{\Ic}\; \Stau \equiv \overline{\Diamond^{s}_{\Ic}\; \overline{\Stau}}$.

\end{itemize}
\revisionTwo{
Notice that in the definition of the spatial eventually operator, we used the universe set $\Suniverse$ as a terminal even though the syntax in Def. \ref{lbl:mtl_s4u_syntax-ext} does not explicitly allow for that.
The universe ($\Suniverse$) and the empty set ($\emptyset$) can be defined as derived spatial expressions, i.e., 
$\Suniverse = p \sqcup \bar p$ and $\emptyset = p \sqcap \bar p$.
Therefore, if we replace $\Suniverse$ in the definition of $\Sval$ for $\Un^{s}_{\Ic}$, we can observe that $\Diamond^{s}_{\Ic}$ corresponds to the spatial union of the expression $\Stau$ over the time interval ${\Ic}$.
}

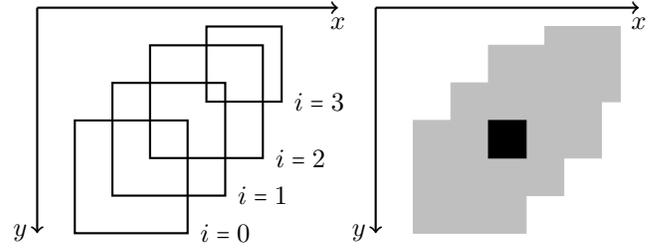
\begin{figure}
\centering
\begin{tikzpicture}
    \draw[thick,->] (0,3) -- (4,3) node[below] {$x$};
    \draw[thick,->] (0,3) -- (0,0) node[left] {$y$};
    \draw[thick] (0.5,0) rectangle (2,1.5);
    \draw[thick] (1,0.5) rectangle (2.5,2);
    \draw[thick] (1.5,1) rectangle (3,2.5);
    \draw[thick] (2.25,1.75) rectangle (3.25,2.75);
    \node at (2.5,0) {$i = 0$};
    \node at (3,0.5) {$i = 1$};
    \node at (3.5,1) {$i = 2$};    
    \node at (3.75,1.75) {$i = 3$};    
    \draw[thick,->] (4.5,3) -- (8,3) node[below] {$x$};
    \draw[thick,->] (4.5,3) -- (4.5,0) node[left] {$y$};
    \filldraw[gray!50] (5,0) rectangle (6.5,1.5);
    \filldraw[gray!50] (5.5,0.5) rectangle (7,2);
    \filldraw[gray!50] (6,1) rectangle (7.5,2.5);
    \filldraw[gray!50] (6.75,1.75) rectangle (7.75,2.75);
    \filldraw[black] (6,1) rectangle (6.5,1.5);
\end{tikzpicture}
\caption{
\revisionTwo{
{\bf Left}: the evolution of a spatial proposition $p$ over four frames with $\tau(0) = 0$, $\tau(1) = 0.4$, $\tau(0) = 0.8$, $\tau(0) = 1.2$; {\bf Right}: {\it Gray}: the set resulting from $\Diamond^{s} p$, and {\it Black:} the set resulting from $\Box^{s}_{[0,1]} p$.
}
}
\label{fig:exmp:spatio-temp-ops}
\end{figure}

\begin{exmp}
\label{exmp:simple:spatial:expr}
\revisionTwo{
A simple example is presented in Fig. \ref{fig:exmp:spatio-temp-ops} for a data-stream with 4 frames.
The spatial expression $\Diamond^{s} p$ corresponds to the union of all the sets represented by $p$ over time (gray set in Fig. \ref{fig:exmp:spatio-temp-ops}) since there are no timing constraints. 
On the other hand, the spatial expression $\Box^{s}_{[0,1]} \, p$ corresponds to the intersection of the sets of $p$ at frames $i = 0, 1, 2$ since the last frame ($i=3$) with $\tau(3) = 1.2$ does not satisfy the timing constraints $[0, 1]$.
The set $\Box^{s}_{[0,1]} \, p$ is represented as a black box in Fig. \ref{fig:exmp:spatio-temp-ops}.
}   
\end{exmp}


\revisionTwo{
Given the definition of the valuation function $\Sval$ for QTTM, we can proceed to define the semantics of MTL$\times S4_u$.
Recall that the valuation of spatial expressions returns sets from some topological space.
On the other hand, MTL$\times S4_u$ formulas are interpreted over Boolean values True ($\top$) / False ($\bot$), i.e., the formulas are satisfied or are not satisfied. 
To evaluate MTL$\times S4_u$ formulas, we use a valuation function $\dle \varphi \dri$ which takes as an input an MTL$\times S4_u$ formula $\varphi$, a data stream $\Data$, a sample $i$, and a timestamp function $\tau$, and returns True ($\top$) or False ($\bot$).
}

\begin{defn}[MTL$\times S4_u$ semantics]
\revisionTwo{
Given an MTL$\times S4_u$ formula $\varphi$, a QTTM $\SVAL=$ $\langle \Ts, \sval, \Data,  \tau \rangle$, and a frame $i \in \nat$,
the semantics of $\varphi$ are defined recursively as follows:
}
\begin{align*} 
	\allowdisplaybreaks
	&
\revisionTwo{
        \dle \Sexists \Stau \dri (\Data,i,\tau) :=   
        \left \{
         \begin{tabular}{ll}
            $\top$ & if  $\; \Sval(\Stau, \Data, i, \tau) \not = \Sempty$\\ 
            $\bot$ & otherwise (i.e., an empty set) \\
        \end{tabular}
        \right .
}
	\displaybreak[2] 
\\
	&
\revisionTwo{
        \dle \neg \phi \dri (\Data,i,\tau) :=   
	\neg 	\dle \phi \dri  (\Data,i,\tau) 
}
	\displaybreak[2] 
\\
	&
\revisionTwo{
	\dle \phi_1 \wedge \phi_2 \dri  (\Data,i,\tau) :=   
        \dle \phi_1 \dri (\Data,i,\tau) \; \wedge \dle \phi_2 \dri  (\Data,i,\tau)
}
	\displaybreak[2] 
\\
	&
\revisionTwo{
	\dle \phi_1 \;\Un_\Ic \; \phi_2 \dri  (\Data,i,\tau) :=  	
}
\\
	&
\revisionTwo{
 \bigvee_{j \in \{ k \in \mathbb{N} \; | \; \tau(k) \in (\tau(i)+\Ic) \} } 
 \bigg( \dle \phi_2 \dri  (\Data,j,\tau) \wedge  \bigwedge_{i \leq k <j} \dle \phi_1 \dri  (\Data,k,\tau) \bigg)
}
	\displaybreak[2] 
\\
	&
\revisionTwo{
 \dle \Next_\Ic \phi \dri (\Data,i,\tau) :=    
}
	\displaybreak[2] 
 \\
    & \;\;\; \left  \{ 
    \begin{tabular}{ll}
    \revisionTwo{$\dle \phi \dri  (\Data,i+1,\tau)$} & \revisionTwo{if  $i+1 \revision{< |\mathcal{D}|} , \tau(i+1)\in (\tau(i) + \Ic)$}\\ 
    \revisionTwo{$\bot$} & \revisionTwo{otherwise} \\
    \end{tabular}
    \right .
    \displaybreak[2] 
\end{align*}

\end{defn}



\revisionTwo{
Notice that the definitions of the propositional and temporal operators closely match the definitions of the spatial and spatio-temporal operators where set operations (union and intersection) have been replaced by Boolean operations (disjunction and conjunction).
Therefore, similar to the spatial expressions, we can define {\it disjunction} as $\varphi_1 \vee \varphi_2 \equiv \neg( \neg \varphi_1 \wedge \neg \varphi_2$), eventually as $\Diamond_\Ic \varphi \equiv \top \Un_\Ic \, \varphi$, and always as $\Box_\Ic \varphi \equiv \neg \Diamond_\Ic \neg \varphi$.
Finally, the constant {\it true} is defined as $\top \equiv \Sforall \, \Suniverse$. 
}

\begin{rem}\label{rem:frame-int}
In some specifications, it is easier to formalize a requirement using frame intervals and reason over frames rather than time intervals.  
To highlight this option, we add a tilde on top of the spatio-temporal operators with frame intervals, 
i.e., in $\tilde{\Un}^s_{\Ic}$, $\tilde{\Next}^s_{\Ic}$, $\tilde{\Un}_{\Ic}$, and $\tilde{\Next}_{\Ic}$ the interval $\Ic$ is over frame interval.
When reasoning over frames, \revision{$\tau(i)$ equals $i$}, i.e., $\tau$ is the identity function.
\revisionTwo{
}
\end{rem}

\begin{exmp}
\revisionTwo{
Revisiting Example \ref{exmp:simple:spatial:expr}, we can now introduce and explain some MTL$\times S4_u$ formulas.
The formulas $\Sexists \Diamond^{s} p$ and $\Sexists \Box^{s}_{[0,1]} \, p$ evaluate to true since the sets $\Diamond^{s} p$ and $ \Box^{s}_{[0,1]} \, p$ are not empty.
On the other hand, the formula $\Sexists \Box^{s} \, p$ is false because the set that corresponds to $\Box^{s} \, p$ is empty (in Fig. \ref{fig:exmp:spatio-temp-ops} there is no common subset for $p$ across all frames).
Notice that the MTL$\times S4_u$ formula $ \Box^{s} \, \Sexists \, p$ is true since the set $p$ is not empty at every frame. 
Finally, considering time stamps versus frames, the set $\Box^{s}_{[0,1]} \, p$, which considers timing constraints, is the same as the set $\tilde \Box^{s}_{[0,2]} \, p$, which considers frame constraints.
}
\end{exmp}



\section{Problem Definition}

Given a data stream $\Data$ as defined in \ref{def:data-obj-stream}, 
the goal of this paper is to:
\begin{enumerate}
    \item formulate object properties in such a data stream, and  
    \item monitor satisfiability of formulas over the stream.
\end{enumerate}

\subsection{Assumptions}
Given a data stream $\Data$, we assume that a tracking perception algorithm uniquely assigns identifiers to classified objects for all the frames \hlRev{(see the work by \cite{gordon2018re})}. 

In order to relax this assumption,
we need to enable probabilistic reasoning within the spatio-temporal framework, which is out of the scope of this work.
However, our framework could do some basic sanity checks about the relative positioning of the objects during a series of frames to detect misidentified objects.

This assumption is only needed for some certain types of requirements and for the rest it can be lifted.

\subsection{Overall Solution}

We define syntax and semantics of Spatio-Temporal Perception Logic (STPL) over sets of points in topological space, and quantifiers over objects.
We build a monitoring algorithm over TPTL, MTL, and $S4_u$ and show the practicality, expressivity and efficiency of the algorithm by presenting examples.
This is a powerful language and monitoring algorithm with many applications for verification and testing of complex perception systems.
Our proposed language and its monitoring algorithm are different than prior works as we discussed in the introduction (see also related works in Section \ref{sec:related}), and briefly summarized below.
STPL
\begin{itemize}
    \item reasons over spatial and temporal properties of objects through functions and relations;
    \item extends the reasoning of prior set-based logics by equipping them with efficient offline monitoring algorithm tools;
    \item focuses on expressing functional requirements for perception systems; and
    \item is supported by open source monitoring tools.
\end{itemize}

\section{Spatio-Temporal Perception Logic}
\label{sec:STPL}

\revisionTwo{
In this section, we present the syntax and semantics of the Spatio-Temporal Perception Logic (STPL).
Our proposed logic evaluates the geometric relations of objects in a data stream and their evolution over time.
Theoretically, any set-based topological space can be used in our logic.
However, we focus on topological spaces and geometric operations relevant to applications related to perception.
}

Next, we are going to define and interpret STPL formulas over data-object streams.
We define our topological space to be axis aligned rectangles in 2D images, 
or arbitrary polyhedral sets (potentially boxes) in 3D environments.
An axis aligned rectangle can be represented by a set of two points that correspond to two non-adjacent corners of the rectangle with four sides each parallel to an axis. 
A polyhedral is a set formed by the intersection of a finite number of closed half spaces.
We assume that this information is contained in the perception datastream as annotations or attributes for each object or region identified by the perception system.
For instance, $\Retrieve(\Data(i),id).\CS$ may store a rectangular axis-aligned convex-polygon that is associated with an object $id$ in the $i$'th frame.
Alternatively, $\Retrieve(\Data(i),id).\CS$ may store a convex-polyhedron.
Without loss of generality, we will typically use definitions for 2D spaces (image spaces), and later in the case study, we will use bounding volumes (3D spaces).

\subsection{STPL Syntax}
The syntax of STPL is based on the syntax of TQTL (\cite{dokhanchi2018evaluating}) and ${\mathcal S}4_u$ (\cite{gabelaia2005combining}).

In the context of STPL, the spatial propositions are the symbols that correspond to objects or regions in the perception dataset.
We use the function symbol $\BB$ to map these objects to their corresponding sets.
\revisionTwo{
Hence, the syntax of spatial terms $\Stau$ in STPL is the same as in MTL$\times S4_u$, but the function symbols $\BB$ replace the spatial propositions $p$.
In addition, we add grammar support under production rule $\mathcal{A}$ for area computation for spatial terms.
Area computation is needed in standard performance metrics for 2D vision algorithms, for example to compute union over intersection (UoI).
In the same spirit, we can also support volume computation for spatial terms in 3D spaces.
}

\revisionTwo{
For many requirements, we also need to access other attributes of the objects in the datastream, e.g., classes, probabilities, estimated velocities, or even compute some basic functions on object attributes, e.g., distances between bounding boxes.
For usability, we define some basic functions to retrieve data and compute the desired properties.
The functions which are currently supported are: 
object class ($C$),
class membership probability ($P$),
latitude ($Lat$) and longitude ($Lon$) coordinates of a point ($CRT$),
area of bounding box ($Area$), and
distance ($Dist$) between two points ($CRT$).
We provide the syntax for using such functions under the production rule $\Theta$.
}

This set of functions is sufficient to demonstrate the generality of our logic.
However, a more general approach would be to define arithmetic expressions over user definable functions which can retrieve any desired data from the datastream.
\revisionTwo{
Such functionality support is going to be among the goals of future software releases. 
}

\begin{defn}[STPL Syntax over Discrete-Time Signals] 
\label{lbl:stql-brief-syntax}
\revision{Let $V_t$ and $V_o$ be sets of time variables and object variables, respectively.
}
Assume that $x \in V_t$ is a time variable,
$id \in V_o$ is an object variable, 
$\Ic$ is any non-empty interval of $\;\mathbb{R}_{\ge 0}$ over time.
The syntax for Spatio-Temporal Perception Logic (STPL) formulas is provided by the following grammar starting from the production rule $\phi$:
\\
\\
\textit{The syntax for spatial terms is:}
\begin{align*}
    \Stau &::= \BB(id) \;|\; 
    \overline{\Stau} \;|\; \Stau \sqcap \Stau \;|\; \Itl\;{\Stau} \;|\;
    \Stau \;\Un^{s}_{\Ic}\; \Stau \;|\; 
    \Next^{s}_{\Ic}\; \Stau 
\end{align*}
\textit{The syntax for the functions that compute the area of a spatial term are:}
\begin{align*}
     \mathcal{A} & \revisionTwo{ \; ::= Area(\Stau) \sim r \;|\; Area(\Stau) \sim r \times Area(\Stau) }
\end{align*}
\textit{The atomic propositions that represent coordinates for a bounding box are:}
\begin{align*}
    \CRT &::= \LM \;|\; \RM \;|\; \TM \;|\; \BM \;|\; \CT
\end{align*}
\textit{The syntax for spatio-temporal functions are:}
\begin{align*}
    \Theta ::=  & \revisionTwo{ \; Dist(id,\CRT,id,\CRT) \sim r \;|\; }
    \\
    & \revisionTwo{ \; Lat(id,\CRT) \sim r \;|\; Lon(id,\CRT) \sim r \;|\; }
	\displaybreak[2] 
    \\
    & \revisionTwo{ \; Lat(id,\CRT) \sim r \times Lat(id,\CRT) \;|\;     }
	\displaybreak[2] 
    \\
    & \revisionTwo{ \;    Lon(id,\CRT) \sim r \times Lon(id,\CRT) \;|\; }
	\displaybreak[2] 
    \\
    & \revisionTwo{ \; Lat(id,\CRT) \sim r \times Lon(id,\CRT) \;|\; }
	\displaybreak[2] 
    \\
    & \revisionTwo{ \; Area(id) \sim r \;|\; Area(id) \sim r \times Area(id) \;|\; }
	\displaybreak[2] 
    \\
    &  \; C(id) = c \;|\; C(id) = C(id) \;|\; 
	\displaybreak[2] 
    \\
    & \revisionTwo{ \; P(id) \sim a \;|\;  P(id) \sim \revision{r} \times P(id) }
\end{align*}
\textrm{The syntax for the STPL formula is:}
\begin{align*}
    \phi ::= & \; \top \;|\; x.\phi \;|\; \exists id@x.\phi \;|\; \exists id.\phi \;|\; id = id \;|\;
    \neg \phi\;|\; \phi \vee \phi \;|\; 
    \\ 
    &  \Ctime - x \sim t \;|\; \Cframe - x \sim n \;|\; \Cframe - x \;\%\; c \sim n \;|\;
    \\
    &  \bigcirc \phi \;|\; \phi \;\Un\; \phi \;|\; \Previous \phi \;|\; \phi \;\Since \;\phi \;|\;
    \\
    &  \Theta \;|\; \Sexists\; \Stau \;|\; \mathcal{A}
\end{align*}
where 
$\top$ is the symbol for {\it true},
$\sim \in \{<,>,\geq,\leq,= \}$, and $r\in \mathbb{R}_{\geq 0}, c \in \mathbb{N}$, and $a \in [0,1]$ are constants.
Here, $\BB: V_o \rightarrow \Pi$ is a function that maps object variables into spatial propositions.
\end{defn}

\revisionTwo{
The syntax of STPL is substantially different from the syntax of \mtlsfu.
}
In the STPL syntax, $x.\phi$ stands for the freeze time quantifier.
When this expression is evaluated, the corresponding frame $i$ is stored in the  clock variable $x$.
The prefix $\exists id$ in the rule $\exists id @x.\varphi$ or the rule $\exists id .\varphi$ is the \textit{Existential} object quantifier.
\revisionTwo{
When the formula $\exists id @x.\varphi(id)$ is evaluated, then it is satisfied (true) when there is an object $id$ at frame/time $x$ that makes $\varphi(id)$ true.
The formula $\exists id .\varphi(id)$ is also searching for an object that makes $\varphi(id)$ true, but in this case we do not need to refer to the time that the object was selected.
}
Similarly, the \textit{Universal} object quantifier is defined as 
$ \forall id @x.\phi\equiv \neg(\exists id @x.\neg\phi)$ or $ \forall id .\phi\equiv \neg(\exists id .\neg\phi)$.
\revisionTwo{
The universal quantifier requires that all the objects in a frame satisfy the subformula $\varphi$.
Notice that the syntax of STPL cannot enforce that all uses of time or object variables are bound, i.e., declared before use.
In practice, such errors can be detected after the parse tree of the formula has been constructed through a simple tree traversal.
}

\revisionTwo{
In contrast to \mtlsfu, the timing constraints in STPL are not annotating the temporal operators, but rather they are explicitly stated as arithmetic expressions in the formulas.
For example, consider the specification ``{\it There must exist a car now which will have class membership probability greater than 90\% within 1.5 sec}".
With timing constraints annotating the temporal operators, the requirement would be: 
\[ \exists id . \big( C(id) = \texttt{car} \wedge \Diamond_{[0,1.5]} P(id) > 0.9\big).\]
Since STPL uses time variables and arithmetic expressions to express timing constraints, the same requirement may be written as:
\[ \exists id @ x . \big( C(id) = \texttt{car} \wedge \Diamond (\tau-x\leq 1.5 \wedge  P(id) > 0.9)\big).\]
The use of time variables enables the requirements engineer to define more complex timing requirements and to refer to objects at specific instances in time.
}
The time, frame, and object constraints of STPL formulas are in the form of $\Ctime - x > r$, $\Cframe - x > n$, and $id=id$, respectively. 
We denote $\Ctime -x$ and $\Cframe -x$ to refer to the elapsed time and frames, respectively.
Note that we use the same variable to refer to the freeze time and frame, but distinguish their type based on how they are used in the constraints ($\Ctime$ represents the current time, and $\Cframe$ represents the current frame number).
\revision{
The operator $\%$ in the expression $\Cframe - x \;\%\; c \sim n$ is used to specify periodic constraints.
For reasoning over the past time, we added $\Previous$ (\textit{previous}) and $\Since$ (\textit{since}) operators as duality for the $\Next$ and $\Un$ operators, respectively.  
}

For STPL formulas $\psi$, $\phi$, we define $\psi\wedge\phi\equiv\neg(\neg\psi\vee\neg\phi)$, $\bot\equiv\neg\top$ (False), $\psi\rightarrow\phi\equiv\neg\psi\vee\phi$ ($\psi$ Implies $\phi$),
$\phi \;\Rc\; \psi \equiv \neg (\neg \phi \;\Un\; \neg \psi)$ ($\phi$ releases $\psi$),
$\phi \;\overline{\Rc}\; \psi \equiv \phi \;\Rc\; (\phi \vee \psi)$ ($\phi$ non-strictly releases $\psi$),
$\Diamond\psi \equiv \top \;\Un\; \psi$ (Eventually $\psi$),
$\Box\psi \equiv \neg \Diamond \neg\psi$ (Always $\psi$)
using syntactic manipulation.

\begin{rem}
\label{rem:multi:classes}
\revision{
In principle, in STPL, it is easy to add multiple classes and classification probabilities for each object. 
We would simply need to replace in Def. \ref{lbl:stql-brief-syntax} the rules
\[ 
C(id) = c \;|\; P(id) \sim a \;|\; P(id) \sim \revision{r} \times P(id) 
\]
in $\Theta$ with 
\[
c \in C(id) \;|\; P(id, c) \sim a \;|\; P(id, c) \sim a \times P(id, c).
\]
where $C(id)$ is now a function which returns a set of classes for the object.
In order to write meaningful requirements over multiple classes, we would also need to introduce quantifiers over arbitrary sets.
That is, we should be able to write a formula such as ``{\it there exists at least one class with probability greater than $a$}": $\exists c \in C(Id) \, . \, P(Id,c) > a$.
This is within the scope of First Order Temporal Logics (e.g., see \cite{BasinKMZ2015acm} and \cite{HavelundPU2020fmsd}) and we plan to consider such additions in the future.
}
\end{rem}


\revisionTwo{
In general, the monitoring problem for STPL formulas is PSPACE-hard (since STPL subsumes Timed Proposition Temporal Logic (TPTL); see \cite{MarkeyR2006tcs}). 
However, there exists a fragment of STPL which is efficiently monitorable (in polynomial time).
}

\begin{defn}[Almost Arbitrarily \revision{Nested} Formula] \label{lbl-AAN-formula}
\hlRev{An Almost Arbitrarily \revision{Nested} (AAN) formula is an STPL formula in which 
no time or object variables are used in the scope of another freeze time operator.}
\end{defn}
\noindent For example,
\begin{align*}
\varphi_1 & = x_1. \Box \big(  \Cframe - x_1 > 2 \wedge 
x_2. \Diamond ( 
\Ctime - x_2 < 0.01 ) \big)
\\
\varphi_2 & = \Box \exists Id_1@x.\Diamond \big(\Box \forall Id_2. 
(Id_1 = Id_2) \wedge (\Ctime - x > 2) \big)
\end{align*}
are AAN formulas.
\hlRev{In formula $\varphi_1$, the time variable $x_1$ is not used in the scope of the $x_2$ freeze time operator. 
In formula $\varphi_2$, there is no nested quantifier/freeze time operators.
Therefore, they are both ANN STPL formulas.
On the other hand,}
\begin{align*}
\varphi_3  = & x_1.\big( \Box x_2. \Diamond 
(\Cframe - x_1 > 2 
\wedge \;\Ctime - x_2 < 0.01 ) \big)
\\
\varphi_4  = & \Box\forall Id_1@x_1.\Next \forall Id_2@x_2.\Next \Box \forall Id_3.
\\
& \big( P(Id_3) > P(Id_1) \wedge P(Id_3) < P(Id_2)\big)
\end{align*}
are not AAN formulas.
That is because in $\varphi_3$, the variable $x_1$ is used in the scope of the nested quantifier operator ``$x_2.$''.
In $\varphi_4$, the $Id_1$ is used in the scope of the second nested quantifier operator ``$x_2.$''.

The authors in \cite{dokhanchi2016efficient} presented an efficient monitoring algorithm for TPTL formulas with an arbitrary number of \textit{independent} time variables.
Our definition of AAN formulas is adopted from their definition of \textit{encapsulated} TPTL formulas that are TPTL formulas with only independent time variables in them. 
\begin{rem}
    \revision{
    Our definition of AAN formulas is syntactical.
    However, there can be syntactically AAN formulas that can be rewritten as semantically equivalent non-AAN formulas. 
    There are other ways to define formulas that are not syntactically ANN but semantically equivalent to AAN formulas.
    The precise mathematical definitions of these formulas is beyond the scope of this paper.
    In the following, we will show examples of AAN formulas and some other formulas that are equivalent to AAN formulas.
    For example, 
    \[ 
    \varphi_5 = \Box \;x. \Box \;y. \big( (\Ctime - x > 1) \implies \Box ( \Ctime - y > 2)\big) 
    \]
    which is not an AAN formula can be rewritten as AAN
    \[
       {\varphi_5}'  = \Box \;x. \Box \big( (\Ctime - x > 1)  \implies y.\Box ( \Ctime - y > 2)\big). 
    \] 
    On the other hand, although 
    \begin{align*}
    \varphi_6  = & \Box \forall id_1@x. \Box \forall id_2@y. \\ 
    & \big( (C(id_1)=C(id_2)) \implies ( P(id_2) > 0.9 \big) 
    \end{align*}
    is not an AAN formula, our monitoring tool supports it since the clock variables are not used.
    That is, it can be written as:
    \begin{align*}
        \varphi_6'  = & \Box \forall id_1 @x. \Box \forall id_2 . \\ 
        & \big( (C(id_1)=C(id_2)) \implies ( P(id_2) > 0.9 )\big) 
    \end{align*}
    }
\end{rem}
\subsection{STPL Semantics}
\label{sec:stpl-semantics}
\revision{
Before getting into the semantics of the STPL, we represent spatio-temporal function definitions that are used in production rules $\mathcal{A},$ and $\Theta$.
The definitions of these functions are independent of the semantics of the STPL language, and therefore, we can extend them to increase the expressivity of the language.
}
\subsubsection{Spatio-Temporal Functions}
\revision{We define functions $f$ as follows:}
\begin{itemize}

\item \revision{Class of an object:}
\\
$\boldsymbol{f_C(id ,\Data,i,\epsilon,\zeta)}$ 
returns the $\Retrieve$ $(\Data$ $(k)$ $,\epsilon(id ))$ $.Class$ as the \revision{class of the $\epsilon(id )$ object} in the $k$'th frame, 
where $k \leftarrow \zeta(id )$ if $\zeta(id )$ is specified, and $k \leftarrow i$ otherwise.  

\item \revision{Probability of a classified object:}
\\
$\boldsymbol{f_P(id ,\Data,i,\epsilon,\zeta)}$ 
returns the $\Retrieve($ $\Data(k)$ $,\epsilon(id ))$ $.Prob$ 
as the probability of \revision{the $\epsilon(id )$ object} in the $k$'th frame,
where $k \leftarrow \zeta(id )$ if $\zeta(id )$ is specified, and $k \leftarrow i$ otherwise. 

\item \revision{Distance between two points from two different objects:}
\\
$\boldsymbol{f_{Dist}(id_j,id_k, \CRT_1 , \CRT_2 ,\Data,i,\epsilon,\zeta)}$ 
computes and returns the \textit{Euclidean Distance} between the $\CRT_1$ point of \revision{the $\epsilon(id_j)$ object} and $\CRT_2$ point of \revision{the $\epsilon(id_k)$ object} in the $k_1$'th and  $k_2$'th frames, respectively; and we have
$k_1 \leftarrow \zeta(id_j)$ if $\zeta(id_j)$ is specified, and $k_1 \leftarrow i$ otherwise, and
$k_2 \leftarrow \zeta(id_k)$ if $\zeta(id_k)$ is specified, and $k_2 \leftarrow i$ otherwise.

\item \revision{Lateral distance of a point that belongs to an object in a coordinate system (for image space, see section \ref{sec:image-space}):}
\\
$\boldsymbol{f_{LAT}(id , \CRT ,\Data,i,\epsilon,\zeta)}$
computes and returns the \textit{Lateral Distance} of the $\CRT$ point of \revision{the $\epsilon(id )$ object} in the $k$'th frame from the \textit{Longitudinal axis},
where $k \leftarrow \zeta(id )$ if $\zeta(id )$ is specified, and $k \leftarrow i$ otherwise.  

\item \revision{Longitudinal distance of a point that belongs to an object in a coordinate system (for image space, see section \ref{sec:image-space}):}
\\
$\boldsymbol{f_{LON}(id , \CRT ,\Data,i,\epsilon,\zeta)}$ 
computes and returns the \textit{Longitudinal Distance} of the $\CRT$ point of \revision{the $\epsilon(id )$ object} in the $k$'th frame from the \textit{Lateral axis},
where $k \leftarrow \zeta(id )$ if $\zeta(id )$ is specified, and $k \leftarrow i$ otherwise.  

\item \revision{Area of a region:}
\\
$\boldsymbol{f_{Area}(\Stau)}$ computes and returns the area of an spatial term $\Stau$ if $\Stau$ is specified, otherwise it returns unspecified.

\end{itemize}

In the above functions, we use identifier variables to refer to objects in a data stream.
Some parameters are point specifiers to choose a single point from all the points that belong to an object.
The only function with a spatial term as an argument is the \textit{area} function that calculates the area of a 2D bounding box.
\revision{
Similar reasoning for a 3D geometric shape (polyhedra in general) using a volume function is possible.
This is supported by our open-source software \cite{stpl_tools}, but we don't provide formal definitions here for the brevity of the presentation.
}

All the functions have $\Data$, $i$, $\epsilon$ and $\zeta$ as arguments.
Functions need the data stream $\Data$ to access and retrieve perception data, 
the frame number $i$ to access a specific frame in time (``current" frame), 
a data structure (typically referred to as ``environment") $\epsilon$ to store and retrieve the values assigned to variables (both for time and objects), 
and, finally, 
\revisionTwo{
a data structure $\zeta$ to acquire the time/frame at which an object was selected through quantification.
The data structure $\zeta$ is used to distinguish whether the time that an object was selected is needed or not (see semantics for $\exists$ in Section \ref{stql:smt:tmp}).
}
For example, if we would like to store in the object variable $Id_1$, the value $k$ as the identifier of an object in a frame, 
and store in the frame variable \textit{x}, the value $i$ as the frozen frame,
then we would write  
\begin{center}
$\epsilon [Id_1 \leftarrow k] \qquad $  and $ \qquad \epsilon [x \leftarrow i]$
\end{center}
respectively. 
Similarly, we write 
\revisionTwo{
\[ \zeta [Id_1 \leftarrow i] \] 
to denote that we store the frame number $i$ when the object variable $Id_1$ was set. 
}
The initial data structures (environments) $\epsilon_0$ and $\zeta_0$ are empty.
That is, no information is stored about any object identifier of the clock variable.
The environment $\zeta$ is needed for the comparison of objects, probabilities, etc over different points in time, and 
environment $\epsilon$ is needed for the comparison of time/frame constraints and quantification of the object variables.

Note that the above functions are application dependent, and we can add to this list based on future needs.
In the next section, when we introduce an example of 3D space reasoning, we present a new spatio-temporal function.


\begin{defn}[Semantics of STPL]\label{stql:smt}
    Consider $\Tm=$ $\langle \Ts,$ $\Sval,$ $\Data,$ $\epsilon,$ $\zeta$, \revision{$\tau$}$\rangle$ as a topological temporal model in which
    $\Ts$ is a topological space,
    $\Sval$ is a spatial valuation function,
    $\Data$ is a data-object stream,
	$\zeta: V_o \rightarrow \mathbb{N} \cup \{\revision{\bot}\}$ is an object evaluating function, and 
	$\epsilon: V_t \cup V_o 
	\rightarrow \mathbb{N}$ is a frame-evaluating function.
	Here, $i \in \mathbb{N}$ is the index of current frame, 
    $\tau: \mathbb{N} \rightarrow \mathbb{R^+}$ is a mapping function from frame numbers to their physical times,
	$\phi,\phi_1,\phi_2 \in STPL$ (i.e., formulas belong to the language of the grammar of STPL),
    $V_t$ is a set of time variables, and 
    $V_o$ is a set of object variables.
	The quality value of formula $\phi$ with respect to $\Data$ at frame i with evaluations $\epsilon$ and $\zeta$ is recursively assigned as follows:

\subsubsection{Semantics of Temporal Operators}\label{stql:smt:tmp}
\begin{align*} 
	\allowdisplaybreaks
	&\dle \top \dri (\Data,i,\epsilon,\zeta,\revision{\tau}) :=   \top
	\displaybreak[2] 
	\\
	\allowdisplaybreaks
	&\dle x.\phi \dri (\Data,i,\epsilon,\zeta,\revision{\tau}) := \dle \phi \dri  (\Data,i,\epsilon[
        x\Leftarrow i
    ], \zeta, \tau)
	\displaybreak[2] 
	\\
    &\dle \exists id @x.\phi \dri  (\Data,i,\epsilon,\zeta,\revision{\tau}):= 
    \\
    &\;\;\;\;\;
    \bigvee_{ k \in\sorder(\Data(i))}
    \Big(\dle \phi \dri  (\Data,i,\epsilon[
    id \Leftarrow k, 
    x\Leftarrow i
    ],
    \zeta[id \Leftarrow i])\Big)
	\displaybreak[2] 
	\\
    &\dle \exists id.\phi \dri  (\Data,i,\epsilon,\zeta,\revision{\tau}):= 
    \\
    &\;\;\;\;\;
    \bigvee_{ k \in\sorder(\Data(i))} \Big(\dle \phi \dri  (\Data,i,\epsilon[
    id \Leftarrow k
    ],
    \zeta[id \Leftarrow \bot])\Big)
    \\
    &\dle \Ctime - x \sim n\dri  (\Data,i,\epsilon,\zeta,\revision{\tau}):= \displaybreak[2] \left\{ \begin{array}{ll}
	\top
	\mbox{ if \;\;} \tau(i) - \tau(x) \sim n
	\\
	\bot
	\mbox{ otherwise }  
	\\
	\end{array} \right.
	\\ 
    &\dle \Cframe - x \sim n\dri  (\Data,i,\epsilon,\zeta,\revision{\tau}):= \displaybreak[2] \left\{ \begin{array}{ll}
	\top
	\mbox{ if \;\;} i - x \sim n
	\\
	\bot
	\mbox{ otherwise }  
	\\
	\end{array} \right.
	\\
    &\dle \Cframe - x \;\%\; c \sim n\dri  (\Data,i,\epsilon,\zeta,\revision{\tau}):= 
    \displaybreak[2] \left\{ \begin{array}{ll}
	\top
	\mbox{ if \;\;} (i - x) \;\%\; c \sim n
	\\
	\bot
	\mbox{ otherwise }  
	\\
	\end{array} \right.
	\\	
    &\dle id_j = id_k \dri  (\Data,i,\epsilon,\zeta,\revision{\tau}):= \displaybreak[2] \left\{ \begin{array}{ll}
	\top
	\mbox{ if \;\;} \epsilon(id_j) = \epsilon(id_k)
	\\
	\bot
	\mbox{ otherwise }  
	\\
	\end{array} \right.
	\\	
	&\dle \neg \phi \dri (\Data,i,\epsilon,\zeta,\revision{\tau}) :=   
	\neg
	\dle \phi \dri  (\Data,i,\epsilon,\zeta,\revision{\tau}) 
	\displaybreak[2] 
	\\
	&\dle \phi_1 \vee \phi_2 \dri  (\Data,i,\epsilon,\zeta,\revision{\tau}) :=   
        \dle \phi_1 \dri (\Data,i,\epsilon,\zeta,\revision{\tau}) \; \vee 
	\displaybreak[2] 
	\\
        & \qquad \qquad \qquad \qquad \qquad \quad 
        \dle \phi_2 \dri  (\Data,i,\epsilon,\zeta,\revision{\tau}) \displaybreak[2]  
	\\
	&\dle \phi_1 \;\Un\; \phi_2 \dri  (\Data,i,\epsilon,\zeta,\revision{\tau}) :=  
	\\
	&\;\;\;\;\;\;\;\;\;\;\;\;\bigvee_{i \le j} \bigg( \dle \phi_2 \dri  (\Data,j,\epsilon,\zeta,\revision{\tau}) \wedge  \bigwedge_{i \leq k <j} \dle \phi_1 \dri  (\Data,k,\epsilon,\zeta,\revision{\tau}) \bigg)
	\\
    &\dle \Next \phi \dri (\Data,i,\epsilon,\zeta,\revision{\tau}) :=    \dle \phi \dri  (\Data,i+1,\epsilon,\zeta,\revision{\tau})
\end{align*}

\hlRev{\subsubsection{Semantics of Past-Time Operators}\label{stpl:past-time}
\revision{
Many applications require requirements that refer to past time.
Past time operators are particularly relevant to online monitoring algorithms.
}
\begin{align*}
	&\dle \phi_1 \;\Since\; \phi_2 \dri  (\Data,i,\epsilon,\zeta,\revision{\tau}) :=  
	\\
	&\;\;\;\;\;\;\;\;\;\;\;\;\bigvee_{i \ge j} \bigg( \dle \phi_2 \dri  (\Data,j,\epsilon,\zeta,\revision{\tau}) \wedge  \bigwedge_{j < k \le i} \dle \phi_1 \dri  (\Data,k,\epsilon,\zeta,\revision{\tau}) \bigg)
	\\
    &\dle \Previous \;\phi \dri (\Data,i,\epsilon,\zeta,\revision{\tau}) :=    \dle \phi \dri  (\Data,i-1,\epsilon,\zeta,\revision{\tau}) 
	\displaybreak[2] 
\end{align*} 
}
\subsubsection{Semantics of Spatio-Temporal Operators}\label{stql:smt:sptmp}
\revision{Now we define the operators and functions that are needed for capturing the requirements 
that reason over different properties of data objects in a data stream.
Note that in some of the following definitions, we used $\Sval$ to avoid rewriting the semantics
we built up upon the syntax and semantics as part of $MTL \times S4_u$ logic.
The semantics of STPL will be based on the valuation function $\Sval$ of $MTL \times S4_u$ 
with the change that $\Sval$ will now also accept the environments $\epsilon$ and $\zeta$, 
and we extend the definition of $\Sval$ over the spatial term $\BB(id)$.
That is, we replace in the semantics of the $MTL \times S4_u$ the valuation of spatial propositions $p$ with:
\[\Sval(\BB(id), \Data, i, \tau, \epsilon, \zeta) = \sval(\epsilon(id), k),\]
where $k \leftarrow \zeta(id )$ if $\zeta(id )$ is specified, and $k \leftarrow i$ otherwise.
The semantics for the rest of the spatial operators are:
}
\begin{align*}
	&\dle \Sexists\; \Stau \dri (\Data,i,\epsilon,\zeta,\revision{\tau}) := 
	\revision{\Sval}(\Sexists\;\Stau, \Data, i, \tau, \revision{\epsilon, \zeta})
	\displaybreak[2] 
	\\
	&\dle Area(\Stau) \sim r \dri (\Data,i,\epsilon,\zeta,\revision{\tau}) := 
	\\
	&\;\;\;\;\;\;\;\;\;\;\;\;\displaybreak[2] \left\{ \begin{array}{ll}
	    \bot
	    \mbox{ if \;\;}  f_{Area}\big(\revision{\Sval}(\Stau,\Data,i,\tau, \revision{\epsilon, \zeta})\big) \mbox{ is unspecified}
	    \\
	    \top
	    \mbox{ if \;\;} f_{Area}\big(\revision{\Sval}(\Stau,\Data,i,\tau, \revision{\epsilon, \zeta})\big) \sim r
	    \\
	    \bot
	    \mbox{ otherwise}  
	    \end{array} \right.
	\displaybreak[2] 
	\\
	&\dle C(id) = r \dri (\Data,i,\epsilon,\zeta,\revision{\tau}) := 
	\\
	&\;\;\;\;\;\;\;\;\;\;\;\;\displaybreak[2] \left\{ \begin{array}{ll}
	    \top
	    \mbox{ if \;\;} f_{C}\big(id,\Data,i,\epsilon,\zeta)\big) = r
	    \\
	    \bot
	    \mbox{ otherwise}^*  
	    \end{array} \right.
		\displaybreak[2] 
    \\
    &\dle P(id) \sim r \dri (\Data,i,\epsilon,\zeta,\revision{\tau}) := 
	\\
	&\;\;\;\;\;\;\;\;\;\;\;\;\displaybreak[2] \left\{ \begin{array}{ll}
	    \top
	    \mbox{ if \;\;} f_{P}\big(id,\Data,i,\epsilon,\zeta)\big) \sim r
	    \\
	    \bot
	    \mbox{ otherwise}^*  
	    \end{array} \right.
		\displaybreak[2] 
    \\
	&\dle Dist(id_j, \CRT_1 ,id_k, \CRT_2 ) \sim r \dri (\Data,i,\epsilon,\zeta,\revision{\tau}) := 
	\\
	&\;\;\;\;\;\;\;\;\;\;\;\;\displaybreak[2] \left\{ \begin{array}{ll}
	    \top
	    \mbox{ if \;\;} f_{Dist}\big(id_j,id_k, \CRT_1 , \CRT_2 ,\Data,i,\epsilon,\zeta)\big) \sim r
	    \\
	    \bot
	    \mbox{ otherwise}^*  
	    \end{array} \right.
	\displaybreak[2] 
	\\
	&\dle LAT(id , \CRT ) \sim r \dri (\Data,i,\epsilon,\zeta,\revision{\tau}) := 
	\\
	&\;\;\;\;\;\;\;\;\;\;\;\;\displaybreak[2] \left\{ \begin{array}{ll}
	    \top
	    \mbox{ if \;\;} f_{LAT}\big(id , \CRT ,\Data,i,\epsilon,\zeta)\big) \sim r
	    \\
	    \bot
	    \mbox{ otherwise}^*  
	    \end{array} \right.
	\displaybreak[2] 
	\\
	&\dle LON(id , \CRT ) \sim r \dri (\Data,i,\epsilon,\zeta,\revision{\tau}) := 
	\\
	&\;\;\;\;\;\;\;\;\;\;\;\;\displaybreak[2] \left\{ \begin{array}{ll}
	    \top
	    \mbox{ if \;\;} f_{LON}\big(id , \CRT ,\Data,i,\epsilon,\zeta)\big) \sim r
	    \\
	    \bot
	    \mbox{ otherwise}^*  
	    \end{array} \right.
	\displaybreak[2] 
\end{align*}
where $\sim \in \{>,<,=,\ge,\le\}$, and ``otherwise$^*$'' has a higher priority to become true in the \textit{if statements} if $\epsilon(id ) \not\in \sorder(\Data(i))$.
\end{defn}

Note that we omit the semantics for some of the spatio-temporal operators in the production rule $\Theta$ except 
the ones stated above
due to their similarity to the semantics of the presented operators.

\begin{defn}[STPL Satisfiability]
We say that the data stream $\Data$ satisfies the STPL formula $\varphi$ under the current environment iff $\;\dle \varphi \dri(\Data,0,\epsilon_0,\zeta_0, \revision{\tau}) = \top$, which is equivalent to denote $(\Data,0,\epsilon_0,\zeta_0, \revision{\tau}) \models \varphi$. 
\end{defn}
Note that by $\epsilon_0$ and $\zeta_0$, we reset all the variables to \revision{be empty}.
\revision{}
This enables the use of the presented proof system of TPTL by \cite{dokhanchi2016efficient}.

\begin{figure*}[hbt!]
  \vspace{0.75em}
    \centering
    \begin{subfigure}{.95\linewidth}
    \includegraphics[width=1\linewidth]
    {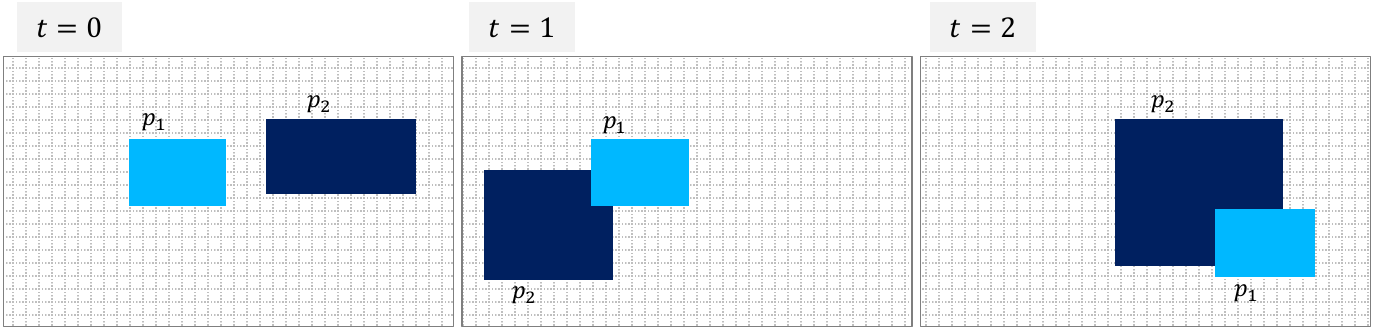}
    \end{subfigure}
    \caption{
    An object-data stream $\Data$ of three frames illustrated in three time steps $t_0,t_1$, and $t_2$.
    In each frame, there are two rectangular geometric shapes denoted by identifiers $1$ and $2$. 
    The lighter blue colored rectangle and the darker blue colored rectangle are identified by $p_1$ and $p_2$, respectively.
    }
    \label{fig:exm:data-stream}
\end{figure*}


\begin{figure*}[hbt!]
    \centering
    \begin{subfigure}{.95\linewidth}
    \includegraphics[width=1\linewidth]
    {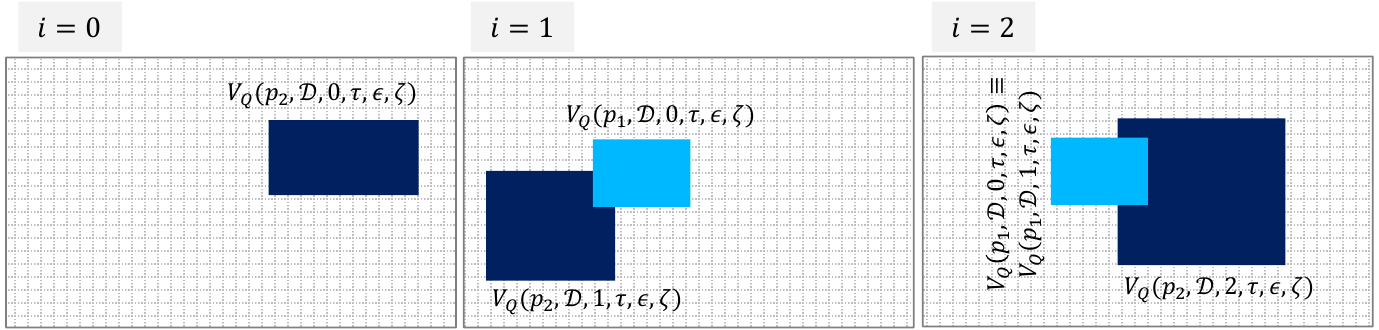}
    \end{subfigure}
    \caption{
    The step-by-step computation of $\Sval( p_1 \Un^{s}_{[0,2]} p_2 , \Data, 0, \tau, \revision{\epsilon})$ for $i=0$ to $2$.
    Here, $p_1$ and $p_2$ are the same as in Fig. \ref{fig:exm:data-stream}.
    }
    \label{fig:exm:spatial-until}
\end{figure*}

\begin{figure}
  \vspace{0.75em}
    \centering
    \begin{subfigure}{.65\linewidth}
    \includegraphics[width=1\linewidth]
    {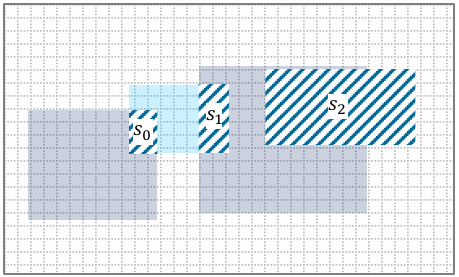}
    \end{subfigure}
    \caption{
    $s_2 = \Sval (p_2, \Data, 0, \tau, \revision{\epsilon})$,
    $s_0 = \Sval (p_2, \Data, 1, \tau, \revision{\epsilon}) \cap \Sval (p_1, \Data, 0, \tau, \revision{\epsilon})$,
    $s_1 = \Sval (p_2, \Data, 2, \tau, \revision{\epsilon}) \cap \Sval (p_1, \Data, 0, \tau, \revision{\epsilon}) \cap \Sval (p_1, \Data, 1, \tau, \revision{\epsilon})$,
    $\Sval( p_1 \Un^{s}_{[0,2]} p_2 , \Data, 0, \tau, \revision{\epsilon}) = s_0 \cup s_1 \cup s_2$.
    }
    \label{fig:exm:spatial-until-result}
\end{figure}

\begin{exmp}[Evaluating a simple spatio-temporal \revision{STPL} formula]
A sample data stream of three frames time stamped at times $t_0$, $t_1$, and $t_2$ is illustrated in Figure \ref{fig:exm:data-stream}.
We labeled rectangular geometric shapes in each frame by $\BB(1)$ and $\BB(2)$, 
and will refer to them by \revision{$p_1$} and \revision{$p_2$,} respectively.
Object $p_1$ does not change its geometric shape in all the frames, but $p_2$ evolves in each frame.
The location of the $p_1$ is the same in the first two frames, 
but in the third frame, it moves to the bottom-right corner of the frame.
The location of the $p_2$ changes constantly in each frame, 
that is, it first horizontally expands and moves to the bottom-left of the frame, 
and then expands horizontally and vertically and moves to the right-center of the frame. 
We are going to compute and evaluate the formula 
\[ \phi := \revision{\exists id_1. \exists id_2.} \Sexists \big ( \BB(id_1) \;\Un^s_{[0,2]}\; \BB(id_2) \big) \]
according to the semantics in Def. \ref{lbl:ext-temporal-topo-model} and Def. \ref{stql:smt}.
\revision{
An English statement equivalent to the above formula is: 
}

\revision{
``There exist two objects $\BB(id_1)$ and $\BB(id_2)$ in the first frame such that, $\BB(id_2)$ is non-empty in the first frame; or, for the next two frames, its intersection with $\BB(id_1)$ in the previous frames is non-empty (given that $\BB(id_1)$ does not change)''.
}

In Figure \ref{fig:exm:spatial-until}, we demonstrate the result of  evaluating the subformula $\BB(id_1) \;\Un^s_{[0,2]}\; \BB(id_2)$ for $i=1$ to $3$, \revision{where we assign to $id_1 \leftarrow 1$ and $id_2 \leftarrow 2$ from the four possible assignments}.
We used $p_1$ and $p_2$ to refer to $\BB(id_1 \leftarrow 1)$ and $\BB(id_2 \leftarrow 2)$ in each frame, respectively.
\revision{That is, $p_1$ at $t=2$ is equivalent to $\sval(\epsilon(1),2)$.}
The evaluation of STPL formula $\phi$ starts with the following equation 
\begin{align*}
	&\dle \Sexists\; \BB(id_1) \;\Un^s_{[0,2]}\; \BB(id_2) \dri (\Data,0,\epsilon_0,\zeta_0, \revision{\tau}) := 
	\\
	&\;\;\;\;
	\Sval(\Sexists\; p_1 \;\Un^s_{[0,2]}\; p_2 , \Data, 0, \tau, \revision{\epsilon, \zeta})
\end{align*}
where, $\epsilon_0[id_1 \leftarrow 1,id_2 \leftarrow 2]$ and $\zeta_0[id_1 \leftarrow \bot,id_2 \leftarrow \bot]$.

The evaluating of the until subformula is computed as below
\begin{align*}
    &\Sval( p_1 \;\Un^{s}_{[0,2]}\; p_2 , \Data, 0, \tau, \revision{\epsilon, \zeta}) :=
        \displaybreak[2] 
    \\
    &\bigcup_{t' \in \{0,1,2\} } 
    \Big(\Sval(p_2, \Data, t', \tau, \revision{\epsilon, \zeta}) \cap
    \\
    &\bigcap_{0 \leq t'' <t'} \Sval(p_1, \Data, t'', \tau, \revision{\epsilon, \zeta})
    \Big),
\end{align*}
where $\tau(0)=0, \tau(1)=1,$ and $\tau(2)=2$.. 
The above formula is equal to the below disjunctive normal formula as shown in Fig. \ref{fig:exm:spatial-until}
\begin{gather*}
    \Sval(p_2, \Data, 0, \tau, \revision{\epsilon, \zeta}) 
    \\
    \cup
    \displaybreak[2] 
    \\
    \big(\Sval(p_2, \Data, 1, \tau, \revision{\epsilon, \zeta}) 
    \cap \;
    \Sval(p_1, \Data, 0, \tau, \revision{\epsilon, \zeta})\big)
    \\
    \cup
    \displaybreak[2] 
    \\
    \big(\Sval(p_2, \Data, 2, \tau, \revision{\epsilon, \zeta}) 
    \cap \;
    \Sval(p_1, \Data, 0, \tau, \revision{\epsilon, \zeta}) 
    \; \cap \;
    \\ \;\;\;\;\;\;\;\;\;\;\;\;\;\;\;\;\;\;\;\;\;\;\;\;\;\;\;\;\;\;\;\;\;\;\;\;\;\;\;\;\;\;\;\;\;\;\;\;\;\;\;\;\;\;\;\;\;\;\;\;
    \Sval(p_1, \Data, 1, \tau, \revision{\epsilon, \zeta})\big).
\end{gather*}
\end{exmp}
The final evaluation of the above formula is depicted as hashed regions $s_0$, $s_1$ and $s_2$ in Fig. \ref{fig:exm:spatial-until-result}.
The union of the regions is a non-empty set, hence, the spatial existential operator returns \textit{true}.


\subsection{MTL/STL Equivalences in STPL}
Below, we represent some rewriting rules to translate time interval based formulas in MTL/STL to their equivalent STPL formulas. 
\begin{align*}
    \phi_1 \;\Un_{[a,b]}\; \phi_2 \equiv& \;
    x. \phi_1 \Un ( (a \leq  \Ctime   - x \leq b) \wedge \phi_2 )
	\displaybreak[2] 
    \\
    \phi_1 \;\Rc_{[a,b]}\; \phi_2 \equiv& \;
    x. \phi_1 \Rc ( (a \leq  \Ctime   - x \leq b) \implies \phi_2 )
	\displaybreak[2] 
    \\
    \Diamond_{[a,b]}\; \phi \equiv& \;
    x. \top \Un ( (a \leq  \Ctime   - x \leq b) \wedge \phi )
	\displaybreak[2] 
    \\
    \Diamond_{[a,b]}\; \phi \equiv& \;
    x. \Diamond ( (a \leq  \Ctime   - x \leq b) \wedge \phi )
	\displaybreak[2] 
    \\
    \Box_{[a,b]}\; \phi \equiv& \;
    x. \bot \Rc ( (a \leq  \Ctime   - x \leq b) \implies \phi )    
	\displaybreak[2] 
    \\
    \Box_{[a,b]}\; \phi \equiv& \;
    x. \Box( (a \leq  \Ctime   - x \leq b) \implies \phi )    
\end{align*}

In the next section, we use specific examples of perception systems to explain each aspect of the logic individually gradually and later combined.

\section{Case Study}\label{sec:case-study}


In this section, we are going to gradually demonstrate the expressivity of the STPL language using a sample perception data stream corresponding to the image frames in Figure \ref{fig:kitti:case-study}.
There are six frames in this case-study that are taken from the KITTI\footnote{\url{http://www.cvlibs.net/datasets/kitti/}} databases (\cite{geiger2013vision}).
We adopt a perception system based on SqueezeDet by which some desirable classes of objects in image frames are recognized.
The perception classifies an object as one class from 
 three desirable classes \textit{Car}, \textit{Cyclist}, and \textit{Pedestrian}.
It also assigns to each detected object a confidence level (normalized between 0 and 1), and
a bounding box that surrounds the object.
We also manually associate a unique object identifier to the detected objects of each frame.
The data is available to the object data stream function $\Data$ as represented in Table \ref{tbl:kitti:case-study-data}, and visualized in the frames in Fig. \ref{fig:kitti:case-study}.

In the following sections, we focus on STPL specifications that  use  quantifiers, and spatial and temporal operators.
\revision{
We present the formalization of 15 requirements in total.
A highlight is the STPL formula (\ref{ex:4:b}) that formalizes Req. \ref{req:obj-occlusion} and its utilization on the KITTI dataset in our monitoring tool.
Our monitoring tool was able to detect some mismatches in the labeled data. 
All of the STPL formulas and their monitoring result except the last two are distributed with our monitoring framework (\cite{stpl_tools}).
}

Note that in the rest of the sections, when we translate a requirement into STPL, we will refer to the following assumptions.
These assumptions enable us to write smaller formulas.
They can be relaxed at the expense of more complex requirements.
\begin{assum}\label{assum:dif-modules}
There are different perception modules to detect, track and classify the objects. That is, the object detector detects objects, and the object tracker assigns unique identifiers to the detected objects, and the classifier assigns classes to the detected objects.
\end{assum}
\begin{assum}\label{assum:always-detect}
Object detector always detects objects. 
\end{assum}
\begin{assum}\label{assum:unique-id}
Each detected and tracked object has a unique id. 
\end{assum}
The tracker can check the detected objects through the sequence of frames but it does not imply that the classifier will assign the same class to them in all the frames.
\revision{
If we cannot assume the existence of unique IDs on the same object, then we can still write requirements in a more complicated form to achieve more conservative results.
For example, in the requirements that follow (Req \ref{req:unique:obj}-\ref{req:point-cloud-occlusion}), whenever we check for the presence of the same ID, e.g., 
\[
\psi_= = \Box \forall Id_i. \Diamond \exists Id_j . (\;Id_i = Id_j \implies \varphi),
\]
then this precondition can be replaced with one where we quantify over all objects that intersect with the original object in the previous frame, e.g.,
\begin{align*}
\psi_\sqcap = & \Box \forall Id_i @x. \Diamond \exists Id_j . 
\\ 
& \Big(\big( \Sexists (\BB(Id_i) \sqcap \BB(Id_j)) \wedge C(Id_i) = C(Id_j)\big) \implies \varphi \Big).
\end{align*}
Notice the increase in complexity between the two formulas.
In $\psi_=$, we just need to store the ID of the object in $Id_i$ and just check if in the future there is another object ($Id_j$) with the same ID.
In $\psi_\sqcap$, we need to store the time $x$ that we observed the object $Id_i$, so that we can retrieve its bounded box and check for intersection with future objects along with checking the agreement of classes.
Clearly, $\psi_=$ and $\psi_\sqcap$ are not syntactically or semantically equivalent.
Nevertheless, under the assumption of sufficient sampling rate, then the objects that satisfy $\psi_=$ will also satisfy $\psi_\sqcap$.
There exist other possible formulas under which we can relax the requirement for persistent unique IDs.
However, in the following presentation, we will always be using the assumptions that result in the simplest STPL formulas.
}


\hlRev{For the requirements that are formalized into STPL in the rest of this section, we used our STPL monitoring tool to apply the data stream in Table \ref{tbl:kitti:case-study-data} to each STPL formula, and presented the result in Table \ref{tbl:kitti:stpl-result}.}

\begin{table}
\begin{center}
\caption{\hlRev{
The result of monitoring STPL formula $\varphi$ on the data stream $\Data$ in Table \ref{tbl:kitti:case-study-data}.
$\dle \varphi \dri$ abbreviates  $\dle \varphi \dri(\Data$,$0$,$\epsilon_0$,$\zeta_0$,$\revision{\tau}$$)$. \\
* We used the dataset ``0018'' from KITTI tracking benchmark for evaluating formula in Eq.(\ref{ex:4:b}). 
}}\label{tbl:kitti:stpl-result}
\resizebox{.48\textwidth}{!}{
\renewcommand{\arraystretch}{1.3}
\begin{tabular}{ |c|c|c|c|c|c|c|c| } 
\hline
$\boldsymbol{\varphi}$ & $\boldsymbol{\dle \varphi \dri}$ & $\boldsymbol{\varphi}$ & $\boldsymbol{\dle \varphi \dri}$ & $\boldsymbol{\varphi}$ & $\boldsymbol{\dle \varphi \dri}$ & $\boldsymbol{\varphi}$ & $\boldsymbol{\dle \varphi \dri}$\\
\Xhline{3\arrayrulewidth}
Eq.(\ref{ex:cs:1}) & $\top$ & Eq.(\ref{ex:cs:2}) & $\bot$ & Eq.(\ref{ex:cs:2:alt}) & $\bot$ & Eq.(\ref{ex:cs:3}) & $\bot$ \\[0.1cm]
\hline
Eq.(\ref{ex:cs:4}) & $\bot$ & Eq.(\ref{lbl:ex:6}) & $\top$ & Eq.(\ref{lbl:ex:7}) & $\top$ & Eq.(\ref{lbl:ex:8}) & $\bot$ \\[0.1cm]
\hline
Eq.(\ref{ex:1:a}) & $\top$ & Eq.(\ref{ex:2:a}) & $\bot$ & Eq.(\ref{ex:3:a}) & $\bot$ & Eq.(\ref{ex:3:b}) & $\bot$ \\[0.1cm]
\hline
Eq.(\ref{ex:4:b}) & $\top$ & *Eq.(\ref{ex:4:b}) & $\bot$ & Eq.(\ref{ex:4:c}) & $\top$ & Eq.(\ref{ex:4:d}) & $\top$ \\[0.1cm]
\Xhline{3\arrayrulewidth}
\end{tabular}
}
\end{center}
\end{table}

\subsection{Object Quantifier Examples}
First, we present a requirement that enables search through a data stream to find a frame in which there are at least two unique objects from the same class.

\begin{requirement}
\label{req:unique:obj}
\textit{There is at least one frame in which at least two unique objects are from the same class.}
\end{requirement}
\noindent \textbf{STPL} \hlRev{(using all the assumptions)}:
\begin{gather}\label{ex:cs:1}
    \phi_{\ref{req:unique:obj}} = \Diamond \exists Id_1.\exists Id_2. 
    \big( Id_1 \neq Id_2 \wedge C(Id_1) = C(Id_2)\big)
\end{gather}
In the above formula, the object variable constraint requires a unique assignment of object identifiers to the variables $Id_1$ and $Id_2$.
Additionally, the equivalence proposition is valid only if the classes of a unique pair of objects are equal.
Also, the existential quantifier requires one assignment of objects to the object variables that satisfy its following subformula.
Lastly, the eventually operator requires one frame in which its following subformula is satisfiable.

\paragraph{Checking the Formula on Real Data:}
From the data stream $\Data$ in Table \ref{tbl:kitti:case-study-data}, and as depicted in the frames in Fig. \ref{fig:kitti:case-study}, 
the frame numbers 0, 2 and 3
have two pedestrians, and frame number 3 has two cars in them.
Therefore, the formula is satisfiable for the given object data stream and we can denote it as $(\Data,0,\epsilon_0,\zeta_0, \revision{\tau}) \models \phi_{\ref{req:unique:obj}}$.
Note that we can push the $\Diamond$ operator after the existential quantifier (i.e., $\Diamond \exists Id_1.\exists Id_2.\phi \equiv \exists Id_1.\exists Id_2.\Diamond \phi$)
and still expect the same result.

\subsection{Examples with Time Constraints}

Next, we formalize a requirement in which we should use nested quantifiers and time/frame constraints in our formalization. 

\begin{requirement}
\label{req:prob:fps}
\revision{
\textit{Always, for all objects in each frame, their class probability must not decrease with a factor greater than 0.1 in the next two seconds, unless Frame Per Second (FPS) drops below 10 during the first second.}
}
\end{requirement}

\revision{
\noindent \textbf{STPL} (using Assumptions \ref{assum:dif-modules}-\ref{assum:unique-id}):
\begin{align}
\phi_{\ref{req:prob:fps}} = & \Box \forall Id_1 @x.  \big( \psi_{\ref{req:prob:fps}1}^{Id_1,x} \implies \psi_{\ref{req:prob:fps}2}^{x} \big)     \label{ex:cs:2}
\end{align}   
where
\begin{align}
\psi_{\ref{req:prob:fps}1}^{Id_1,x} \equiv & \Diamond \exists Id_2. \big ( Id_1 = Id_2 \; \wedge \; \Ctime - x \leq 2 \; \wedge 
            \nonumber \\
           &  \qquad\qquad P(Id_2) < 0.9 \times P(Id_1) \big )
        \nonumber \\
\psi_{\ref{req:prob:fps}2}^{x} \equiv & \Next \Diamond \big( Ratio(\Cframe - x,\Ctime - x) < 10  \wedge \Ctime - x \leq 1 \big) \nonumber
\end{align}   
}

\revision{
Notice that the function ``$Ratio(\Cframe - x,\Ctime - x)$" is not directly allowed by the syntax of STPL (Def. \ref{lbl:stql-brief-syntax}).
However, the arithmetic expression uses a single freeze time variable $x$ and, hence, its evaluation is no more computationally expensive than of the individual expressions ``$(\Cframe - x)$" and ``$(\Ctime - x)$".
We remark that the function \textit{Ratio} introduces the possibility of division by zero. 
In the formula $\psi_{\ref{req:prob:fps}2}^{x}$, the issue is avoided in the specification by using the next time operator ($\Next$). 
However, an implementation of the function \textit{Ratio} should handle the possibility of division by zero.
The subformula $\psi_{\ref{req:prob:fps}1}^{Id_1,x}$ searches for a vehicle for which within 2 sec in the future, i.e., ($\Ctime - x \leq 2$), its classification probability drops below the desired threshold. 
Notice that with the expression $P(Id_2) < 0.9 \times P(Id_1)$, we compare probabilities for the same object across different points in time.
The subformula $\psi_{\ref{req:prob:fps}2}^{x}$ checks whether within 1 sec in the future the FPS drops below 10.
The implication $\psi_{\ref{req:prob:fps}1}^{Id_1,x} \implies \psi_{\ref{req:prob:fps}2}^{x}$ enforces that if there is a vehicle for which the classification probability drops, then at the same time the FPS should be low.
}

\revision{
\paragraph{Checking the Formula on Real Data:}
Except for the object with $ID=2$ in the first frame in Table \ref{tbl:kitti:case-study-data}, the rest of the objects satisfy the requirement. Therefore, the whole requirement is not satisfiable, that is $(\Data,0,\epsilon_0,\zeta_0, \tau) \not \models \phi$. 
Formula (\ref{ex:cs:2}) is too strict.
That is, it checks the maximum allowed drop in probabilities for a classified object in all the frames rather than for the newly classified objects.
For example, assume a sample data stream of size three with only one object classified as a car with the probabilities $0.7$, $0.9$, $0.64$ during the first, second, and third frames, respectively.
Formula (\ref{ex:cs:2}) without the always operator is not satisfiable starting from the second frame (more than $28\%$ probability decrement), but satisfiable for the first and the last frame. 
}

\revision{
We can relax  Formula (\ref{ex:cs:2})  by adding a precondition to the antecedent of the implication to only consider the newly detected objects.
The relaxed requirement is possible by using the \textit{weak previous operator} $\Previous_{w}$.
The weak previous operator does not become unsatisfiable in the first frame on a data stream where there is no previous frame.
Similarly, by $\Next_{w}$, we denote the \textit{weak next operator}.
It is similar to the next operator, but it does not become unsatisfiable in the last frame on a finite data stream.
For more information about weak temporal operators and past LTL see the works by \cite{eisner2006weak} and \cite{cimatti2004bounded}, respectively.
}

The following formula is a revised version of Formula (\ref{ex:cs:2}):

\noindent \textbf{STPL} (using Assumptions \ref{assum:dif-modules}-\ref{assum:unique-id}):
\revision{
\begin{equation}
    \phi_{\ref{req:prob:fps}}' = \Box \forall Id_1 @x.  \Big( \psi_{\ref{req:prob:fps}0}^{Id_1} \Rightarrow \big(\psi_{\ref{req:prob:fps}1}^{Id_1,x} \Rightarrow \psi_{\ref{req:prob:fps}2}^{x} \big) \Big )
        \label{ex:cs:2:alt} 
\end{equation}
where
\[
   \psi_{\ref{req:prob:fps}0}^{Id_1}  \equiv \Previous_{w} \forall Id_3. (Id_1 \neq Id_3) 
\]
}

In formula $\phi_{\ref{req:prob:fps}}'$, the antecedent subformula $ \psi_{\ref{req:prob:fps}0}^{Id_1}$ checks if an object did not exist in the previous frame.
Therefore, the consequent $\psi_{\ref{req:prob:fps}1}^{Id_1,x} \implies \psi_{\ref{req:prob:fps}2}^{x}$ is only checked  for new objects that appear in the frame for the first time.


\begin{requirement}
\label{req:time:constraints}
\textit{\hlRev{Always, each object in a frame must exist in the next 2 frames during 1 second.}}
\end{requirement}
\noindent \textbf{STPL} \hlRev{(using Assumption \ref{assum:always-detect}-\ref{assum:unique-id}):}
\begin{equation}
\begin{array}{rl}
\label{ex:cs:3}
    \phi_{\ref{req:time:constraints}} = & \Box \forall Id_1 @x. \bigg(  \big(
    \Previous_{w} 
    \forall Id_3. (Id_1 \neq Id_3) \big) \Rightarrow 
    \\
    & \qquad \qquad  \Box \Big(
    \big(\Ctime - x \le 1 \; \wedge \; \Cframe - x \le 2 \big) 
    \Rightarrow
    \\
    & \qquad \qquad  \exists Id_2.
    \big( Id_1 = Id_2 \big)\Big)\bigg)
\end{array}
\end{equation}

Similar to the previous example, the equalities that need to hold are in the consequent of the second implication, but its antecedent is a conjunction of time and frame constraints.
In this formula, there is a freeze time variable after the first quantifier operator, which is followed by an always operator.
Thus, the constraints apply to the elapsed time and frames between the freeze time operator and the second always operator. 
Therefore, for any three consecutive frames, if the last two frames are within 1 second of the first frame, then all the objects in the first frame have to reappear in the second and third frames.

\paragraph{Running Formula in Real Data Stream:}
\hlRev{The same result as in the previous example holds here for the data stream $\Data$ in Table \ref{tbl:kitti:case-study-data} that is $(\Data,0,\epsilon_0,\zeta_0, \revision{\tau}) \not \models \phi_{\ref{req:time:constraints}}$.}

Below, we translate the requirement previously presented as in Req. \ref{lbl:req-intro-exm-1}.
\begin{requirement} 
\label{req:uber:formal}
{Whenever a new object is detected, then it is assigned a class within 1 sec, after which the object does not change class until it disappears.}
\end{requirement}
\noindent \textbf{STPL} \hlRev{(using Assumptions \ref{assum:dif-modules}-\ref{assum:unique-id}):}
\begin{equation}
\label{ex:cs:4}
\begin{array}{rl}
    \phi_{\ref{req:uber:formal}} =  & \Box \forall Id_1 @x. \\
    & \bigg( 
    \Big( C(Id_1) = 0 \Rightarrow \psi_{\ref{req:uber:formal}1}^{Id1,x} \Big) \;\wedge
    \\
    & \Big( C(Id_1) > 0 \Rightarrow \psi_{\ref{req:uber:formal}2}^{Id1,x} \Big) \bigg)
\end{array}
\end{equation}
where the subformulas are defined as:
\begin{align*}
\psi_{\ref{req:uber:formal}1}^{Id1,x} \equiv & 
    \Diamond \Big(
    \big( \Ctime - x \le 1 \; \wedge \; \Cframe -x \ge 1 \big)
    \wedge
    \\
    & \qquad \Box \exists Id_2. \big( Id_1 = Id_2 \wedge C(Id_2) > 0 \big)\Big)     
    \\
\psi_{\ref{req:uber:formal}2}^{Id1,x} \equiv 
    & \Box \forall Id_3 \Big(
    \big( \Cframe -x \ge 1
    \wedge Id_3 = Id_1 \big)
    \Rightarrow 
    \\
    & \qquad  \qquad  C(Id_1)=C(Id_3)     \Big)    
\end{align*}
In the above formula, we evaluate the two implication-form subformulas for any objects in all the frames.
If there is an object that is not classified (i.e., $C(Id_1)=0$), then we check the consequence of the corresponding subformula.
The subformula corresponding to the subformula $\psi_{\ref{req:uber:formal}1}^{Id1,x}$ requires that when an unclassified object was observed, then eventually in less than a second afterward, the object always has a class being assigned to it. 
If there is an object that is already classified, then the second predicate has to be evaluated.
The subformula equivalent to the $\psi_{\ref{req:uber:formal}2}^{Id1,x}$ requires that when a classified object was observed, then afterward the same object only can take the same class. 

\paragraph{Running Formula in Real Data Stream:}
\hlRev{In the frame $1$ in Table \ref{tbl:kitti:case-study-data}, the object with $ID=2$ changes its class from \textit{cyclist} to \textit{pedestrian}.
Therefore, the data stream $\Data$ does not satisfy the requirement $(\Data,0,\epsilon_0,\zeta_0, \revision{\tau}) \not \models \phi_{\ref{req:uber:formal}}$.}

\begin{table}
\begin{center}
\caption{
Data stream $\Data$ of image frames in Fig. \ref{fig:kitti:case-study}. Each row for the column headers: 
Frame, $\tau$, ID (object identifiers), class, and Prob represent 
the frame number, 
the sampling time (e.g., here we assume that frame-per-second is 25 fps), the associated identifier to the objects, 
the classification confidence, respectively.
Moreover, the other four headers are the data attributes (minimum and maximum lateral and longitudinal positions of coordinates of the bounding boxes) by which a bounding box is associated with each classified object. Note that some objects are not tracked correctly throughout the data stream.
}\label{tbl:kitti:case-study-data}
\resizebox{.48\textwidth}{!}{
\renewcommand{\arraystretch}{1.3}
\begin{tabular}{ |c|c|c|c|c|c|c|c|c| } 
\hline
\textbf{Frame} & $\boldsymbol{\tau}$ & \textbf{ID} & \textbf{Class} & \textbf{Prob} & $\boldsymbol{x_{min}}$ & $\boldsymbol{y_{min}}$ & $\boldsymbol{x_{max}}$ & $\boldsymbol{y_{max}}$\\
\Xhline{3\arrayrulewidth}
0 & 0 & 1 & \textit{car} & 0.88 & 58 & 151 & 220 & 287 \\[0.1cm]
\hline
0 & 0 & 2 & \textit{cyclist} & 0.75 & 479 & 124 & 690 & 382 \\[0.1cm]
\hline
0 & 0 & 3 & \textit{pedestrian} & 0.63 & 522 & 130 & 632 & 377 \\[0.1cm]
\hline
0 & 0 & 4 & \textit{pedestrian} & 0.64 & 861 & 133 & 954 & 329 \\[0.1cm]
\Xhline{3\arrayrulewidth}
1 & 0.04 & 1 & \textit{car} & 0.88 & 61 & 152 & 217 & 283 \\[0.1cm]
\hline
1 & 0.04 & 2 & \textit{cyclist} & 0.57 & 493 & 111 & 699 & 383 \\[0.1cm]
\hline
1 & 0.04 & 3 & \textit{pedestrian} & 0.64 & 877 & 136 & 972 & 330 \\[0.1cm]
\Xhline{3\arrayrulewidth}
2 & 0.08 & 1 & \textit{car} & 0.89 & 58 & 143 & 220 & 271 \\[0.1cm]
\hline
2 & 0.08 & 2 & \textit{pedestrian} & 0.65 & 511 & 107 & 724 & 367 \\[0.1cm]
\hline
2 & 0.08 & 3 & \textit{pedestrian} & 0.64 & 911 & 115 & 1001 & 340 \\[0.1cm]
\Xhline{3\arrayrulewidth}
3 & 0.12 & 1 & \textit{car} & 0.92 & 56 & 139 & 216 & 266 \\[0.1cm]
\hline
3 & 0.12 & 2 & \textit{cyclist} & 0.59 & 493 & 111 & 705 & 380 \\[0.1cm]
\hline
3 & 0.12 & 3 & \textit{pedestrian} & 0.72 & 541 & 125 & 649 & 351 \\[0.1cm]
\hline
3 & 0.12 & 4 & \textit{car} & 0.58 & 926 & 107 & 1004 & 302 \\[0.1cm]
\hline
3 & 0.12 & 5 & \textit{pedestrian} & 0.76 & 938 & 118 & 998 & 332 \\[0.1cm]
\Xhline{3\arrayrulewidth}
4 & 0.16 & 1 & \textit{car} & 0.91 & 53 & 139 & 217 & 265 \\[0.1cm]
\hline
4 & 0.16 & 2 & \textit{pedestrian} & 0.80 & 551 & 126 & 658 & 356 \\[0.1cm]
\Xhline{3\arrayrulewidth}
5 & 0.2 & 1 & \textit{car} & 0.92 & 52 & 140 & 216 & 264 \\[0.1cm]
\hline
5 & 0.2 & 2 & \textit{cyclist} & 0.62 & 506 & 104 & 695 & 368 \\[0.1cm]
\hline
5 & 0.2 & 3 & \textit{pedestrian} & 0.68 & 552 & 115 & 669 & 362 \\[0.1cm]
\Xhline{3\arrayrulewidth}
\end{tabular}
}
\end{center}
\end{table}

\subsection{Examples of Space and Time Requirements with 2-D Images}

In the following examples, we want to demonstrate the expressivity and usability of the STPL language to capture requirements related to the detected objects and their spatial and temporal properties.

\subsubsection{Basic Spatial Examples without Quantifiers.}
\label{sec:exmp:spat:no:quant}
Let us assume that we have a spatial proposition $p$ as in Fig. \ref{fig:exmp:spatio-temp-ops}.

\begin{exmp}
\revision{
One way to formalize that the sampling rate is sufficient is to check if the bounding box of an object overlaps with itself across consecutive frames.
The following requirement can be used as a template in more complex requirements.
}
\begin{requirement}\label{req:topo-intersect}
\textit{\hlRev{The intersection of a spatial predicate $p$ with itself across all frames must not be empty.}}
\end{requirement}
\noindent\textbf{STPL: }
\begin{align}
\phi_{\ref{req:topo-intersect}}= \Sexists \; \Box^s p
\label{form:sexists:box}
\end{align}
 \revision{The above requirement is clearly too strict.
 In order to make the requirement more realistic, we introduce frame intervals (see Rem. \ref{rem:frame-int}) and also use quantification over objects.
 If we change the above requirement to:
 \textit{``All the objects in all the frames must intersect with themselves in the next three frames''}, 
 then we can modify Eq. (\ref{form:sexists:box}) to the following:}
 \begin{align}
\revision{\phi_{\ref{req:topo-intersect}}'= \Box \forall Id_1. \;\Sexists \; \Tilde{\Box}^s_{[0,3]} \;\BB(Id_1)}
\end{align}
\end{exmp}

\begin{exmp}
\revision{It is interesting to check the occupancy of an object across a series of frames.
This can be used for identifying regions of interest.}
\begin{requirement}\label{req:topo-union}
\textit{The union of a spatial proposition $p$ with itself in all the frames is not equal to the universe.}
\end{requirement}
\noindent\textbf{STPL: }
\begin{align}
\phi_{\ref{req:topo-union}} = \neg\; \Sforall \; \Diamond^s p  
\label{form:diamond}
\end{align}
The result of applying the formula $\Diamond^s p$ on a sequence of three frames is shown in Fig. \ref{fig:exmp:spatio-temp-ops}.
 The resulting set is not equal to the universe; 
 hence; the formula $\phi_{\ref{req:topo-union}}$ evaluates to \textit{true}.
\end{exmp}

\subsubsection{Basic Spatial Examples with Quantifiers.}
In the following examples, the spatial requirements cannot be formalized without the use of quantifiers.

\begin{exmp}
\revision{We can check if some properties hold for all the objects across all the frames.}
\begin{requirement}\label{req:in-bbx}
\textit{Always, all the objects must remain in the bounding box $(x_1,y_1,x_2,y_2)$.}
\end{requirement}
\noindent\textbf{STPL} \hlRev{(using Assumption \ref{assum:always-detect}):}
\begin{multline}\label{lbl:ex:6}
\phi_{\ref{req:in-bbx}} = \Box \forall Id. \big( LAT(Id,\LM) \ge x_1 \wedge  LAT(Id,\RM) \le x_2 \; \wedge 
\\
LON(Id,\TM) \ge y_1 \wedge  LON(Id,\BM) \le y_2 \big) 
\end{multline}
\end{exmp}

\paragraph{Running Formula in Real Data Stream:}
\hlRev{The result of evaluating the above formula depends on the size and position of the bounding box.
For example, if the bounding box is the same as the image frame in the data stream $\Data$ in Table \ref{tbl:kitti:case-study-data}, then we have $(\Data,0,\epsilon_0,\zeta_0, \revision{\tau}) \models \phi_{\ref{req:in-bbx}}$.}

\begin{exmp}
\revision{The combination of eventually/globally operators and a quantifier operator 
is very useful for specifying properties of an object across time.}
\begin{requirement}\label{req:evn-shift}
\textit{\hlRev{Eventually, there must exist an object that at the next frame shifts to the right.}}
\end{requirement}
\noindent\textbf{STPL} \hlRev{(using Assumptions \ref{assum:always-detect}-\ref{assum:unique-id}):}
\begin{multline}\label{lbl:ex:7}
\phi_{\ref{req:evn-shift}} = \Diamond \exists Id_1@x. \Next \exists Id_2. \big(
\\
Id_1 = Id_2 \wedge
LAT(Id_1,\LM) <  LAT( Id_2,\LM)
\big)  
\end{multline}
Note that in the above formula, the $Id_1$ is quantified in a freeze time quantifier to get access to the data-object values in the frozen time. 
\hlRev{If we change the first existential quantifier to the universal quantifier, then the formula represents the following requirement: ``Eventually, there must exist a frame in which all the objects shift to the right in the next frame''.}
\end{exmp}

\paragraph{Running Formula in Real Data Stream:}
\hlRev{The data stream $\Data$ as in Table \ref{tbl:kitti:case-study-data} satisfies the above formula $(\Data,0,\epsilon_0,\zeta_0, \revision{\tau}) \models \phi_{\ref{req:evn-shift}}$.
For example, the pedestrian with $ID=3$ moved to the right in the frame $1$.
Note that the pedestrian and the recording camera both moved to the left, while the camera's movement was faster.}

\begin{exmp}
\revision{The requirement below checks the robustness of the tracking algorithms in a perception system using time and spacial constraints.}
\revision{
\begin{requirement}
\label{req:disappear} 
\textit{If an object disappears from the right of the image in the next frame, then during the next 1 seconds, 
it can only reappear from the right.
}
\end{requirement}
}
\revision{
\noindent\textbf{STPL} (using Assumptions \ref{assum:always-detect}-\ref{assum:unique-id}):
\begin{equation}
\label{lbl:ex:8}
\phi_{\ref{req:disappear}} =
\Box \forall Id_1@x. \Next_w \forall Id_2@y. \Big (
\psi_{\ref{req:disappear}1}^{Id_1,Id_2} \Rightarrow \psi_{\ref{req:disappear}2}^{Id_1,Id_2,y}
\Big)
\end{equation}
where
\begin{align*}
\psi_{\ref{req:disappear}1}^{Id_1,Id_2} \equiv & \;Id_1 = Id_2 \wedge 
\\
& \;LAT(Id_1,\LM) <  LAT( Id_2,\LM) \;\wedge
\\
& \;LAT(Id_2,\LM) > \frac{3}{4}W \;\wedge
\\
& \;\Next_w \forall Id_3. \;(Id_3 \neq Id_1)
\\
 \psi_{\ref{req:disappear}2}^{Id_1,Id_2,y} \equiv & \;\Next_w \Box  
 \\
& \;\Big( \big ( (\Ctime - y < 1 ) \wedge \forall Id_4 .(Id_2 \neq Id_4) \;\wedge 
 \\
& \;\Next_w \exists Id_5.(Id_2 = Id_5 ) \big )
 \Rightarrow 
 \\
& \;\Next_w LAT(Id_5,\LM) > \frac{3}{4}W
 \Big)
\end{align*}
Note that in the above formula, the $Id_1$ and $Id_2$ are quantified in a nested freeze time quantifier structure, and hence, it is not an AAN formula.
There, $W$ is a constant denoting the width of the image in pixels.
This formalization will check the reappearance of an object even if it happened more than once.
In order to force the formula to only check once for the reappearance of an object, one should use the release operator. 
}
\end{exmp}

\paragraph{Checking the Formula on Real Data:}
\revision{
In Table \ref{tbl:kitti:case-study-data}, in the third frame, the pedestrian with $ID=3$ has disappeared in the fourth frame and reappeared in the last frame.
However, the preconditions captured in subformula $\psi_{\ref{req:disappear}1}^{Id_1,Id_2}$ for that object do not hold assuming that $W=1100$ for the data stream.
That is the least $x$ position of the bounding box for the object with $ID=3$ is $911$ in frame $2$, and then it changes to $541$ in frame $3$ which violates the condition for shifting toward right before disappearing (the $LAT$ condition in the subformula $\psi_{\ref{req:disappear}1}^{Id_1,Id_2}$).
Note that, from the second frame afterward, the tracking algorithm mixes the ID association to the objects, and as a result, the geometrical positions of the objects are not consistent.
The inconsistent object tracking makes the subformula $\psi_{\ref{req:disappear}1}^{Id_1,Id_2}$ not hold for viable cases.
Therefore, for the data stream $\Data$ in Table \ref{tbl:kitti:case-study-data}, we have $(\Data,0,\epsilon_0,\zeta_0,\tau) \models \phi_{\ref{req:disappear}}$.
}

\begin{figure*}[tbp]
  \vspace{0.75em}
\centering
\begin{subfigure}{0.48\linewidth}
    \centering
    \includegraphics[width=1\linewidth]{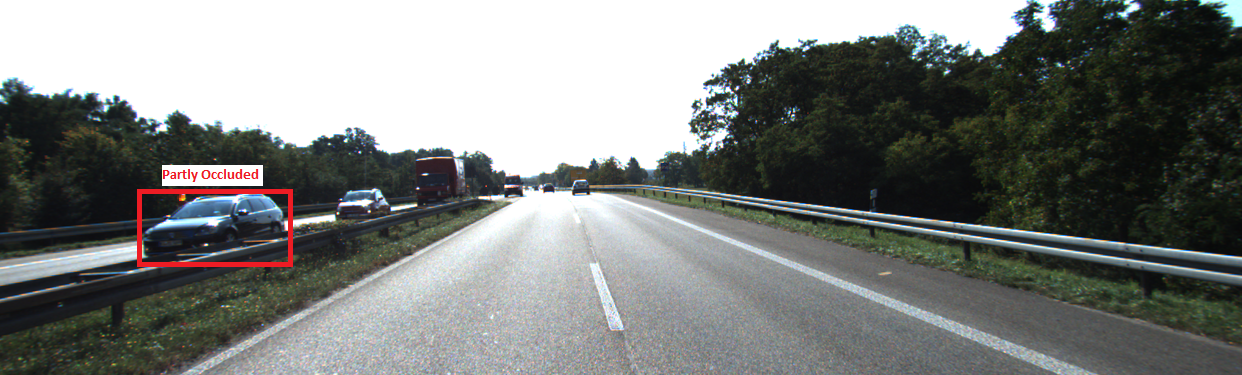}
    \caption{ The car inside the red rectangle is identified with ID $4$ in frame $10$, and it is annotated as ``\textit{partly occluded}''.}\label{fig:kitti-occ-1}
\end{subfigure}    
\begin{subfigure}{0.48\linewidth}
    \centering
    \includegraphics[width=1\linewidth]{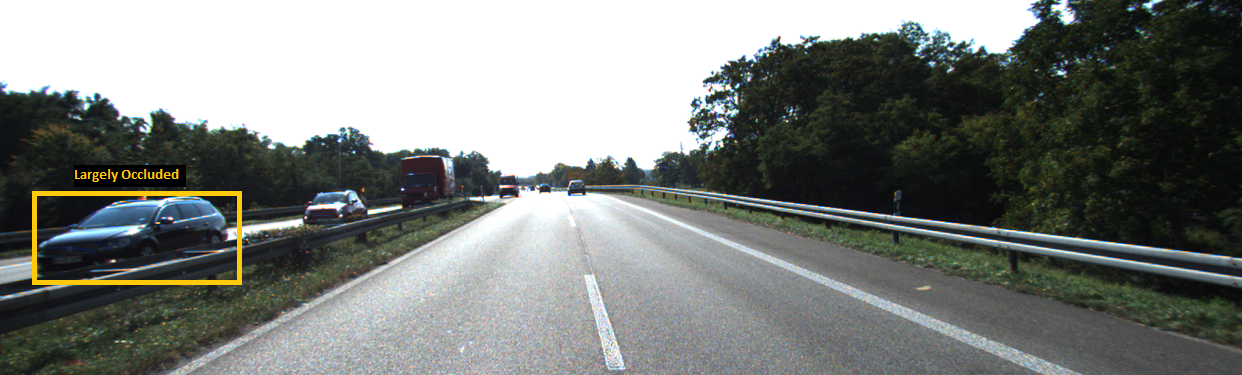}
    \caption{ The same car as in Image (a) is bounded by the yellow rectangle, and it is annotated as ``\textit{largely occluded}'' in frame $11$.}\label{fig:kitti-occ-2}
\end{subfigure}    
\begin{subfigure}{0.48\linewidth}
    \centering
    \includegraphics[width=1\linewidth]{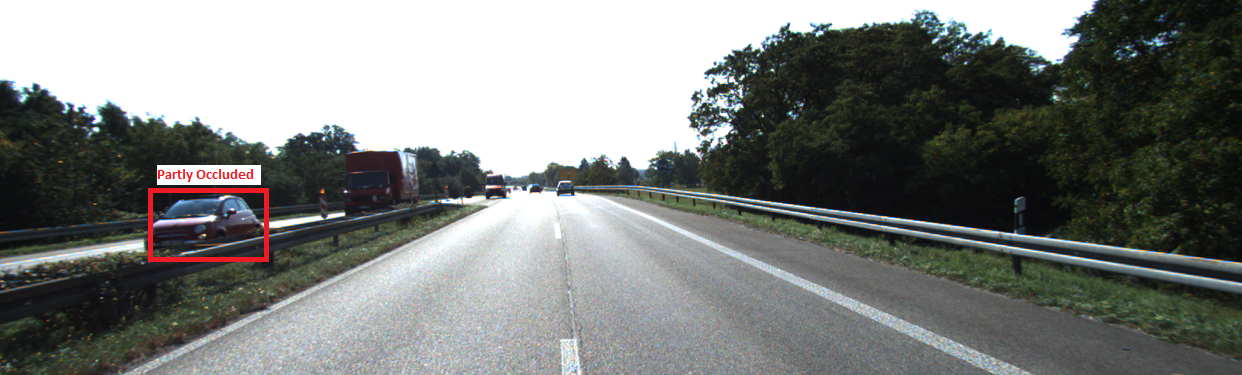}
    \caption{ The car inside the red rectangle is identified with ID $5$ in frame $14$, and it is annotated as ``\textit{partly occluded}''.}\label{fig:kitti-occ-3}
\end{subfigure}  
\begin{subfigure}{0.48\linewidth}
    \centering
    \includegraphics[width=1\linewidth]{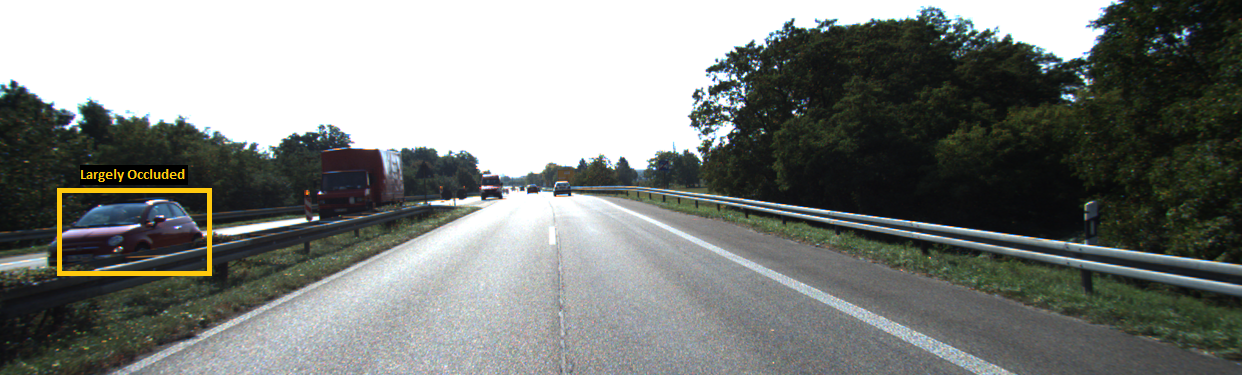}
    \caption{ The same car as in Image (c) is bounded by the yellow rectangle, and it is annotated as ``\textit{largely occluded}'' in frame $15$.}\label{fig:kitti-occ-4}
\end{subfigure}  
\hfill
\begin{subfigure}{0.48\linewidth}
    \centering
    \includegraphics[width=1\linewidth]{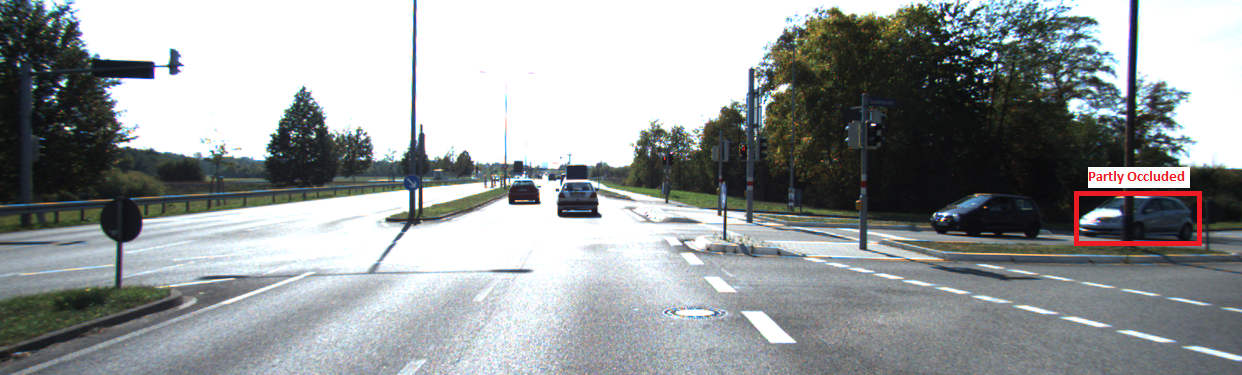}
    \caption{ The car inside the red rectangle is identified with ID $16$ in frame $260$, and it is annotated as ``\textit{partly occluded}''.}\label{fig:kitti-occ-5}
\end{subfigure}    
\begin{subfigure}{0.48\linewidth}
    \centering
    \includegraphics[width=1\linewidth]{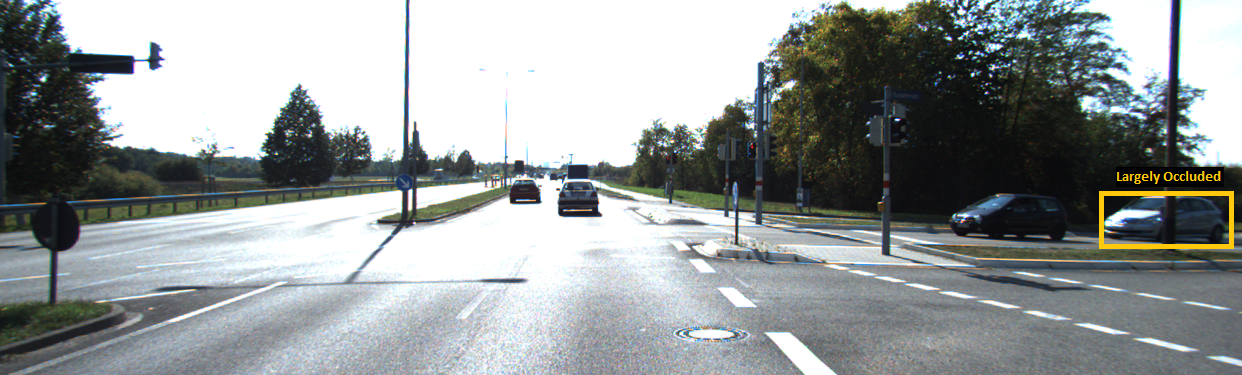}
    \caption{ The same car as in Image (e) is bounded by the yellow rectangle, and it is annotated as ``\textit{largely occluded}'' in frame $261$.}\label{fig:kitti-occ-6}
\end{subfigure}  
\caption{\hlRev{We used the labels corresponding to the 390 images in the folder ``$0008$'' from the KITTI tracking benchmark.
Our STPL monitoring tool detected frames 11, 15, and 261 with ``largely occluded'' labels as inconsistent annotations. 
The training data and their format are available from the links in the footnote.} \protect\footnotemark}
\label{fig:kitti-occ-example}
\end{figure*}

For the examples in the rest of this section, we focus on the real-time requirements that concern detected objects in perception systems.


\begin{exmp} \revision{The requirement below checks the robustness of the classification of the pedestrians in a perception system using time and spatial constraints.}
\begin{requirement}\label{req:rob-ped}
\textit{If a pedestrian is detected with probability higher than $0.8$ in the current frame, then for the next $1$ second, the probability associated with the pedestrian should not fall below $0.7$, and the bounding box associated with the pedestrian in the future should not overlap with another detected bounding box.}
\end{requirement}
\noindent\textbf{STPL} \hlRev{(using Assumptions \ref{assum:always-detect}-\ref{assum:unique-id}):}
\begin{multline}\label{ex:1:a}
    \phi_{\ref{req:rob-ped}} = \Box \forall Id_1@x. \Big( 
    \big( C(Id_1)=Ped \wedge P(Id_1) > 0.8 \big) \implies 
    \\
    \Box \Big( \Ctime - x \le 1 \implies 
    \\
    \exists Id_2. \big(Id_1 = Id_2 \wedge P(Id_2) > 0.7 \wedge C(Id_2)=Ped \;\wedge
    \\
    \forall Id_3. \big( Id_2 \neq Id_3 \implies 
    \\
    \neg\; \Sexists (\BB(Id_2) \sqcap \BB(Id_3))
    \big)\big)\Big)\Big)  
\end{multline}
\end{exmp}

\paragraph{Running Formula in Real Data Stream:}
\hlRev{In Table \ref{tbl:kitti:case-study-data}, there is no pedestrian associated with probability higher than $0.8$, therefore the data stream $\Data$ satisfies the above formula $(\Data,0,\epsilon_0,\zeta_0, \revision{\tau}) \models \phi_{\ref{req:rob-ped}}$.}

\begin{exmp}
The requirement below represents a situation in which all the adversarial cars in the frames from the current time forward are moving at least as fast as the ego car and in the same direction.

\begin{requirement}\label{req:bbx-not-expand}
\textit{Always all the bounding boxes of the cars in the image frames do not expand.}
\end{requirement}
\noindent\textbf{STPL} \hlRev{(using Assumptions \ref{assum:always-detect}-\ref{assum:unique-id}):}
\begin{multline}\label{ex:2:a}
    \phi_{\ref{req:bbx-not-expand}} = \Box \forall Id_1 @x. \Big( C(Id_1)=Car \implies 
    \\
    \Box \forall Id_2. \big( (Id_1 = Id_2 \wedge C(Id_2)=Car) \implies
    \\
    Area(Id_1) \ge Area(Id_2) \big)\Big)  
\end{multline}
Notice that in the above example, the geometric position of the objects in the space is not required. 
\end{exmp}

\paragraph{Running Formula in Real Data Stream:}
\hlRev{In Table \ref{tbl:kitti:case-study-data}, the bounding box of the car with $ID=1$ has expanded in the frame $2$.
Therefore, the data stream $\Data$ does not satisfy the above formula $(\Data,0,\epsilon_0,\zeta_0, \revision{\tau}) \not \models \phi_{\ref{req:bbx-not-expand}}$.}

\revision{
To ease the presentation in the formalization of the  requirements in the following examples, we introduce some derived spatial operators:
\begin{itemize}
    \item Subset: $p_1 \sqsubseteq p_2 \; \equiv \; \Sforall \big(\overline{p_1} \sqcup p_2 \big) $ 
    \item Set equality: $p_1 = p_2 \; \equiv \; p_1 \sqsubseteq p_2 \wedge p_2 \sqsubseteq p_1$
\end{itemize}
}

\begin{exmp}
We want to formalize a requirement specification that applies restrictions on the positions and movements of the adversarial and the ego car in all the image frames.
The following requirement can be thought of as a template to be used within other requirements.

\begin{requirement}\label{req:fix-rel-vel-pos}
\textit{The relative position and velocity of all the cars are fixed with respect to the ego car.}
\end{requirement}
\noindent\textbf{STPL} \hlRev{(using Assumptions \ref{assum:always-detect}-\ref{assum:unique-id}):}
\revision{
\begin{equation*}
\label{ex:3:a:mod}
    \phi_{\ref{req:fix-rel-vel-pos}} = \forall Id_1 @x. \Box \exists Id_2 . \Big( Id_1 = Id_2 \wedge \BB(Id_1) = \BB(Id_2)    \Big)
\end{equation*}
}
\revision{
Notice that we quantify over all objects in the first frame ($Id_1$), and then we require that for all future times there is an object ($Id_2$) with the same ID and bounding box.
If in addition, we would like to impose the requirement on any new objects that we observe, then the formalization of the requirement would be: 
}
\revision{
\begin{equation}
\label{ex:3:a}
\begin{array}{rl}
    \phi_{\ref{req:fix-rel-vel-pos}}' = & \Box \forall Id_1 @x. \Box \exists Id_2 . \\
    & \qquad \qquad \Big( Id_1 = Id_2 \wedge \BB(Id_1) = \BB(Id_2)    \Big)   
\end{array}
\end{equation}
}
Another formalization for the last requirements is 
\revision{
\begin{equation}\label{ex:3:b}
    \phi_{\ref{req:fix-rel-vel-pos}}'' = \Box \forall Id_1. \Big( \Box^s \BB(Id_1) = \Diamond^s \BB(Id_1) \Big)
\end{equation}
}
\end{exmp}

\paragraph{Running Formula in Real Data Stream:}
\hlRev{All of the objects in Table \ref{tbl:kitti:case-study-data} have different bounding boxes in different frames.
Thus, the data stream $\Data$ does not satisfy the above formulas $(\Data,0,\epsilon_0,\zeta_0, \revision{\tau}) \not \models \phi_{\ref{req:fix-rel-vel-pos}}$.}

\begin{exmp}
In this example, we represent a sample requirement for detecting occluded objects.
\hlRev{The example is about a high confidence OBJECT which suddenly disappears without getting close to the borders. 
The requirement checks if such an object existed before and next disappeared, then it has to be occluded by a previously close proximity object.}
\begin{requirement}\label{req:obj-occlusion}
\textit{\hlRev{If there exists a high confidence OBJECT, and in the next frame, it suddenly disappears without being close to the borders, then it must be occluded by another object.}}
\end{requirement}
\noindent\textbf{STPL} \hlRev{(using Assumptions \ref{assum:always-detect}-\ref{assum:unique-id}):}
\begin{multline}\label{ex:4:b}
    \phi_{\ref{req:obj-occlusion}} = \Box \forall Id_1@x. \Big( \big( \varphi_{high}^{prob} \wedge \varphi_{far}^{borders} \wedge \hlRev{\Next} \varphi_{disap}\big) \implies
    \\
    \varphi_{occ}
    \Big)  
\end{multline}
and the predicates are defined as:
\begin{align*}
    &\varphi_{high}^{prob} \equiv 
    P(Id_1) > 0.8 
    \\
    &\varphi_{far}^{borders} \equiv LON(Id_1,\TM) > d_1 \wedge LON(Id_1,\BM) < d_2 
    \\
    &\;\;\;\;\;\;\;\;\;\;\;\;\;\;
    \wedge LAT(Id_1,\LM) > d_3 \wedge LAT(Id_1,\RM) < d_4 
    \\
    &\varphi_{disap} \equiv 
        \hlRev{\forall Id_2.(Id_1 \neq Id_2)}
\end{align*}
\vspace{-20pt}
\begin{multline*}
    \varphi_{occ} \equiv 
    \exists Id_3. \exists Id_4.
    \bigg(
    Id_1 \neq Id_3 \wedge Id_1 = Id_4 \;\wedge \\
    \revision{\Sexists \Big( \BB(Id_4) \sqcap}
    \revision{\big(\BB(Id_3) \sqcup \Next^s \BB(Id_3)\big)} 
    \Big) 
    \bigg)
\end{multline*}
\hlRev{In the above formulas, the subformula $\varphi_{far}^{borders}$ checks if the object identified by $Id_1$ is close to the borders of the image frame at the current time, \revision{where $d_1$, $d_2$, $d_3$, and $d_4$ are some integers}. 
The subformula $\forall Id_2.(Id_1 != Id_2)$ is to check if an object disappeared.
In the subformula $\varphi_{occ}$, the $Id_4$ refers to the object which disappears in the next frame, and the $Id_3$ refers to the object that occludes the other object.
The spatial subformulas check if the bounding box of the two objects intersect each other in the current frame, or one bounding box in the current frame intersects with the other bounding box in the next frame. 
}
Similarly, we can formalize the occlusion without using STE operators by rewriting $\varphi_{occ}$ as below
\begin{multline}\label{ex:4:c}
    \varphi_{occ} \equiv 
    \exists Id_3. \exists Id_4.
    \big( Id_1 \neq Id_3 \wedge Id_1 = Id_4 \;\wedge 
    \\
     Dist(Id_4,\CT,Id_3,\CT) < d_5 \big ) 
\end{multline}
Note that the second formalization is less realistic because it puts a threshold on the \textit{Euclidean distance} \revision{(i.e., $d_5$ is a positive real number)} between two objects to infer their overlap.
\end{exmp}

\paragraph{Running Formula in Real Data Stream:}
\hlRev{All the objects in Table \ref{tbl:kitti:case-study-data} are close to the borders of the images, and the data stream does not satisfy the subformula $\varphi_{far}^{borders}$.
Therefore, the data stream $\Data$ satisfies the the above formula $(\Data,0,\epsilon_0,\zeta_0, \revision{\tau}) \models \phi$.}

\hlRev{In the following, we used the above formula to partially validate the correctness of a training data stream that is used for object tracking.}

\paragraph{Running Formula in Real Data Stream:}
\hlRev{We used the formula in Eq. (\ref{ex:4:b}) to verify the correctness of the labeled objects as ``\textit{largely occluded}'' in the KITTI tracking benchmark.
From the 21 datasets, except for 4 of them (i.e., datasets ``0013'', ``0016'', ``0017'', and ``0018''), we detected inconsistencies in labeling objects as ``\textit{largely occluded}''.
For example, in the dataset ``0008'' (to refer to as data stream $\Data$), we identified 3 frames in each a car was labeled as ``\textit{largely occluded}'', while they were inconsistent with the rest of the occluded labels.
That is, we have $(\Data,0,\epsilon_0,\zeta_0, \revision{\tau}) \not \models \phi$.
Our STPL monitoring tool detected the wrong labels by applying the STPL formula in Eq. (\ref{ex:4:b}) on the data stream in less than 2 seconds.
The result of this experiment is shown in the Figure \ref{fig:kitti-occ-example}.
We provide all the datasets as part of our monitoring tool.}

\footnotetext{Training data: \url{http://www.cvlibs.net/datasets/kitti/eval_tracking.php}, and 
data format: \url{https://github.com/JonathonLuiten/TrackEval/blob/master/docs/KITTI-format.txt}
}

Below, we translate the requirement previously presented as in Req. \ref{lbl:req-intro-exm-2}.
\begin{requirement}\label{self-overlap}
The frames per second of the camera is high enough so that for all detected cars, 
their bounding boxes self-overlap for at least 3 frames and for at least 10\% of the area.
\end{requirement}
We interpret the above requirement to require an overlap check over a batch of three frames 
rather than three consecutive frames. 

\noindent\textbf{STPL} \hlRev{(using Assumptions \ref{assum:always-detect}-\ref{assum:unique-id}):}
\begin{multline}\label{ex:4:d}
    \phi_{\ref{self-overlap}} = \Box \forall Id_1 @x. \bigg( 
    \hlRev{\big( 
    \Previous_{w} 
    \forall Id_3. (Id_1 \neq Id_3) \big) \implies}
    \\
    \Box \Big(
    \big(  \Cframe -x \ge 1  \; \wedge \; \Cframe -x \le 3 \big)
    \implies
    \\
    \forall Id_2.
    \big( Id_1 = Id_2 \;\implies 
    \\
    Ratio(Area(\BB(Id_1) \sqcap \BB(Id_2)), 
    \\
    Area(\BB(Id_2))) \ge 0.1 
    \big) \Big) \bigg)
\end{multline}
In the above formula, the spatial subformula is in the form of a non-equality \textit{ratio} function.
Its first parameter represents an occupied area for the intersection of the bounding boxes of any two objects $obj_1$ and $obj_2$ referred to by $Id_1$ and $Id_2$, respectively.
The second parameter denotes the area for $obj_2$.
For the non-equality to be satisfiable, first, there must be a non-empty intersection between the two objects.
Second, the ratio of the intersected area to the area of the second object must be at least $10\%$. 

\paragraph{Running Formula in Real Data Stream:}
\hlRev{The data stream $\Data$ in Table \ref{tbl:kitti:case-study-data} satisfies the above formula $(\Data,0,\epsilon_0,\zeta_0, \revision{\tau}) \models \phi_{\ref{self-overlap}}$.}


\subsection{Examples of Space and Time Requirements with 3D environment}

\subsubsection{Missed classification scenario}

\begin{figure*}[tbp]
  \vspace{0.75em}
\centering
\begin{subfigure}{0.48\linewidth}
    \centering
    \includegraphics[width=1\linewidth]{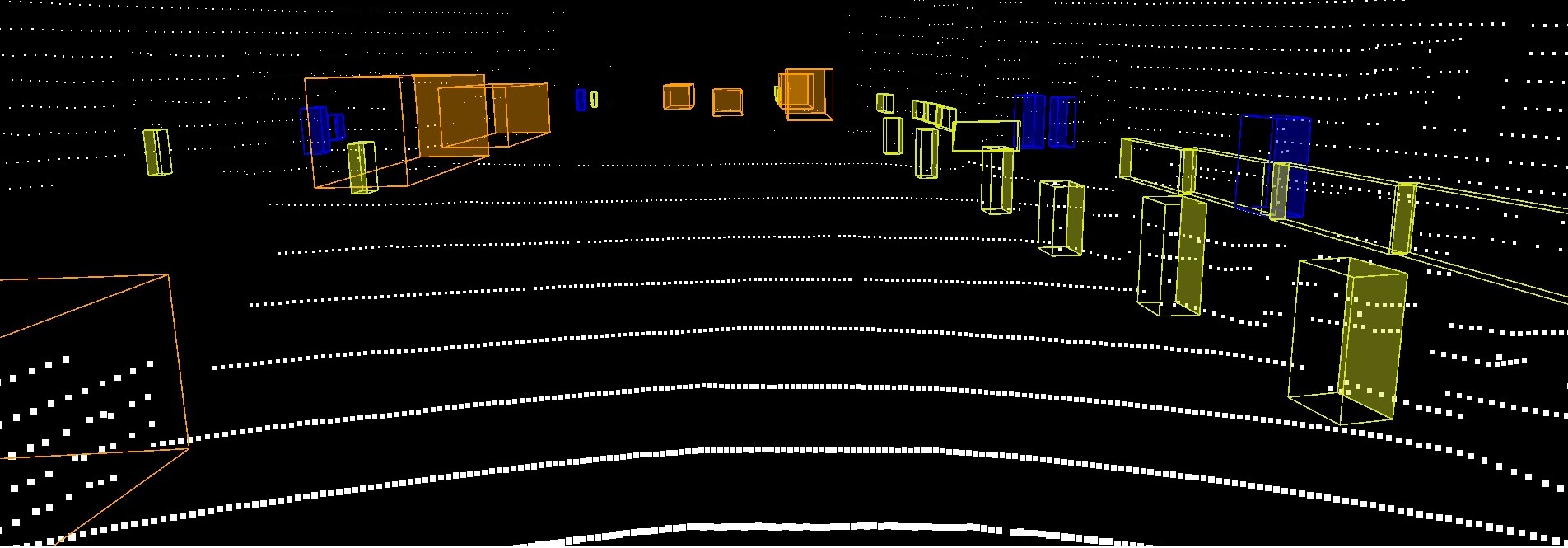}
    \caption{ Annotated LiDAR point cloud image at time $t_0$.}
\end{subfigure}    
\begin{subfigure}{0.48\linewidth}
    \centering
    \includegraphics[width=1\linewidth]{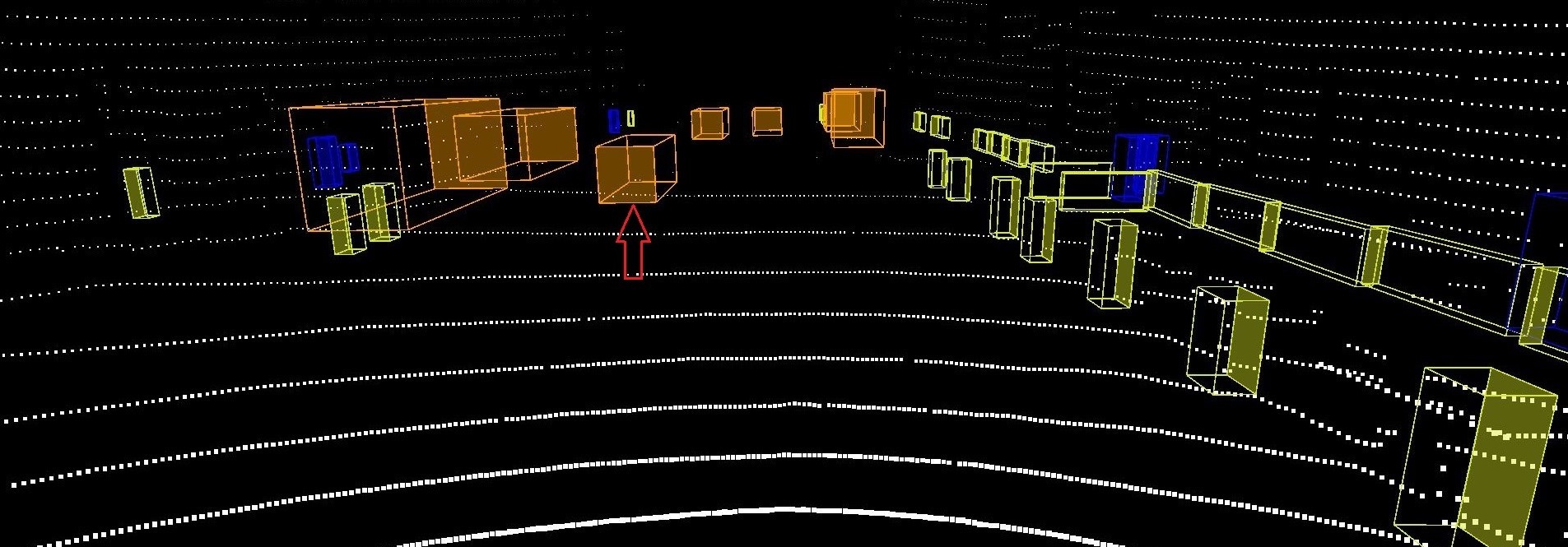}
    \caption{ Annotated LiDAR point cloud image at time $t_1$.}
\end{subfigure}    
\begin{subfigure}{0.48\linewidth}
    \centering
    \includegraphics[width=1\linewidth]{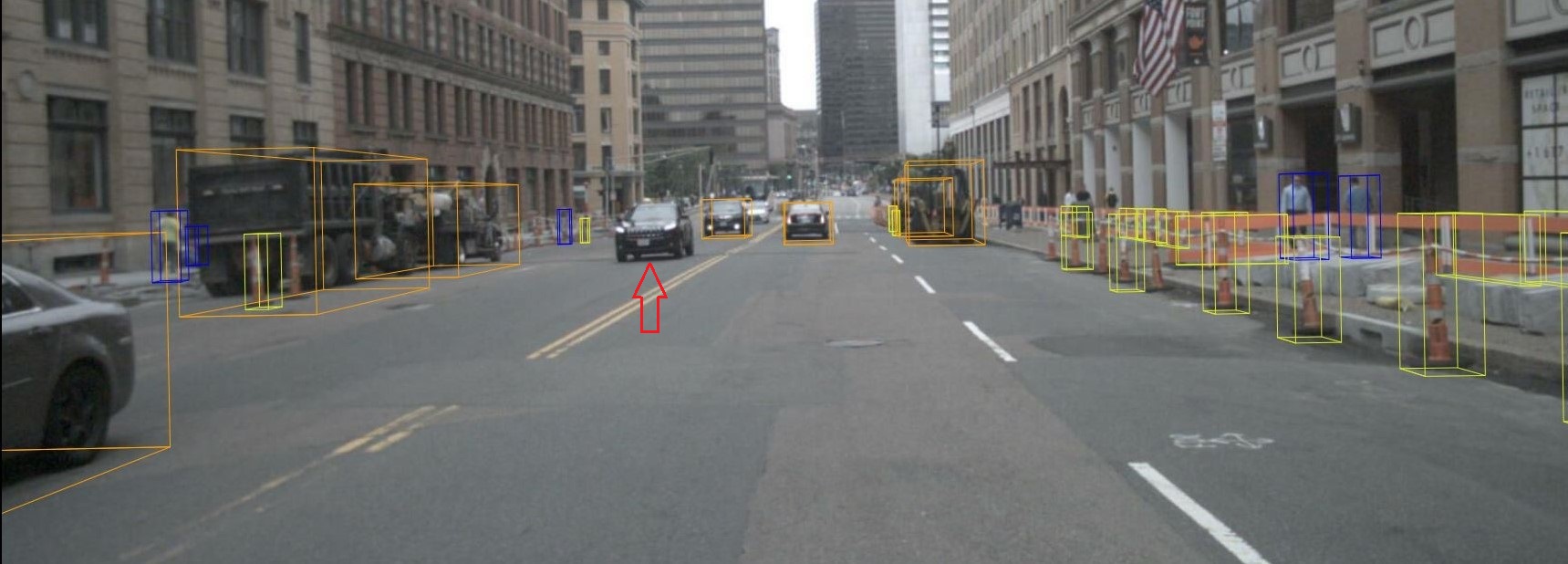}
    \caption{ Annotated camera image for the LiDAR image on the left.}
\end{subfigure}  
\begin{subfigure}{0.48\linewidth}
    \centering
    \includegraphics[width=1\linewidth]{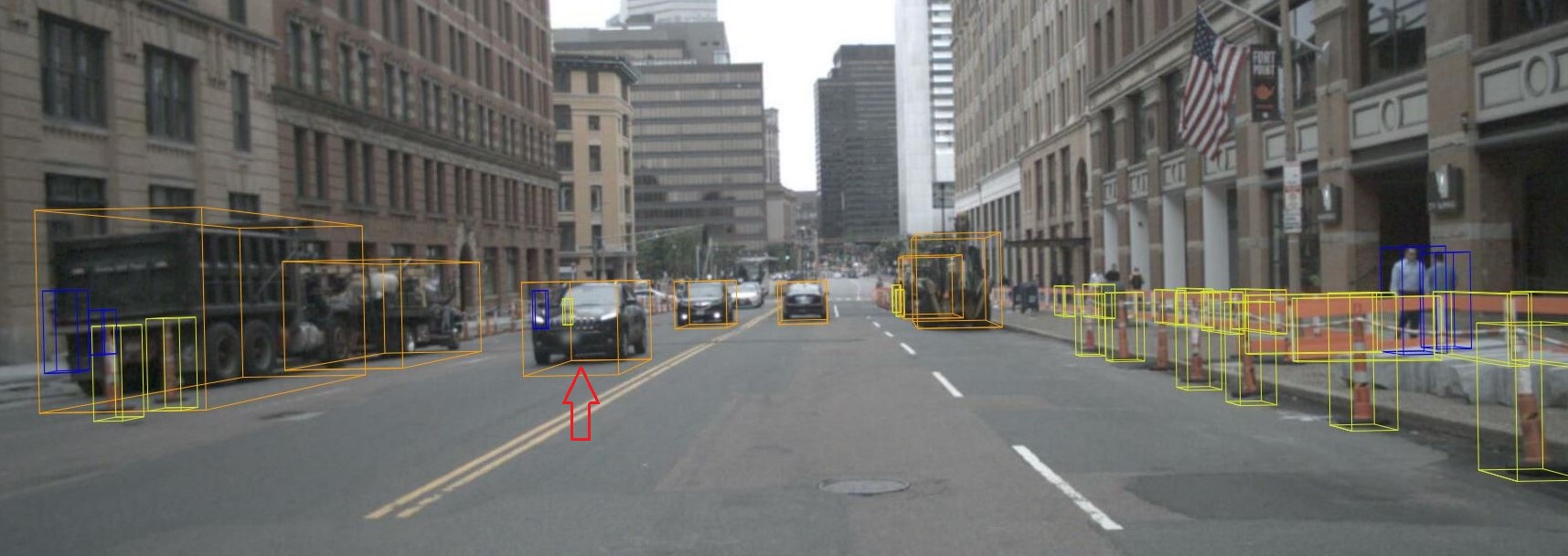}
    \caption{ Annotated camera image for the LiDAR image on the left.}
\end{subfigure}  
\hfill
\begin{subfigure}{0.48\linewidth}
    \centering
    \includegraphics[width=1\linewidth]{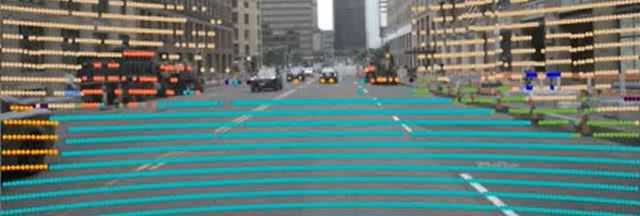}
    \caption{ Colored semantic segmented of the LiDAR data at time $t_0$.}
\end{subfigure}    
\begin{subfigure}{0.48\linewidth}
    \centering
    \includegraphics[width=1\linewidth]{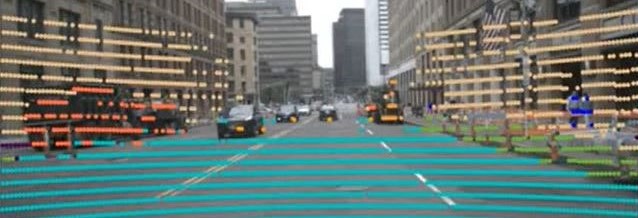}
    \caption{ Colored semantic segmented of the LiDAR data at time $t_1$.}
\end{subfigure}  
\caption{Two LiDAR and Camera frames annotated data along with the semantic segmentation data are taken from NuScenes LidarSeg dataset (Scene-0247) (\cite{caesar2020nuscenes}).}
\label{fig:3D-example-1}
\end{figure*}

\begin{figure}[tbp]
  \vspace{0.75em}
\centering
\begin{subfigure}{0.48\linewidth}
    \centering
    \includegraphics[width=1\linewidth]{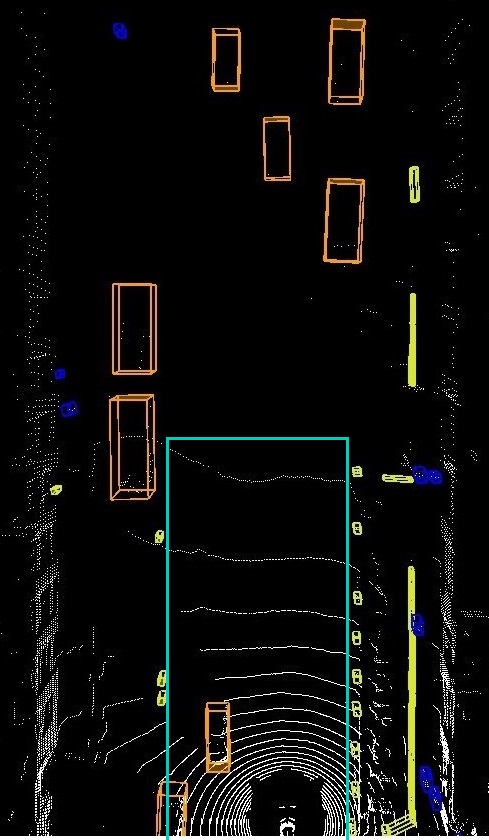}
    \caption{ \textit{x-y plane} bird-eye view of LiDAR point cloud at time $t_0$.}
\end{subfigure}    
\begin{subfigure}{0.48\linewidth}
    \centering
    \includegraphics[width=1\linewidth]{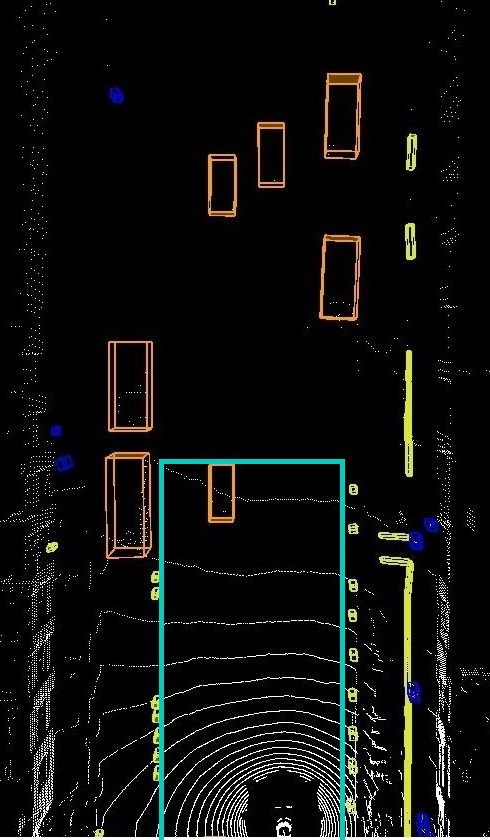}
    \caption{ \textit{x-y plane} bird-eye view of LiDAR point cloud at time $t_1$.}
\end{subfigure}  
\caption{The bird-view of LiDAR point clouds for the two consecutive frames in Fig. \ref{fig:3D-example-1} where the drivable area in front of the ego is shown.}
\label{fig:3D-example-1-topview}
\end{figure}

Here, we are going to show by example how 3D reasoning is possible using STPL.
Specifically, we are going to show by example if the perception system misses some important objects.
The six images in Figure \ref{fig:3D-example-1} illustrates annotated information of two consecutive frames that are taken from nuScenes-lidarseg\footnote{\url{https://www.nuscenes.org/nuscenes?sceneId=scene-0274&frame=0&view=lidar}} dataset (\cite{caesar2020nuscenes}).
For each detected class of objects, they adopt color coding to represent the object class in the images.
The color orange represents cars, and the color cyan represents drivable regions.
The color-coded chart of the classes, along with their frequency in the whole dataset can be found online (\cite{nuscenes2020website}).
In the first frame, at time $t_0$, the LiDAR annotated image is shown in Fig. \ref{fig:3D-example-1}(a).
The corresponding camera image for the LiDAR scene is shown in Fig. \ref{fig:3D-example-1}(c).
There is a car in this image that we pointed to in a red arrow. 
As one can check, there is no point cloud data associated with the identified car in the LiDAR image, and as a result, it is not detected/recognized as a car.
In the next images illustrated in Fig. \ref{fig:3D-example-1}(b,d), the undetected car at $t_0$ is identified and annotated as a car at time $t_1$.
Unlike before, there are some point cloud data associated with the car by which it was detected.
The semantic segmentation of the corresponding images is shown in Fig. \ref{fig:3D-example-1}(e-f).
First, we will discuss how to decide if there is a missed object in the datasets, similar to what we presented here.
Second, we represent a requirement specification in STPL that can formalize such missed-detection scenarios.

\begin{exmp}
For this and the next example scenarios, we have 3D coordinates of bounding volumes and 3D point clouds representing each detected object in the frames. 
However, the height of those points and cubes are not helpful in these cases, and therefore, we mapped them to the \textit{x-y plane} 
by ignoring the \textit{z-axis} of sensor positions (i.e., see \cite{nuscenes2020website}).

Let assume that the origin of the coordinate system is the center of the ego vehicle, and the \textit{y-axis} is positive for all the points in front of the ego car, and the \textit{x-axis} is positive for all the points to the right of the ego.
That is, we have 2D bounding boxes of classified objects that are mapped to a flat ground-level environment as shown in Fig. \ref{fig:3D-example-1-topview}.
Therefore, 
we do not need to define new spatial functions for 3D reasoning.

\begin{requirement}\label{req:drive-toward}
\textit{
If there is a car driving toward the ego car in the current frame, it must exist in the previous frame.
}
\end{requirement}

\noindent\textbf{STPL} \hlRev{(using Assumptions \ref{assum:always-detect}-\ref{assum:unique-id}):}

\begin{multline}\label{ex:3d-1}
     \phi_{\ref{req:drive-toward}} = \Box \forall Id_1. \forall Id_3 @x. \Big( \big(
     \varphi_{exist}^{car}
     \wedge \varphi_{close}^{dist} 
     \big)
     \implies
     \\
     \hlRev{\Previous_{w}}
     \;\varphi_{existed}^{before}
     \Big)
\end{multline}
and the predicates are defined as:
\begin{gather*}
\varphi_{exist}^{car} \equiv 
     \big(C(Id_1) = Car\big) \wedge 
     \big( C(Id_3)= Drv \big)  
\end{gather*}
\begin{gather*}
\varphi_{close}^{dist} \equiv
     LON(Id_1,\BM) \le 1.2 \times LON(Id_3,\BM)
     \wedge 
	\displaybreak[2] 
     \\
     LON(Id_1,\TM) \ge 0.5 \times LON(Id_3,\BM)
     \wedge 
	\displaybreak[2] 
     \\
     LAT(Id_1,\CT) \ge LAT(Id_3,\LM)
     \wedge 
	\displaybreak[2] 
     \\
     LAT(Id_1,\CT) \le LAT(Id_3,\RM)
\end{gather*}
\begin{gather*}
\varphi_{existed}^{before} \equiv
     \exists Id_2. \big( C(Id_2)=Car \wedge Id_1 = Id_2 
     \wedge
	\displaybreak[2] 
     \\ 
     Dist(Id_2,\CT,\Suniverse,\CT) > Dist(Id_1,\CT,\Suniverse,\CT)
     \big)
\end{gather*}
Note that in the above 2D spatial functions, we used $\BM$ and $\TM$ in contrast to their original semantics in the previous examples due to the change of origin of the universe.
Also, we assume that the perception system uses an accurate object tracker that assigns unique IDs to objects during consecutive frames.
We used $Id_1$, $Id_2$, and $Id_3$ to refer to a car in the current frame, a car in the previous frame, and the drivable region.
Note that we needed to use the previous frame operator to check if an existing car was absent in the previous frame.
That is, in $\varphi_{existed}^{before}$, we check if the car identified by $Id_2$ matches with the same car in the previous frame.
\hlRev{The antecedent of the formula as in Eq. (\ref{ex:3d-1}) holds in the current frame, but the consequent of it fails because the car that exists in the current frame did not exist in the previous frame.
Thus, the data stream shown as image frames in Fig. \ref{fig:3D-example-1-topview} falsifies the formula.}

It is more realistic to formulate the same requirement using lane-based coordinate systems, if the perception system detects different lanes.
Consequently, we can simplify the subformula $\varphi_{close}^{dist}$ to encode if the cars drive in the same lane.
\end{exmp}

\subsubsection{3D occlusion scenario}\label{lbl:3d-occ}

\begin{figure*}[tbp]
  \vspace{0.75em}
\centering
\begin{subfigure}{0.23\linewidth}
    \centering
    \includegraphics[width=1\linewidth]{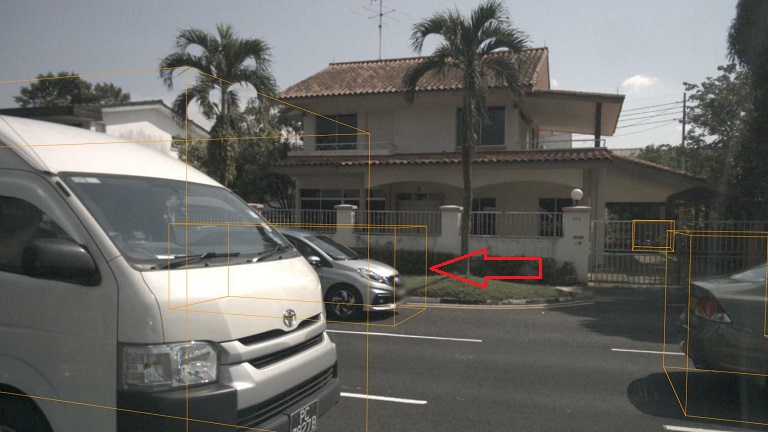}
    \caption{ Annotated camera image for the LiDAR image on the right.}
\end{subfigure}    
\begin{subfigure}{0.23\linewidth}
    \centering
    \includegraphics[width=1\linewidth]{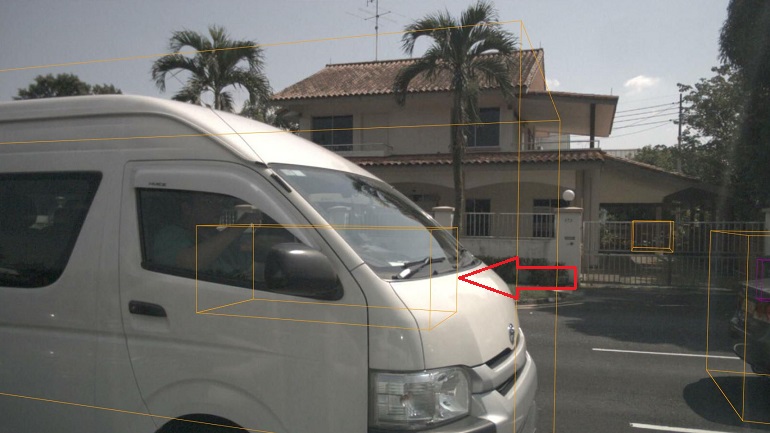}
    \caption{ Annotated camera image for the LiDAR image on the right.}
\end{subfigure}    
\begin{subfigure}{0.23\linewidth}
    \centering
    \includegraphics[width=1\linewidth]{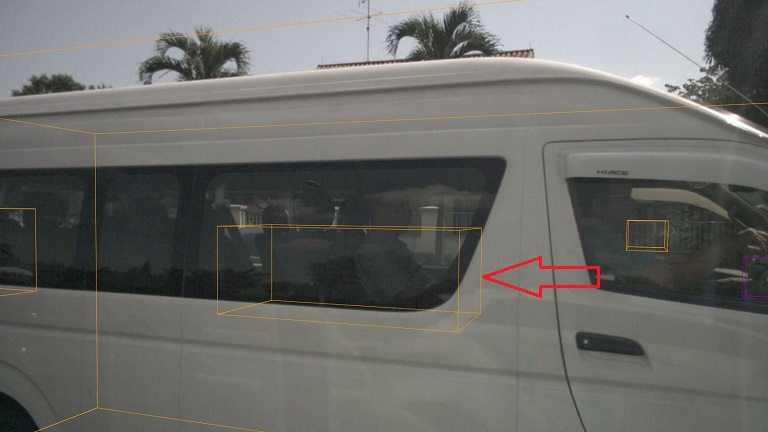}
    \caption{ Annotated camera image for the LiDAR image on the right.}
\end{subfigure}    
\begin{subfigure}{0.23\linewidth}
    \centering
    \includegraphics[width=1\linewidth]{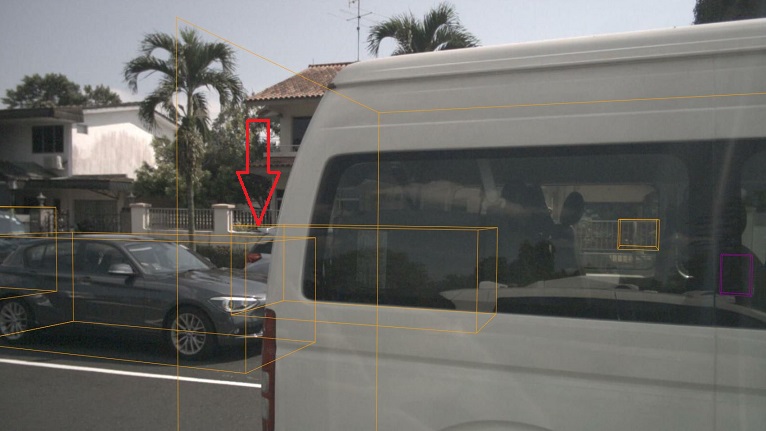}
    \caption{ Annotated camera image for the LiDAR image on the right.}
\end{subfigure}    
\begin{subfigure}{0.23\linewidth}
    \centering
    \includegraphics[width=1\linewidth]{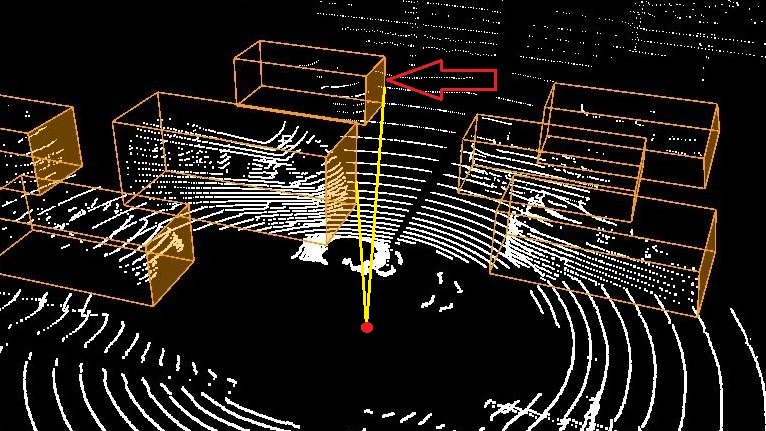}
    \caption{ Annotated LiDAR point cloud image at time $t_0$.}
\end{subfigure}  
\begin{subfigure}{0.23\linewidth}
    \centering
    \includegraphics[width=1\linewidth]{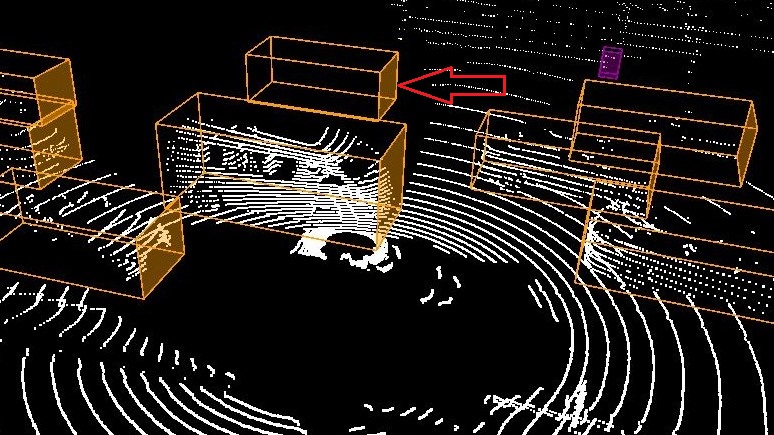}
    \caption{ Annotated LiDAR point cloud image at time $t_1$.}
\end{subfigure}  
\begin{subfigure}{0.23\linewidth}
    \centering
    \includegraphics[width=1\linewidth]{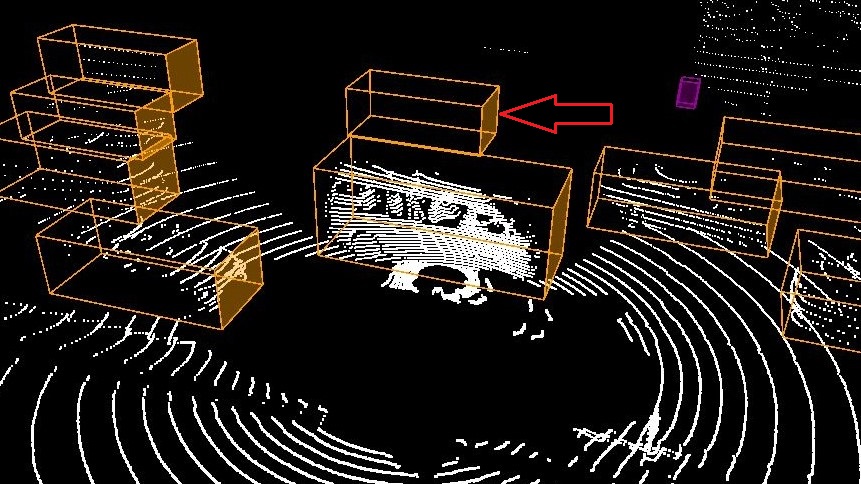}
    \caption{ Annotated LiDAR point cloud image at time $t_2$.}
\end{subfigure}  
\begin{subfigure}{0.23\linewidth}
    \centering
    \includegraphics[width=1\linewidth]{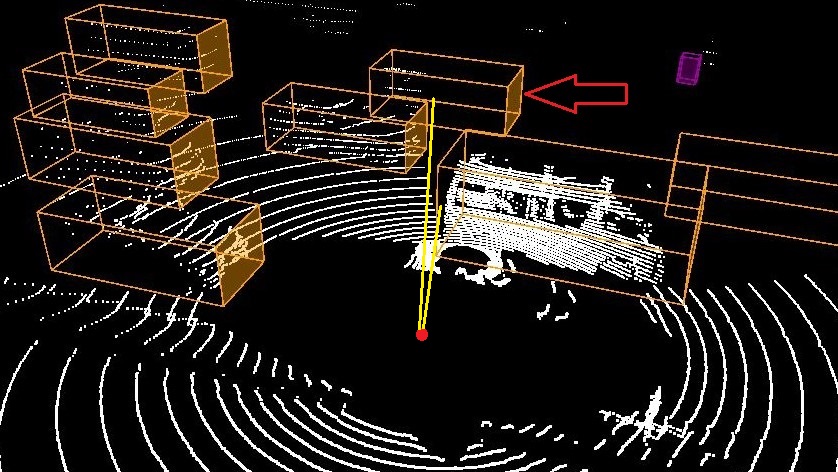}
    \caption{ Annotated LiDAR point cloud image at time $t_3$.}
\end{subfigure}  

\caption{Two LiDAR and Camera frames annotated data are taken from NuScenes LidarSeg dataset (Scene-0399) (\cite{caesar2020nuscenes}). The images are taken by the back-left camera as represented in sensor-position image available at nuScenes website (\cite{nuscenes2020website}).
}
\label{fig:3D-example-2}
\end{figure*}

In Figure \ref{fig:3D-example-2}, there are four frames captured at $t_0$ to $t_3$.
In the second and third frames, an occluded car (pointed to a red arrow) is detected. 
The perception system detected a car for which there is no information from the sensors (the cubes in frames 2 and 3 are empty, and the cameras cannot see the occluded vehicle).
Although this can be the desired behavior for a perception system that does object tracking, here we are going to formalize a requirement that detects similar cases as wrong classifications.

For this example, we need to define a new 3D spatial function \textit{Visible} by which we can check if given two cars are visible from the view of a third car.
Additionally, we need to add a data attribute \textit{Empty} to the data-object stream $\Data$ that determines if a bounding volume of an object is empty (e.g., $\Retrieve(\Data(i),ID).{PC}=\emptyset$ checks if there is no point could data associated with an object identified by $ID$) or not.
For instance, by $\Retrieve(\Data(i),ID).Empty$ we check if an object identified by $ID$ in the $i$'th frame is empty.
We use function $E(id)$ as a new syntax that serves as above.
\begin{align*}
    &\dle V(id,\CRT,id,id) \sim r \dri (\Data,i,\epsilon,\zeta,\revision{\tau}) := 
	\\
	&\;\;\;\;\;\;\;\;\;\;\;\;\displaybreak[2] \left\{ \begin{array}{ll}
	    \top
	    \mbox{ if \;\;} f_{visible}\big(id,\CRT,id,id ,\Data,i,\epsilon,\zeta)\big) \sim r
	    \\
	    \bot
	    \mbox{ otherwise}^*  
	    \end{array} \right.
	\displaybreak[2]     
\end{align*}

$\boldsymbol{f_{visible}(id_1 , \CRT,id_2,id_3 ,\Data,i,\epsilon,\zeta)}$ 
returns \textit{true} if the $\epsilon(id_2)$ and $\epsilon(id_3)$ objects are visible from the view point $\CRT$ of $\epsilon(id_1)$ in the $k$'th frame,
where $k \leftarrow \zeta(id_1)$ if $\zeta(id_1)$ is specified, and $k \leftarrow i$ otherwise. 
If one or both are invisible from the viewpoint of object $\epsilon(id_1)$, then the function returns \textit{false}.

\begin{exmp}
\revision{The below requirement checks the robustness of the object detection in a perception system using spatial constraints and functions on point cloud data.}
\begin{requirement}\label{req:point-cloud-occlusion}
\textit{
Any detected car (using the LiDAR data) driving the same direction on the left side of the ego car must include some point cloud data in its bounding volume unless it is occluded.}
\end{requirement}
\noindent\textbf{STPL} \hlRev{(using Assumptions \ref{assum:dif-modules}-\ref{assum:unique-id}):}
\begin{multline}\label{ex:3d-2}
    \phi_{\ref{req:point-cloud-occlusion}} = \Box \forall Id_1. \forall Id_2. \bigg( \Big ( 
    \neg E(Id_1) \wedge \neg E(Id_2) \wedge (Id_1 \neq Id_2)
    \\
    \wedge 
    \big( C(Id_1) = Car \big) \wedge \big( C(Id_2) = Car \big) 
    \\
    \wedge
    LAT(Id_1,\RM) < -1 \wedge LAT(Id_2,\RM) < -1
	\displaybreak[2] 
    \\
    \wedge
    LAT(Id_1,\RM) > -6 \wedge LAT(Id_2,\RM) > -6
    \\
    \wedge
    \hlRev{MD(Id_1) = MD(Id_2)}
    \\
    \wedge
    V(\Suniverse,\CT,Id_1,Id_2) 
    \\
    \wedge
    \Next \big(\neg V(\Suniverse,\CT,Id_1,Id_2) 
    \big)
    \Big)
    \implies
    \\
    \Next \Big( V(\Suniverse,\CT,Id_1,Id_2) 
    \\
    \;\Rc\; 
    \\
     \big ( E(Id_1) \wedge \neg E(Id_2) \big )
    \Big)
    \bigg)
\end{multline}
In the above formula, we keep track of two cars using $Id_1$ and $Id_2$ that are to the left of the ego car (their distance has to be limited between 1 to 6 meters from the left side of the ego car).
\hlRev{We assume that the perception system returns the moving direction of each classified object, and we use function $MD$ to retrieve the direction for a given object identifier.
Note that this function can be replaced by a spatial formula that uses the coordinates of an object in different frames to calculate its moving direction. 
}
The subformula $V(\Suniverse,\CT,Id_1,Id_2)$ requires a visible angle between the two cars from the point of view of the ego car (i.e., the center point of the $\Suniverse$) to return true.
We draw a sample visible angle in the first and fourth frames in yellow.
For the two cars that we could find a visible angle for them in the first frame, there is none in frames 2 and 3.
Additionally, there is no point cloud in the occluded car's bounding volume in frames 2 and 3 that satisfies the whole formula until the implies operator.
The consequence of the implication is a release formula to be satisfiable from the next time/frame.
In the release formula, the right-hand side requires the occluded object to be empty, and only when it becomes visible again, it can be non-empty.
The whole formula is satisfiable for the pair of cars we discussed here.
\hlRev{The data stream shown as image frames in Fig. \ref{fig:3D-example-2} satisfies the formula in Eq. (\ref{ex:3d-2}).}
\end{exmp}

\revision{
\section{Monitoring Algorithm}
\label{lbl:apx:algorithm}
}
\revision{
The offline monitoring algorithm for STPL is one of the main contributions of this work.
\revisionTwo{
We use a dynamic programming (DP) approach similar to \cite{dokhanchi2016efficient,fainekos2012verification} since we are operating over data streams (timed state sequences).
To keep the presentation as succinct as possible, we focus the presentation only to the future fragment of STPL.
}
We divide our monitoring algorithm into four algorithms that work in a modular way.
We present the pseudocode of the algorithms to analyze the worst case complexity of our STPL monitoring framework.
}
In the following, we are going to use the formula 
\begin{multline*}
    \varphi := \Box \forall Id_1@x. \Box \exists Id_2.
    \Big( \Sforall \big(\overline{\BB(Id_1)} \sqcup \BB(Id_2)\big) \wedge 
    \\
    \Sforall \big(\overline{\BB(Id_2)} \sqcup \BB(Id_1)\big) \wedge Id_1 = Id_2 \Big)
\end{multline*}
to explain some of the aspects of the future fragment of our monitoring algorithm.

\subsection{Algorithm-\ref{alg:DP-main}}
\begin{figure}
    \centering
    \begin{subfigure}{1\linewidth}
    \includegraphics[width=1\linewidth]{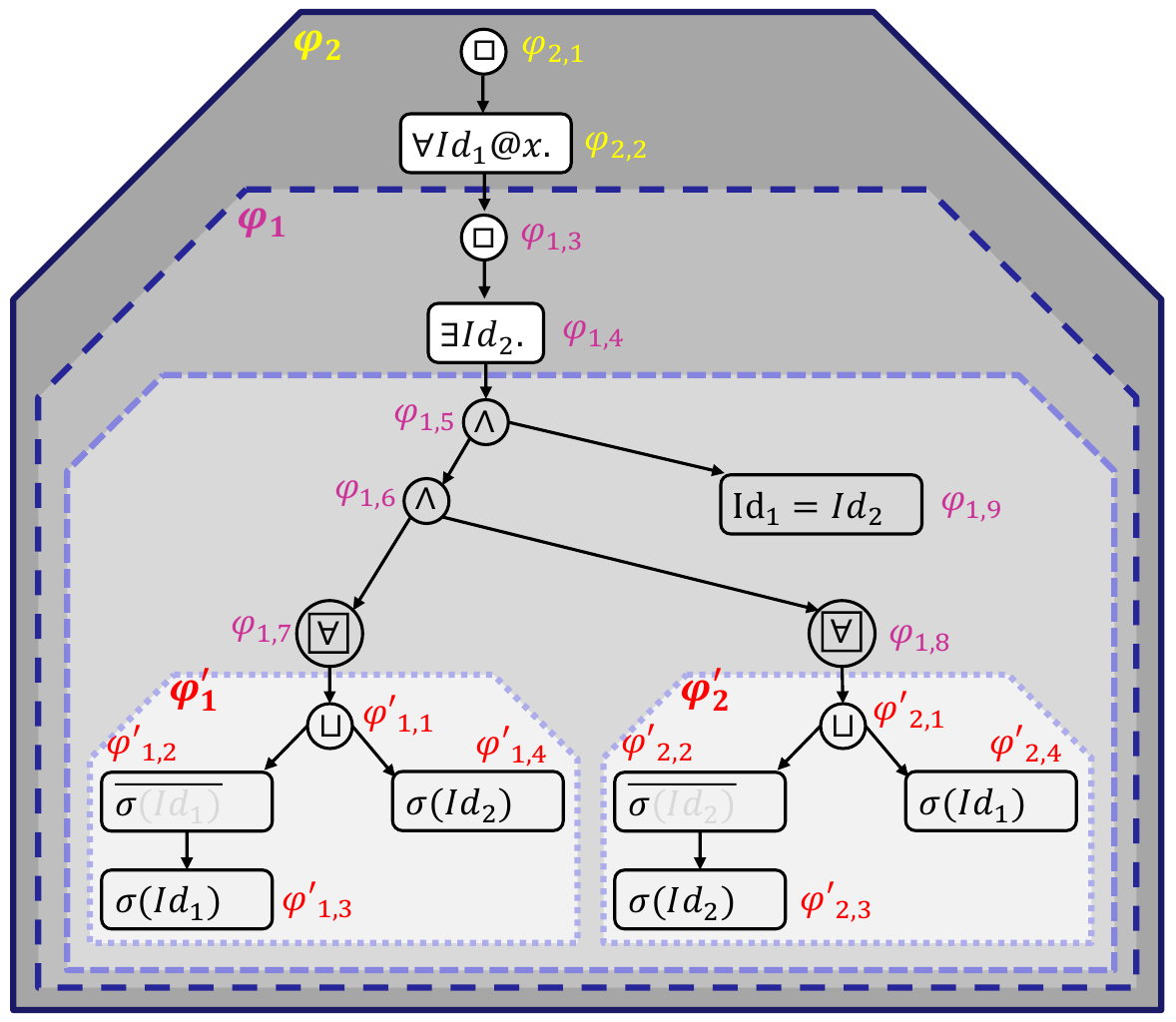}
    \end{subfigure}
    \caption{
    A parse tree for the formula in Eq. \ref{ex:3:a}. Distinct regions represent subformulas $\varphi_2$, $\varphi_1$, $\varphi'_1$, and $\varphi'_2$. In each region, the subformulas subscript into the operators and predicates levels.
    }
    \label{fig:monitor-exm-tree}
\end{figure}

This algorithm represents the main dynamic programming body of the monitoring algorithm.
It receives an STPL formula $\varphi$, a data stream $\hat{\rho}$, and dynamic programming tables $M$ and $F$.
\revisionTwo{
The formula is divided into subformulas as in the example in Fig. \ref{fig:monitor-exm-tree}.
That is, the formula is divided into subformulas based on the presence of freeze time operators ($\varphi_i$) and spatial quantifiers ($\varphi_i'$).
It is important to note that the nested subformulas have lower index, e.g., $\varphi_1$ is nested in $\varphi_2$ as in Fig. \ref{fig:monitor-exm-tree}.
\revisionTwo{Notice that when we have subindexes, i.e., $\varphi_{i,j}$, we refer to the formula in the scope of $i$'th freeze variable and the $j$'th index from the root in the depth-first-search order.}
}

\revisionTwo{
The main loop of Algorithm \ref{alg:DP-main} has 4 nested loops (lines \ref{alg1:v_loop}-\ref{alg1:v_loop:end}). 
Further nested loops exist in $ComputeFnExpr$ (Algorithm \ref{alg:DP-FNEXPR}).
These nested loops are also the source of the computational complexity in our monitoring algorithm.
The loops in lines \ref{alg1:v_loop} and \ref{alg1:t_loop} are responsible for exploring all possible assignments to the freeze time variables starting from the inner nested freeze time operators.
Line \ref{alg1:u_loop} explores the data stream backward in time to resolve the future fragment of Linear Temporal Logic (LTL) and the $S4_u$ terms.
Lines \ref{alg:main:subform:type}-\ref{alg:main:subform:type:end} identify whether the subformula $\varphi_{k,j}$ is a spatio-temporal operator and calls the respective function.
}

\subsection{Algorithm-\ref{alg:DP-LTL}} 
This algorithm computes the values of the DP table $M$ as defined in the temporal semantics of the STPL logic in Def. \ref{stql:smt:tmp}.
For a given subformula, first, it determines the operator and then fills the current columns of the DP tables accordingly.
Most of the lines of the algorithm are straightforward where there is no DP table $F$ in the assignments.
The idea of using the separate DP table $F$ for the freeze subformulas is to store the result of their evaluation at each frozen time, and then update the DP table $M$ accordingly. 
Table $M$ has two columns for each subformula, one for the current time and another for the next time.
We use $u_{01}$ and $u_{10}$ to refer to the current column (current frame) for updating and the next column (next frame) for reading, respectively.
Therefore, we need to toggle the value of these variables (between 0 and 1) to change the read column to write column and vice versa.
Table $F$ has two rows for each frame, one for the current time and the other for the next time.
We use $t_{01}$ and $t_{10}$ to refer to the current row for updating and the next row for reading, respectively.
Therefore, we need to toggle the value of these variables (between 0 and 1) to change the read row to write row and vice versa.
The $M[j,u_{01}].val$ data-member of the DP table $M$ keeps the current evaluating result of the $j$'th subformula $\varphi_j$ at the current frame.
The other data-members are used to update \textit{min/max} values for the quantifier operators.
The $IT$ data-member is a table of size $(S_O)^{|V_{id}|}$, where $S_O$ is the maximum number of objects in all input frames, and $V_{id}$ is the set of the $Id$ variables in the scope of the evaluating subformula $\varphi_j$.
\revisionTwo{
In Alg. \ref{alg:DP-LTL}, the only computationally expensive part is the update of the table in line \ref{alg2:update-table}.
In practice, the table update is not an issue since the number of $id$ variables is not high (at most 4-5 object variables in the examples in this paper) and the table update can be parallelized.
}

\begin{algorithm}[t]
\footnotesize{
	\caption{STPL Monitor}
	{\bf Input}: $\varphi$, 
	$\hat{\rho}$; 
	{\bf Global variables:} 
    $M_{|\varphi|\times 2}$,
	{$F_{2 \times |\hat{\rho}|}$};
	{\bf Output}: $M[1,0]$.

\revisionTwo{
 {\it Comments: \\
 $\overline{M}_{|\varphi'|\times |\hat{\rho}|}$ is a table used to store the evaluation of spatial subformulas. For all the tables, the out-of-index accesses are handled case by case (i.e., $\top$ or $\bot$ is a default value)
 }
}
 
 \label{alg:DP-main}

		{\bf   Procedure  }{\sc STPL-Monitor}($\varphi$,
		$\hat{\rho}$)
		\begin{algorithmic}[1]
			\State $u_{01} \leftarrow 0$ and $t_{01} \leftarrow 0$ \Comment{toggle between 0 and 1}
			\State $\varphi' \leftarrow $ \revisionTwo{Collect all spatial terms, arithmetic expressions, and functions}
			\State $V \leftarrow $ \revisionTwo{Collect all freeze time variables in $\varphi$}
			\For{$k\gets 1\mbox{ to }|V|$}\label{alg1:v_loop} 
			\For{$t\gets 0\mbox{ to }|\hat{\rho}|-1$}\label{alg1:t_loop} \Comment{\revisionTwo{freeze (store) the frame}}
			\For{$u\gets |\hat{\rho}|-1\mbox{ to }t$}\label{alg1:u_loop} \\
                        \Comment{backward iteration for the future fragment of STPL}
			\For{$j\gets \varphi_k.max\mbox{ down to }\varphi_k.min$}\label{alg1:j_loop} \\
                    \Comment{iterate over subformulas under each freeze time operator}
            \If {$\revisionTwo{\varphi_{j}} \in \varphi'$} \label{alg:main:subform:type}
            \State {$ComputeFnExpr(\varphi,\revisionTwo{j},u,u_{01},t,\hat{\rho})$}
			\Else
			\color{black}
			\State  $ComputeLTL(\revision{\varphi,j},u,u_{01},t,t_{01},k)$
			\EndIf \label{alg:main:subform:type:end}
			\EndFor
			\State ${u_{01} \leftarrow toggle(u_{01})}$
			\EndFor	\label{alg1:t_loop:end}
			\EndFor 
			\State \revision{${t_{01} \leftarrow toggle(t_{01})}$}
			\EndFor \label{alg1:v_loop:end}
            \State $k \leftarrow |V|+1$ \label{alg1:end_loop_1} \Comment{\revisionTwo{handle the root node}}
			\If{{$\varphi_{k} \mbox{ is not a freeze operator}$}}
                \State Repeat steps \ref{alg1:u_loop} to \ref{alg1:t_loop:end} for $t=0$
			\Else
			\State ${u_{01} \leftarrow toggle(u_{01})}$ , ${t_{01} \leftarrow toggle(t_{01})}$
		    \State ${M[1][u_{01}] = F[t_{01}][0]}$
			\EndIf
			\State\Return$M[1,u_{01}]$			
		\end{algorithmic}
		\hspace {5pt}{\bf   end procedure}
		}
\end{algorithm}

\subsection{Algorithm-\ref{alg:DP-FNEXPR}}
This algorithm computes the identifier table $IT$ values of the DP table $M$.
Here, we compute spatial functions and quantifier operators.
For a given spatial function $Fn(\dots)$, if it has nested function operators, then this algorithm recursively computes the final values.
The $IT$ table stores the values for any functional expression based on the combinatorial assignments of objects to the object variables.
In line \ref{alg3:if-id-exists}, if the combination of object identifiers does not satisfy the correct range of existing objects in the current frame or a frozen frame, then it assigns 
$\bot$ (false).
Otherwise, if any parameter is a spatial formula, then it runs algorithm \ref{alg:DP-S4U-INT}.
 The size of the $IT$ table is the maximum number of combinations for objects with respect to the object variables.
 For example, if there are a maximum of five objects per frame in an object stream, and a formula has a maximum of three nested object variables, then the size of $IT$ is $5^3=125$.
We choose the size of the $IT$ table based on the worst case scenario, but for efficient computation, we only compute as many as needed combinations for each quantifier subformula based on its scope.
Therefore, we mark the remaining cells of the table with $\nan$ for extended computation that merges these tables for different subformulas in Algorithm \ref{alg:DP-LTL} by calling function $UpdateIdTable$ at line \ref{alg2:update-table}.


\begin{algorithm}[t]
\footnotesize{
	\caption{LTL Monitor}
	{\bf Input}: $\revision{\varphi,j},u,u_{01},t,t_{01},\revision{k}$; 
	\label{alg:DP-LTL}
	
\hspace {10pt}{\bf   procedure  }{\sc ComputeLTL}($\revision{\varphi,j},u,u_{01},t,t_{01},\revision{k}
$)
	\begin{algorithmic}[1]
	\State {$u_{10}\leftarrow toggle(u_{01})$,$t_{10}\leftarrow toggle(t_{01})$}
	    \If {{$\varphi_j \equiv \top$ }}
	    \State {$M[j,u_{01}].val \leftarrow \top$}
	    \ElsIf {{$\varphi_j \equiv \Ctime - v_k\sim r$}}
			\If {$(\tau_u-\tau_t)\sim r$}
			\State $M[j,u_{01}].val \leftarrow \top$
			\Else
			\State $M[j,u_{01}].val \leftarrow \bot$
			\EndIf
	    \ElsIf {{$\varphi_j \equiv \Cframe - v_k\sim r$}}
			\If {$(u-t)\sim r$}
			\State $M[j,u_{01}].val \leftarrow \top$
			\Else
			\State $M[j,u_{01}].val \leftarrow \bot$
			\EndIf
		\ElsIf {{$\varphi_j$ is a freeze subformula}}\label{alg2:freeze-if}
		\State {$M[j,u_{01}] \leftarrow M[j+1,u_{01}]$}
		    \If {{$\varphi_j$ has an $\exists$ quantifier}}
		    \State {$M[j,u_{01}].val \leftarrow M[j,u_{01}].max$}
		    \Else
		    \State {$M[j,u_{01}].val \leftarrow M[j,u_{01}].min$}
		    \EndIf
	    \State {$F[t_{01}][t] \leftarrow M[j,u_{01}]$} \label{alg2:f-update}
	    \State {$UpdateIdTable(M[j,u_{01}].IT)$}\label{alg2:update-table} 
		\ElsIf {$\varphi_j\equiv\neg\varphi_m$}
		 \State $M[j,u_{01}] = \neg M[m,u_{01}]$ \Comment{\revisionTwo{Apply $\neg$ to val, min, max, IT}}
  \label{alg2:IT-neg}
		\ElsIf {$\varphi_j\equiv\varphi_m \vee \varphi_n$}
		\State {$M[j,u_{01}] \leftarrow$ max($M[m,u_{01}],M[n,u_{01}]$)}
		\ElsIf {$\varphi_j\equiv \bigcirc \varphi_m$}
		\If{$u=|\hat{\rho}|-1$}
		\State {$M[j,u_{01}] \leftarrow \bot$}
		\ElsIf{$\varphi_m$ is followed by $@\{x,\dots\}$}\label{alg2:next-follow-freeze}
		\State {$M[j,u_{01}] \leftarrow F[t_{10},u+1]$}
		\Else
		\State {$M[j,u_{01}] \leftarrow M[m,u_{10}]$}
		\EndIf
		\ElsIf {$\varphi_j\equiv\varphi_m \Un \varphi_n$}
		\If{$u=|\hat{\rho}|-1$}
		\State {$M[j,u_{01}] \leftarrow M[n,u_{01}]$}
		\Else
		\If{$\varphi_m$ is followed by $x.$}
		\State {$M[m,u_{01}] \leftarrow$ 
		$F[t_{10},u]$}
		\EndIf
		\If{$\varphi_n$ is followed by $x.$}
		\State {$M[n,u_{01}] \leftarrow$ 
		$F[t_{10},u]$}
		\EndIf
		\State {$M[j,u_{01}] \leftarrow$ }
		\par
		\hskip\algorithmicindent
		{max($M[n,u_{01}],$ min($M[m,u_{01}],M[j,u_{10}]$))}
		
		\EndIf
		\EndIf
	\end{algorithmic}
	\hspace {5pt}{\bf   end procedure}
	}
\end{algorithm}


\subsection{Algorithm-\ref{alg:DP-S4U-INT}} 
This algorithm computes the values of the DP table $\overline{M}$.
\revisionTwo{
It evaluates spatial-temporal formulas similar to the temporal formulas of Algorithm \ref{alg:DP-LTL} with the modification that now set operations are used as opposed to Boolean operations.
The data structures for the set representation and the corresponding set operations can affect the computational complexity of \ref{alg:DP-S4U-INT}.
Depending on the application, e.g., 2D vs 3D, a number of solutions are possible ranging from a linked list of pairs of 2D points (i.e., union of rectangles) to quadtrees or octrees (see next section).
}


\color{black}
\begin{algorithm}[t]
\footnotesize{
	\caption{Compute Function Expression}
	{\bf Input}: $\varphi,k,u,u_{01},t,\hat{\rho}$;
	\label{alg:DP-FNEXPR}
	
	\hspace {10pt}{\bf   procedure  }{\sc ComputeFnExpr}($\varphi,k,u,u_{01},t,\hat{\rho}$)
	\begin{algorithmic}[1]
	    \State $\varphi_k \equiv Fn(\Omega_1,\dots,\Omega_n) \sim r$ where
		\par
		\hskip\algorithmicindent
	    $r \in \mathbb{R}$, $\sim\; \in$ \textnormal{\scriptsize{$\{>,<,\ge,\le,=,!=\}$}},
	    \par
	    \hskip\algorithmicindent
	    and $Fn$ is a reserved function name
	    \State $P_{n \times 1}$ is a list used to store the evaluated arguments of $Fn$
	    \State $S_{id} \leftarrow (S_O)^{|V_{id}|}$ where
	    \par
	    \hskip\algorithmicindent
	    $S_O$ is \textit{the maximum number of objects in all the frames}  
	    \par
	    \hskip\algorithmicindent
	    and $V_{id}$ is \textit{the set of the object variables in the scope of} $\varphi_k$
	    \State $S_O^{cf} \leftarrow$ \textit{number of objects in the current frame $u$}
	    \State $S_O^{ff} \leftarrow$ \textit{number of the objects in the freeze frame $t$}
	    \State $F_{ID} \leftarrow$ is a map s.t.   
	    \par
	    \hskip\algorithmicindent
	    {$ \forall Id \in {V_{id}}, \{F_{ID}[Id] = true \;|\; \exists \; 
	    ``\exists Id @ v_x." \in \varphi$} 
	    \par
	    \hskip\algorithmicindent
	    or $ ``\forall Id @ v_x." \in \varphi\}, v_x \in V_t 
	    \}$ 
		\For{$i\gets 1 \;{ to }\; S_{id}$}\\ \Comment{iterates over combinatorial assignments of values to the IDs}
		\State {$\{Id_1,\dots Id_{|V_{id}|}\} \leftarrow$ i'th  combinatorial of $\{1,\dots S_O\}$} 
		\par
		\hskip\algorithmicindent
		to $\{Id_j \;|\; Id_j\in V_{id}, 1 \le j \le |V_{id}|\}$
		\If { {$\exists Id \in V_{id}$ and \big(($F_{ID}[Id]=true$ and $Id > S_O^{ff})$ or} 
	    \par
	    \hskip\algorithmicindent
		{$(F_{ID}[Id]=false$ and $Id > S_O^{cf})\big)$}}\label{alg3:if-id-exists}
        \State {$M[k,u_{01}].IT[i-1] \leftarrow \bot$}
		\Else
		\For{$j\gets 1 \;{ to }\; n$} \Comment{iterates over $Fn$ arguments}
		\If {$\Omega_j$ is a const number or reserved word}
		\State $P[j].val \leftarrow \Omega_j$
		\ElsIf{$\Omega_j \equiv Fn(\dots)$}
		\State {$P[j].val \leftarrow$} 
		\par
		\hskip\algorithmicindent
		\hskip\algorithmicindent
		\hskip\algorithmicindent
		$ComputeFnExpr(\Omega_j,k,u,u_{01},t,\hat{\rho})$
		\ElsIf {$\Omega_j \equiv \BB(Id)$}
		\State $P[j].region \leftarrow BoundingBox(Id,F_{ID},u,t)$
		\ElsIf {$\Omega_j$ is a spatial formula}
		\State {$P[j].region \leftarrow $}
		\par
		\hskip\algorithmicindent
		\hskip\algorithmicindent
		\hskip\algorithmicindent
		{$ComputeSpatioTemporal(\Omega_j,u,t,i,\hat{\rho})$}
		\EndIf
		\EndFor
        \State {$M[k,u_{01}].IT[i-1] \leftarrow Fn(P) \sim r$}
		\EndIf
		\EndFor
	    \State {$S_{ID} \leftarrow (S_O)^{L}$ where}
	    \par
	    \hskip\algorithmicindent
	    {$L$ is \textit{the maximum number of object variables that can be used}}
	    \par
	    \hskip\algorithmicindent
	    {\textit{in the scope of any } $@\{\dots\}$ subformula}
		\For{$i\gets S_{id} \;{ to }\; S_{ID}-1$} \Comment{iterates over the remaining of the ID Table}
        \State {$M[k,u_{01}].IT[i] \leftarrow \nan$}
		\EndFor
	\end{algorithmic}
	\hspace {5pt}{\bf   end procedure}
	}
\end{algorithm}
\color{black}

\begin{algorithm}[t]
\footnotesize{
	\caption{ Spatio-Temporal Monitor (with intervals)}
	{\bf Input}: $\theta,u',t,i,\hat{\rho}$;
	\label{alg:DP-S4U-INT}

	\hspace {10pt}{\bf   procedure  }{\sc ComputeSpatioTemporal}($\theta,u',t,i,\hat{\rho}$)
	\begin{algorithmic}[1]
		\State {$\{Id_1,\dots Id_{|V_{id}|}\} \leftarrow i$'th  combinatorial assignment of $\{1,\dots S_O\}$ to} 
		\par
		\hskip\algorithmicindent
		object variables
		\For{$u\gets |\hat{\rho}|-1\mbox{ down to }u'$}\Comment{iterates backward over time stamps}
	    \State {$S_O^{cf} \leftarrow$ \textit{number of objects in the current frame $u$}}
	    \State $S_O^{ff} \leftarrow$ \textit{number of the objects in the freeze frame $t$}
	    \For{$j\gets \theta.max\mbox{ down to }\theta.min$} \Comment{iterates over subformulas}
		\If{$\varphi_j\equiv \BB(Id)$}
		\If { {$\big((F_{ID}[Id]=true$ and $Id > S_O^{ff})$ or} 
	    \par
	    \hskip\algorithmicindent
		{$(F_{ID}[Id]=false$ and $Id > S_O^{cf})\big)$}}
        \State {$\overline{M}[j,u] \leftarrow \Sempty$} \Comment{mark as NaN}
		\Else
		\State {$\overline{M}[j,u]\leftarrow BoundingBox(Id,F_{ID},u,t)$}
		\EndIf
		\ElsIf{$\varphi_j\equiv\overline{\varphi_m}$}
		\State $\overline{M}[j,u]\leftarrow Complement(\overline{M}[m,u])$
		\ElsIf {$\varphi_j\equiv\varphi_m\sqcap\varphi_n$}
		\State $\overline{M}[j,u]\leftarrow Intersection(\overline{M}[m,u], \overline{M}[n,u])$
		\ElsIf {$\varphi_j\equiv\varphi_m\sqcup\varphi_n$}
		\State $\overline{M}[j,u]\leftarrow Union(\overline{M}[m,u], \overline{M}[n,u])$
		\ElsIf {$\varphi_j\equiv \Itl \; {\varphi_m}$}
		\State $\overline{M}[j,u]\leftarrow Interior(\overline{M}[m,u])$
		\ElsIf {$\varphi_j\equiv \Csl \; {\varphi_m}$}
		\State $\overline{M}[j,u]\leftarrow Closure(\overline{M}[m,u])$
		\ElsIf {$\varphi_j\equiv\varphi_m \;\tilde{\Un}_{\Ic}^{s}\;\varphi_n$}
		\If{$u=|\hat{\rho}|-1$ and $0 \in \Ic$}
		\State $\overline{M}[j,u]\leftarrow \overline{M}[n,u]$
		\ElsIf{$u=|\hat{\rho}|-1$ and $0 \not \in \Ic$}
		\State $\overline{M}[j,u]\leftarrow \bot$
		\ElsIf{$\Ic = [0,+\infty)$}
		\State 
		$\overline{M}[j,u]\leftarrow \overline{M}[n,u]\;\sqcup$ 
		\par
		\hskip\algorithmicindent
		$(\overline{M}[m,u]\sqcap \overline{M}[j,u+1])$
		\Else
		\State $b_l \leftarrow min\{j+\Ic\}$; $b_u \leftarrow max\{j+\Ic\}$
		\State $r_{min} \leftarrow \sqcap_{j\le j' < b_l}$ $\overline{M}[m,j']$
		\State $\overline{M}[j,u] \leftarrow \bot$
		\For {$j' \leftarrow b_l$ to $b_u$} \Comment{iterates over frame indices}
		\State $\overline{M}[j,j'] \leftarrow \overline{M}[j,j'] \sqcup (\overline{M}[n,j'] \sqcap r_{min})$
		\State $r_{min} \leftarrow r_{min} \sqcap \overline{M}[m,j']$
		\EndFor
		\If {$sup\; \Ic = +\infty$}
		\State 
		$\overline{M}[j,u]\leftarrow \overline{M}[j,u]$ 
		$\sqcup(\overline{M}[m,u]\sqcap \overline{M}[j,u+1])$
		\EndIf
		\EndIf
		\EndIf
		\EndFor
		\EndFor	
		\State \Return {$\overline{M}[\theta.min,u']$}
	\end{algorithmic}
	\hspace {5pt}{\bf   end procedure}
	}
\end{algorithm}
\color{black}

\subsection{Correctness and Complexity Analysis}\label{sec:complexity}
Let $\hat{\rho}$ be an input signal of size $|\hat{\rho}|$,
$\varphi$ be an AAN STPL formula of size $|\varphi|$ (note that size of a formula is the summation of the number of spatial, temporal and spatio-temporal subformulas),
$|V_t|$ be the size of freeze time variables (or 1 if there is none),
$S_{id}$ be the maximum number of used object variables in the scope of a quantifier operator, and
$S_{obj}$ be the maximum number of objects in a single frame in $\hat{\rho}$.
If $\varphi$ is a TPTL formula, then it is known from \cite{dokhanchi2016efficient} that the upper bound time complexity for the variable-bounded TPTL monitoring algorithm is
$O(|V_t| \times |\varphi| \times |\hat{\rho}|^2)$, which is polynomial.
Additionally, the upper bound space complexity of the presented algorithm by \cite{dokhanchi2016efficient} is $O(|\varphi| \times |\hat{\rho}|)$.
Our STPL monitoring algorithm is founded based on the TPTL monitoring algorithm, but
there are two major additions to the TPTL syntax and semantics.
\revision{
The first addition to the STPL grammar presented in Def. \ref{lbl:stql-brief-syntax} is the  existential/universal quantifiers that precedes the freeze time operator.
The second addition is the production rule $\Theta$.
}
We call this extension of the TPTL language the Quantifiable TPTL (QTPTL).  
Next is the production rule $\Stau$ that produces purely spatio-temporal formulas ($\Pi$ and $\Sexists\;\Stau$ quantify the spatio-temporal in a constant time).
The last addition to the QTPTL results in the STPL language.
Therefore, we only analyze the time and space complexity of our algorithm for parts that concern the two aforementioned additions.

Our proposed monitoring algorithm is based on the Dynamic Programming (DP) algorithm by \cite{dokhanchi2016efficient} and, therefore, to consider the first addition, we need to evaluate each object variable related subformula at most $(S_{obj})^{S_{id}}$ times.
This also requires extra space to build the DP tables.
Therefore, the upper bound time and space complexity of the QTPTL algorithms increases to $O((S_{obj})^{S_{id}} \times |V_t| \times |\varphi| \times |\hat{\rho}|^2)$ and $O((S_{obj})^{S_{id}} \times |\varphi| \times |\hat{\rho}|)$, respectively.

Finally, we consider spatial and spatio-temporal subformulas denoted as $\varphi_s$ in addition to temporal ones denoted as $\varphi_t$ to do complexity analysis of the STPL algorithm.
It is easy to see that the production rule $\Stau$ has the same grammar as MTL/STL, except that the logical operators are replaced with spatial ones.
Therefore, the time/space complexity of monitoring these formulas follows the same complexity as in MTL/STL monitoring algorithms except for the spatial operations.
In MTL/STL, all the logical operations compute in constant time.
However, for spatial operations, depending on the used data structure for representing the spatial terms, this might not hold.
In \nameref{appendix:complexity}, we calculate exponential lower bound for some spatio-temporal formulas where the linked-list was used to represent the spatial terms.
That is, if we do not exploit the geometrical properties of the spatial terms while storing them, then we get exponential complexity for the spatial operations. 
Whereas, if we use some geometry-sensitive data structures such as \textit{region Quadtrees} and \textit{region Octrees} for storing and computing 2D and 3D spatial terms (e.g., see \cite{aluru2018quadtrees,shneier1981calculations}), respectively, then, we get a polynomial time and space complexity, e.g., \cite{bournez1999orthogonal}. 
Assume that we can decompose our \textit{d-}dimensional topological space (dimension is fixed to be 2D or 3D) into $r^{d}$ cells, \revision{where $r$ is the constant resolution along each axis.
Construction of an image Quadtree/Octree is linear in the size of the image (see \cite{shneier1981calculations}).}
The union and intersection algorithm for Quadtrees, in the worst case, requires visiting all the nodes in the trees, which can be done in $K$ times.
For computing a spatio-temporal formula $\varphi_s$, at each time-step, in the worst case, it requires as many spatial operations as linear to the size of the formula.

Therefore, the time complexity of computing the formula $\varphi_s$ against an input signal $\hat{\rho}$ is as follows:
\begin{itemize}
    \item $O(|\hat{\rho}| \times |\varphi_s| \times K)$, if there is no time/frame intervals in the formula and no frozen object variable is used in the formula.
    \hlRev{Note that $K$ is a big constant (i.e., $K$ is a function of the dimension and the resolution of the space) and we did not omit it to emphasize its impact on the computation.}
    \item $O(|\hat{\rho}| \times |\varphi_s| \times K \times c)$, if there are time/frame intervals in the formula and no frozen object variable is used in the formula. 
    \revision{
    Here, $c$ is defined as
    \[ c = \max_{0\leq j \leq |\Data|, \Ic \in T(\phi)} |[j,\max J(j,\Ic)]| \]
    where $T(\phi)$ contains all the timing constraints $\Ic$ which are either attached on the temporal operators in formula $\phi$, or are timing constraints of the form $ \Ctime - x \sim n$ or  $\Cframe - x \sim n$ in the scope of a temporal operator in $\phi$.
    For example, if $\phi = \Box_{[0.1, \infty)} \psi$, then $[0.1, \infty) \in T(\phi)$, or if $\phi = x \; . \, \Box ( \tau-x\leq 5.5 \Rightarrow \psi)$, then $[0, 5.5] \in T(\phi)$.
    Intuitively, the function $J$ returns all the samples that satisfy a constraint $\Ic$ from the set $T(\phi)$ starting from a sample $j$.
    Formally,
    \begin{equation*}
        J(j,\Ic) = \left \{
        \begin{array}{ll}
            \tau^{-1}((\tau(j)+_R \Ic)\cap \\
            \qquad (\tau(j+1)+_R \Ic)) & \text{ if } \sup \Ic = +\infty \\
            \tau^{-1}(\tau(j)+_R \Ic) & \text{ otherwise }
        \end{array} \right.
    \end{equation*}
    with  $t+_R\Ic = \{ t'' \in R \; | \; \exists t' \in \Ic \, .\, t'' = t + t' \}$.
    Note that when considering constraints on the number of frames, i.e., $\Cframe - x \sim n$, then the timestamp mapping $\tau$ is the identity function.  
    For a discussion on $c$ for  STL/MTL with discrete time semantics see \cite{fainekos2012verification}.
    }
    \item $O(|\hat{\rho}| \times |\varphi_s| \times K \times |V_t|)$, if there is no time/frame intervals in the formula, but there are frozen object variables used in the formula.
    \item $O(|\hat{\rho}| \times |\varphi_s| \times K \times c \times |V_t|)$, if there are time/frame intervals in the formula, and there are frozen object variables used in the formula.    
\end{itemize}

Overall, the upper bound time and space complexity of the STPL algorithm are $O((S_{obj})^{S_{id}} \times |V_t| \times |\varphi| \times |\hat{\rho}|^2 \times K \times c)$ and $O((S_{obj})^{S_{id}} \times |\varphi| \times |\hat{\rho}| \times K)$, respectively.

\hlRev{In our implementation of the monitoring algorithms (discussed in \nameref{lbl:apx:algorithm}), we focus on the future fragment of the STPL logic to avoid complexity (e.g, by excluding past-time operators as presented in \nameref{lbl:apx:grammar}).} 
That is, we improved the space complexity of STPL formulas without spatial terms by decoupling the DP tables into two separate tables: one dedicated to the values of subformulas at the current time step, and the other for their frozen values along the time horizon.
Therefore, we reduced the space complexity to
$O\big((S_{obj})^{S_{id}} \times (|\varphi_t| + |\hat{\rho}|)\big)$ 
for non-spatial STPL formulas and spatial formulas without time/frame intervals in them.
For improving the exponential complexity of the spatial STPL formulas, we merge the fragmented subsets after spatial operations.
Additionally, if there is no frozen object variable used in a spatial formula, we only evaluate it once.
Some optimizations can be done based on the content of the formulas, for example, if a temporal formula does not have \textit{globally, eventually, until} and \textit{release} operators in it, then we deduce the needed horizon length of the input signal accordingly (\hlRev{i.e., next time operator only requires the evaluation of the first two frames}).
Also, we interpret the time/frame constraints in a formula to possibly ignore the evaluation of the affected subformulas accordingly. 

The correctness of the algorithm with respect to the presented syntax and semantics of STPL can be proven by using the correctness proofs that are presented for the TPTL and MTL monitoring algorithms by \cite{dokhanchi2016efficient} and \cite{fainekos2012verification}.

\revision{
We have released an open-source version of the code for both Linux and Windows OS in a public repository in GitLab (see \cite{stpl_tools}).
}
Our tool can be run in standalone mode or as part of Matlab.
Additionally, there are data stream files and input configuration files that cover most of the examples in the previous sections, as well as the sensitivity analysis result in the following section.

\revision{
\subsection{A Polynomial Time/Space Fragment}\label{sec:poly-frag}
}

\revision{In this section, we identify STPL specification templates for which the monitoring problem becomes  polynomial time in the worst case.}
In the template below, we assume that the size of the spatial formula $\varphi_s$ is bounded.

\subsubsection{Example}
The complexities of evaluating an arbitrary SPE formula $\varphi_s$ (the formula is in conjunctive form) on a data stream $\hat{\rho}$ is
\begin{itemize}
    \item $O\big(2^{a} 3^{b}\big)$ for time and space, if there is no complement operator in the formula, where $a,b \in \mathbb{N}$, and we have:
    $\arg\max \big( 2^a 3^b \;|\;$$(|\varphi_s|+1)/2 = 2a + 3b\big)$.
    For instance, $\varphi_s := (\Stau \sqcup \Stau \sqcup \Stau)$$\;\sqcap\; (\Stau \sqcup \Stau \sqcup \Stau)$.
    
    \item $O\big(4^{((|\varphi_s|+1)/3)}\big)$ for time and space, if there are complement operators in the NNF formula.
    For instance, $\varphi_s := \overline{\Stau} \;\sqcap\; \overline{\Stau}$$\;\sqcap\; \overline{\Stau} \;\sqcap\; \overline{\Stau}$$\;\sqcap\; \overline{\Stau} \;\sqcap\; \overline{\Stau}$.
\end{itemize}

\subsubsection{Example}
An arbitrary $l$-level nested spatial globally formula $\varphi_s^{\Box}$, with a spatial term as its right-most subformula, is of $O(|\hat{\rho}|)$ time and $O(1)$ space complexity.
\[
\varphi_s^{\Box} := \Box^{s}_{\Ic}\Big( \Stau \sqcap \Box^{s}_{\Ic} \big( \Stau \sqcap \Box^{s}_{\Ic} (\Stau \sqcap \Box^{s}_{\Ic} \Stau) \big)\Big)
\]

\subsubsection{Example}
An arbitrary $l$-level nested spatial eventually formula $\varphi_s^{\Diamond}$, with a spatial term as its right-most subformula, is of $O(|\hat{\rho}|^{(l+1)})$ time and space complexity.
\[
\varphi_s^{\Diamond} := \Diamond^{s}_{\Ic}\Big( \Stau \sqcup \Diamond^{s}_{\Ic} \big( \Stau \sqcup \Diamond^{s}_{\Ic} (\Stau \sqcup \Diamond^{s}_{\Ic} \Stau) \big)\Big)
\]

\subsubsection{Example}
An arbitrary $l$-level nested spatial eventually subformula $\varphi_s^{\Diamond}$ followed by an arbitrary $k$-level nested spatial globally subformula $\varphi_s^{\Box}$, with a spatial term as its right-most subformula, is of $O(|\hat{\rho}|^{(l+1)})$ time and space complexity.
\begin{gather*}
\varphi_s^{\Diamond,\Box} := \Diamond^{s}_{\Ic}\Big( \Stau \sqcup \Diamond^{s}_{\Ic} \big( \Stau \sqcup \Diamond^{s}_{\Ic} (\Stau \sqcup \Diamond^{s}_{\Ic} 
\\
\Big(\Box^{s}_{\Ic} \big( \Stau \sqcap \Box^{s}_{\Ic} (\Stau \sqcap \Box^{s}_{\Ic} \Stau) \big)\Big) ) \big)\Big)
\end{gather*}

\subsubsection{Example}
A right-hand-side $l$-level nested spatial until formula $\varphi_s^{\Un}$, where the left-hand-side of all the until operators are a single spatial term, is of $O(|\hat{\rho}|^{(l+2)})$.
\begin{gather*}
\varphi_s^{\Un} := \Stau \;\Un^{s}_{\Ic} 
\Big( \Stau \;\Un^{s}_{\Ic} \big( \Stau \;\Un^{s}_{\Ic} (\Stau \;\Un^{s}_{\Ic}\; \Stau ) \big) \Big)
\end{gather*}

\subsubsection{Example}
A left-hand-side $l$-level nested spatial release formula $\varphi_s^{\Rc}$, where the right-hand-side of all the until operators are a single spatial term, is of $O(|\hat{\rho}|^{(l+2)})$.
\begin{gather*}
\varphi_s^{\Rc} :=  
\Big(\big( ( \Stau \;\Rc^{s}_{\Ic}\; \Stau) \Rc^{s}_{\Ic}\; \Stau \big) \Rc^{s}_{\Ic}\; \Stau \Big) \Rc^{s}_{\Ic}\; \Stau
\end{gather*}

\section{Experiments and Results}
\label{sec:exp}

\begin{table}[tbp]
\vspace{6pt}
\begin{center}
\caption{Statistics on execution-time for different formulas and data stream sizes. We used the Berkeley DeepDrive (BDD) dataset to compute the results. 
$m-time$ and $e-time$ represent the required time (in second) for releasing memories and executing the monitoring algorithm, respectively.
}\label{tbl:experiment-stats}
\resizebox{.45\textwidth}{!}{
\renewcommand{\arraystretch}{1.3}
\begin{tabular}{ ?{.4mm} c | c | c | c | c | c | c | c ?{.4mm} } 
\Xhline{3\arrayrulewidth}
$\boldsymbol{|\hat{\rho}|}$ & $\boldsymbol{|\varphi_t|}$ &$\boldsymbol{|\varphi_s|}$ & $\boldsymbol{|V_t|}$ & $\boldsymbol{S_{obj}}$ & $\boldsymbol{S_{id}}$ & \textbf{m-time} & \textbf{e-time}\\
\Xhline{3\arrayrulewidth}
\multicolumn{8}{c}{ 
\textit{quantifier-formed} STPL Formula (\ref{lbl:ex:6}) without spatial operators}\\
\Xhline{3\arrayrulewidth}
\textbf{25} & 9 & 0 & 0 & 20 & 1 & 0 & \textbf{0.002}\\
\hline
\textbf{50} & 9 & 0 & 0 & 20 & 1 & 0 &\textbf{0.001}\\
\hline
\textbf{100} & 9 & 0 & 0 & 23 & 1 & 0 &\textbf{0.005}\\
\hline
\textbf{200} & 9 & 0 & 0 & 24 & 1 & 0 &\textbf{0.008}\\
\Xhline{3\arrayrulewidth}
\multicolumn{8}{c}{ 
\textit{mix-formed} STPL Formula (\ref{lbl:ex:7}) without spatial operators}\\
\Xhline{3\arrayrulewidth}
\textbf{25} & 7 & 0 & 1 & 20 & 2 & 0 &\textbf{0.132}\\
\hline
\textbf{50} & 7 & 0 & 1 & 20 & 2 & 0 &\textbf{0.519}\\
\hline
\textbf{100} & 7 & 0 & 1 & 23 & 2 & 0 &\textbf{2.76}\\
\hline
\textbf{200} & 7 & 0 & 1 & 24 & 2 & 0 &\textbf{11.31}\\
\Xhline{3\arrayrulewidth}
\multicolumn{8}{c}{ mix-formed STPL Formula (\ref{ex:3:a}) with SPE operators}\\
\Xhline{3\arrayrulewidth}
\textbf{25} & 9 & 8 & 1 & 20 & 2 & 0.004 &\textbf{1.25}\\
\hline
\textbf{50} & 9 & 8 & 1 & 20 & 2 & 0.003 &\textbf{4.13}\\
\hline
\textbf{100} & 9 & 8 & 1 & 23 & 2 & 0.005 &\textbf{16.32}\\
\hline
\textbf{200} & 9 & 8 & 1 & 24 & 2 & 0.005 &\textbf{63.52}\\
\Xhline{3\arrayrulewidth}
\multicolumn{8}{c}{ quantifier-formed STPL Formula (\ref{ex:3:b}) with STE operators}\\
\Xhline{3\arrayrulewidth}
\textbf{25} & 3 & 4 & 0 & 20 & 1 & 0 &\textbf{0.006}\\
\hline
\textbf{50} & 3 & 4 & 0 & 20 & 1 & 0 &\textbf{0.029}\\
\hline
\textbf{100} & 3 & 4 & 0 & 23 & 1 & 0 &\textbf{0.119}\\
\hline
\textbf{200} & 3 & 4 & 0 & 24 & 1 & 0 &\textbf{0.176}\\
\Xhline{3\arrayrulewidth}
\multicolumn{8}{c}{ 
\textit{mix-formed} STPL Formula (\ref{ex:4:b}) with spatial terms}\\
\Xhline{3\arrayrulewidth}
\textbf{25} & 29 & \revision{6} & 1 & 20 & 3 & 0.023 &\textbf{2.23}\\
\hline
\textbf{50} & 29 & \revision{6} & 1 & 20 & 3 & 0.023 &\textbf{4.97}\\
\hline
\textbf{100} & 29 & \revision{6} & 1 & 23 & 3 & 0.037 &\textbf{16.07}\\
\hline
\textbf{200} & 29 & \revision{6} & 1 & 24 & 3 & 0.043 &\textbf{46.88}\\
\Xhline{3\arrayrulewidth}
\multicolumn{8}{c}{ 
\textit{mix-formed} STPL Formula (\ref{ex:4:c}) without spatial terms}\\
\Xhline{3\arrayrulewidth}
\textbf{25} & 29 & 0 & 1 & 20 & 3 & 0 &\textbf{1.16}\\
\hline
\textbf{50} & 29 & 0 & 1 & 20 & 3 & 0 &\textbf{2.73}\\
\hline
\textbf{100} & 29 & 0 & 1 & 23 & 3 & 0 &\textbf{10.41}\\
\hline
\textbf{200} & 29 & 0 & 1 & 24 & 3 & 0 &\textbf{33.69}\\
\Xhline{3\arrayrulewidth}
\end{tabular}
}
\end{center}
\end{table}


We selected some of the presented example formulas to cover different possible 
combinations for the operators and quantifiers, and to demonstrate how the 
computation time scales concerning the size of the data stream and the formulas.
For this experimental analysis, we used the  DeepDrive dataset \footnote{\url{https://bdd-data.berkeley.edu/}} (\cite{yu2020bdd100k}). 
As indicated in the last column of Table \ref{tbl:experiment-stats}, the performance of the STPL monitoring algorithm for hundreds of monitoring frames (including thousands of objects) is feasible for offline monitoring.
Some statistics about the experiments are summarized in Table \ref{tbl:experiment-stats}.
We used a Windows 10 machine with Intel Core i7 CPU $8550$U $@$ $1.8$GHZ, $16$GB RAM, and Matlab R2020a.
The STPL monitoring algorithm is implemented in $C$ language and compiled for Matlab.

\revision{
The formulas in Table \ref{tbl:experiment-stats} were selected based on the types of operators and their computational complexity.
The maximum number of nested quantifiers is a source of exponential complexity (i.e., $(S_{obj})^{S_{id}}$) in our monitoring algorithm.
The highest number of nested quantification operators was 3 in Formulas (\ref{ex:4:b}) and (\ref{ex:4:c}).
Another source of complexity is the number of spatial operators in the formula.
For instance, the worst execution time is observed for the Formula (\ref{ex:3:a}) due to its higher number of spatial operators.
These results are not surprising and are in agreement with our theoretical analysis in Section \ref{sec:complexity}.
}

\revision{
In order to study the impact of the number of objects in each frame on the monitoring algorithm, we manipulated the BDD dataset to create artificial datasets with specific number of objects. 
The result is presented in Table \ref{tbl:runtime-obj-id}.
In the first row, the maximum number of objects in all the frames is 5. For all the following rows, this number is doubled.
Based on the $O((S_{obj})^{S_{id}})$, the \textit{e-time} of rows should increase with the ratio of $2^3=8$.
To remedy the exponential complexity of the nested quantified formulas, we can use parallel computation to efficiently evaluate quantified subformulas.
More specifically, Algorithm \ref{alg:DP-FNEXPR} can be parallelized. 
}

\revision{
Since the theoretical time complexity of offline STPL monitoring is the same as the time complexity of online past-time STPL monitoring (see \cite{BalakrishnanEtAl2021rv} for a toolbox), the problem of online monitoring of bounded time short duration STPL properties is practically feasible and relevant.
However, the offline STPL monitoring problem over extremely large perception datasets may not be feasible without further optimizations or expressivity restrictions.
One promising direction is to prefilter very large perception datasets and extract subsequences that are interesting for STPL specifications.
One such possibility for filtering is the perception query language SpRE (\cite{AndersonEtAl2023rv}) -- see Section \ref{sec:related} for a brief discussion.
We plan to pursue such a direction in the future.
}

        \begin{table}
        \begin{center}
        \caption{
        Statistics on execution time for the Formula (\ref{ex:4:b}) on an artificial perception data stream.
        }\label{tbl:runtime-obj-id}
        \resizebox{.48\textwidth}{!}{
        \renewcommand{\arraystretch}{1.3}
        \begin{tabular}{ ?{.4mm} c | c | c | c | c | c | c | c ?{.4mm} } 
        \Xhline{3\arrayrulewidth}
        $\boldsymbol{|\hat{\rho}|}$ & $\boldsymbol{|\varphi_t|}$ &$\boldsymbol{|\varphi_s|}$ & $\boldsymbol{|V_t|}$ & $\boldsymbol{S_{obj}}$ & $\boldsymbol{S_{id}}$ & \textbf{m-time} & \textbf{e-time}\\
        \Xhline{3\arrayrulewidth}
        \textbf{25} & 29 & 6 & 1 & 5 & 3 & 0.001 & \textbf{0.060}\\
        \hline
        \textbf{25} & 29 & 6 & 1 & 10 & 3 & 0.005 &\textbf{0.441}\\
        \hline
        \textbf{25} & 29 & 6 & 1 & 20 & 3 & 0.043 &\textbf{2.419}\\
        \hline
        \textbf{25} & 29 & 6 & 1 & 40 & 3 & 0.312 &\textbf{20.027}\\
        \Xhline{3\arrayrulewidth}
        \end{tabular}
        }
        
        \end{center}
        \end{table}


\section{Related Works}
\label{sec:related}

        \begin{table*}[h]
        \begin{center}
        \caption{
\revision{
        Overview comparison of spatio-temporal formal languages. 
        The $\exists \forall$ (data) column indicates whether the logic supports quantification over non-spatial data.
        The $\exists \forall$ (spatial column indicates whether the logic supports some form of spatial quantification, e.g., emptiness operator (i.e., there exists a point), or directional operator (there exists a direction).
        FSM: Finite state machines.
}
        }\label{tbl:logic-comp}
        \resizebox{.98\textwidth}{!}{
        \renewcommand{\arraystretch}{1.5}
        \begin{tabular}{   |c | c |  c | c  c  c | c | } 
\Xhline{3\arrayrulewidth}
        \textbf{Language} & \makecell{\textbf{Temporal} \\ \textbf{Foundation}} & \textbf{$\exists\forall$ (data) } & \textbf{$\exists\forall$ (spatial) } & \makecell{\textbf{Spatial} \\ \textbf{Foundation}} & \makecell{\textbf{Domain} \\ \textbf{in practice}} & \textbf{Applications}\\
\Xhline{3\arrayrulewidth}
        \textbf{STPL} & AAN-TPTL & $\checkmark$   & $\checkmark$ & $S4_u$ & \makecell{Sets in \\ Euclidean spaces} & \makecell{Perception systems} \\
        \hline
        \textbf{TQTL} & AAN-TPTL & $\checkmark$ & & \makecell{Predicates} & \makecell{Perception data} & \makecell{Vision based \\ (2D) perception}\\
        \hline
        \textbf{SpRE} & FSM &    & $\checkmark$ & $S4_u$ & \makecell{Sets in \\ Euclidean spaces} & \makecell{Perception systems} \\
        \hline
        \textbf{STSL} & STL &  &  $\checkmark$ & $S4_u$ & \makecell{Distances in \\ Euclidean spaces}  & \makecell{System level \\ requirements for CPS } \\
        \hline
        \textbf{SpaTel} & STL &  & $\checkmark$ & TSSL & Quadrants & Pattern recognition \\
        \hline
        \textbf{SSTL} & STL &  & $\checkmark$ & \makecell{Graph-based\\ modeling} & \makecell{Distances in \\ Euclidean spaces} & Pattern recognition \\
        \hline
        \textbf{GSTL} & STL &  & $\checkmark$ & Mereotopology & Cubics & \makecell{Knowledge\\ representation} \\
        \hline
        \end{tabular}
        }
        
        \end{center}
        \end{table*}

\revision{
In this section, we provide a detailed overview of related works.
We primarily focus on comparing STPL with other spatio-temporal logics.
In Table \ref{tbl:logic-comp}, we provide a summary comparison of the most relevant logics for easy reference.
We conclude the section with some references on promising directions on the analysis of perception systems which are not directly related to spatio-temporal logics.
}

Region Connection Calculus by \cite{CohnBGG1997GeoInformatica}, Shape Calculus by \cite{BartocciCBMT2010sacs}, and Situation Calculus by \cite{BhattL2008scc} are just some examples of the vast literature on logics about topology and space.
Comprehensive surveys and comparisons of reasoning methods about topology and time are provided by \cite{DyllaEtAl2017csur,AielloPHB2007handbook}. 
Spatio-temporal logics and calculi are also frequently used in robotics for human-robot communication (\cite{KressGazitP2010icra,SummersStayCV2014wvl}) and for specifying and verifying properties of AV (\cite{LinkerH2013ictac,LoosPN11fm}).
All the aforementioned works that deal with topology and time primarily focus on deductive reasoning, theorem proving, knowledge representation, axiomatization, and -- in some cases -- planning.
In contrast, STPL requires computationally efficient tools for spatio-temporal reasoning in order to monitor data streams from perception systems. 

Even though there exist spatio-temporal logics that can process spatio-temporal data (offline or online) such as SpaTeL (\cite{haghighi2015spatel}), SSTL (\cite{nenzi2015qualitative}), or SaSTL (\cite{MaBLSF2020iccps}) (for a short survey see \cite{BartocciEtAl2018survey}), or even images, e.g., SLCS (\cite{buonamici2019spatial}), all these logics are application dependent and cannot support the topological reasoning needed for perception data in AV.
To highlight the fundamental differences between the aforementioned logics and STPL, we provide a detailed comparison with SpaTeL and SSTL.
The differences with the other listed logics and conceptually similar in scope.
The two spatio-temporal languages (SSTL and SpaTeL) are explicitly developed for describing high-level spatial patterns that evolve.
Both languages are founded based on a graph-based representation of discrete models of space. 
For the SSTL, undirected weighted graphs are used to model space, and in SpaTeL, a networked system whose states encapsulate an image are represented as quad transition systems. 

In more detail, in SSTL, the syntax of the language adds two spatial operators  $\diamonddiamond_{[w_1,w_2]} \varphi$ and $\varphi_1 \mathcal{S}_{[w_1,w_2]} \varphi_2$ into Signal Temporal Logic (STL) (\cite{MalerN04formats}), which are named bounded somewhere and bounded surround, respectively. 
The first operator requires $\varphi$ to hold in a location that can be reached from the current location with a cost between $w_1$ and $w_2$. 
The cost is usually the distance between the two locations.
For the second operator, the notation of external boundary of a set is required. 
An external boundary of a given set of nodes is defined as the set of nodes that are directly connected to the elements of the given set but are not members of it. 
The semantics of the second operator requires that for the current location $l$ and a given trace $x$, $l$ belongs to a set of locations $A$ that all satisfy the formula $\varphi_1$, and for all the locations in the external boundary of $A$, they satisfy the formula $\varphi_2$. 
An SSTL formula can be arbitrarily nested.

In SpaTeL, a combination of Tree Spatial Superposition Logic (TSSL) (\cite{gol2014formal}) and STL is proposed to reason on spatial patterns. 
TSSL uses quad-trees to represent the space by partitioning the space recursively into quadrants. 
The TSSL logic is similar to the classic Computation Tree Logic (CTL) (e.g., see \cite{huth2004logic}), with the main difference that the next and until operators are not temporal, but spatial. 
That is, evolution happens by a change of resolution (or zoom in). 
All the spatial operators are augmented by a set $B$ that selects the spatial directions (i.e., NW, NE, SW, and SE) in which the operators are allowed to work. 
Additionally, similar to temporal operators, there is a parameter $k$ that limits the operator’s evaluation range on a finite sequence of states. 
For example, $\exists_B \varphi_1  U_k  \varphi_2$ means that there exists a set of directions in $B$ by which the $i$-th label of a path $\pi^B$ 
satisfies the formula $\varphi_2$ and all the other labels on the path until $i$ satisfy the formula $\varphi_1$. 
A difference between the former and the latter is that in the former one, the TSSL fragment of a formula does not include temporal subformulas.

In summary, the key differences of these logics with our proposed STPL logic are:
\begin{itemize}
    \item 
	\textit{Modeling}: 
	SSTL and SpaTeL are not designed to model physical objects in 2D or 3D spaces. 
	On the other hand, our logic is explicitly designed to handle physical objects.

	\item
	\textit{Expressivity}: SpaTeL is inherently less expressive than SSTL due to its modeling and traversing approach on quad-trees and the decoupled syntax for spatial and temporal formulas. 
	Therefore, we are going to compare STPL with SSTL. 
	There are two significant differences. 
	The first one is the presence of time freeze operator and time variables and, hence, STPL is more expressive. 
	The second one is the presence of the quantifiers and set operations
	over spatial terms/locations. 
	As an example, SSTL cannot reason on whether the same object over two different frames overlaps or not with itself. 

	\item
	\textit{Application}: SSTL and SpaTeL are mostly helpful for pattern recognition purposes, while STPL is a more general-purpose language.
	Quantitative semantics: there is quantitative and qualitative semantics for SSTL and SpaTeL, but currently, we only presented qualitative semantics for STPL. 
	The graph-based modeling of the spatial environment and the fixed metric properties such as distance makes it more straightforward to define quantitative semantics for their underlying logic.
\end{itemize}

In another line of work, a graph-based spatio-temporal logic -- GSTL by \cite{liu2020graph} -- is presented for knowledge representation and reasoning.
GSTL deals with spatial elements as regions, and uses \textit{mereotopology} (combination of \textit{mereology} and \textit{topology} to support parthood and connectivity, respectively) to represent relations between spatial elements.
It exploits rectangle/cubic algebra to represent spatial objects.
GSTL combines STL temporal logic with mereotopology-based spatial operators enriched with interval algebra.
The satisfiability problem for GSTL is decidable by restricting the evolution of spatial elements.
GSTL was primarily designed for model checking which restricts its expressivity for decidability reasons.

\revision{
The works closest to ours stem from combining temporal logics with spatial logics (for a historical overview and a discussion on ${\mathcal S}4_u$ see \cite{kontchakov2007spatial}).
\cite{gabelaia2005combining} combine Linear Temporal Logic (LTL) (\cite{MannaP92}) with ${\mathcal S}4_u$ to define the logic $\mathcal{PTL}\times{\mathcal S}4_u$.
They further define several fragments of $\mathcal{PTL}\times{\mathcal S}4_u$ and they study the decidability of the satisfiability problem. 
However, the problem of offline monitoring is not investigated in this line of work.
}

\revision{
More recently, STSL was proposed by \cite{LiEtAl2021mna} where STL is combined with ${\mathcal S}4_u$.
Even though the monitoring problem is studied for STSL, STSL falls short of the goals of STPL in multiple directions.
First and foremost, STSL does not support generic data and quantification over such data. 
That is, it is not possible to express a property such as Req. \ref{lbl:req-intro-exm-2} where we need to quantify over the bounding boxes of all the cars in a frame.
Second, STPL is based on TPTL which is a strictly more expressive logic than STL used in STSL. 
Third, a theoretical or experimental computational complexity analysis is not presented for STSL to identify what fragments are computationally important while still being practically relevant.
Finally, and most importantly, the applications presented for STSL are restricted to properties over numerical trajectories produced by Cyber-Physical Systems, and it is clear that with the metric space chosen for STSL, the corresponding formal specifications can be expressed in STL. 
That is, in practice, there is no gain in expressive power in STSL over STL.
}

\revision{
Another line of research relevant to our work is formal languages for analysis of perception systems.
Timed Quality Temporal Logic (TQTL) by \cite{dokhanchi2018evaluating} was designed to reason over streams of perception data.
TQTL is built upon the AAN fragment of Timed Propositional Temporal Logic (AAN-TPTL) (\cite{dokhanchi2016efficient}) by introducing quantification ($\exists$, $\forall$) over the objects in each frame, and by introducing functions that retrieve data relevant to each object, e.g., class, probabilities, bounding box coordinates, etc.
The AAN fragment of TPTL was chosen due to its polynomial time complexity while still being strictly more expressive then STL.
Note that further algorithmic improvements on AAN-TPTL are possible, e.g., see \cite{ElgyuttFH2018formats,GhorbelP2022hscc}.
Nevertheless, TQTL cannot reason directly about properties of bounding boxes.
For example, TQTL cannot reason about self-overlap of bounding boxes across time, i.e., TQTL cannot express Req. \ref{lbl:req-intro-exm-2}.
More generally, TQTL cannot reason about 3D scenes (e.g, bird-eye view of the world) since this requires a mechanism to  relate spatially different objects in the environment.
STPL resolves these shortcomings of TQTL to enable a versatile framework to reason about perception systems in both 2D and 3D (along with other state variables included in the perception data).
}

\revision{
Spatial Regular Expressions (SpREs) were recently introduced by \cite{AndersonEtAl2023rv} to find patterns in streaming data.
That is, given a SpRE, the goal is to find all the sequences of frames that match the pattern specified by SpRE.
SpREs were designed to closely resemble regular expressions, and, hence, the underlying model of computation for processing streaming perception data is automata (\cite{Sipser06}).
The current version of SpRE does not support quantification over data in order to enable online real-time processing.
However, SpRE supports ${\mathcal S}4_u$ operators on per frame basis.
Clearly, SpRE is less expressive than STPL, but we envision an interplay between the two languages. 
SpRE can potentially find very quickly the subsequences of streaming data over which we need to run the more expressive STPL requirements. 
}

\revision{
The PerSyS monitoring system by \cite{antonante2021monitoring} presents a mathematical model for fault detection in perception systems.
The base of their work is  Perfect Minicomputer Corporation (PMC) model in multiprocessor systems, which is generalized to account for models with heterogeneous outputs (i.e., perception systems), and equipped with temporal dimensions to support interaction among PMC models. 
Their system supports consistency checking among different sensory outputs of a perception system with some formal guarantees on the maximum number of inconsistencies. 
In PerSyS, it is possible to design models that identify faults, but, similar to any other graph-based modeling technique, it is highly reliant on a correct model to begin with, and then adding formalized requirements as a set of constraints on the models (i.e., constraints on the edges of the PMC graphs). 
STPL monitoring goals are orthogonal to PerSyS.
STPL is a specification language that formalizes assumptions and guarantees on the functional performance of the perception system.
As such a language, it is more expressive than the constraints used in PerSyS.
As a byproduct, STPL can also function as a comparison framework between different perception stacks.
}

Finally, the language Scenic by \cite{FremontEtAl2018arxiv} has been developed for creating single scene images for testing object detection and classification algorithms.
However, our work is complementary to languages that generate static scenes.

\section{Conclusions}
In this paper, we presented Spatio-Temporal Perception Logic (STPL), which is a logic which is specifically designed for reasoning over data streams of perception systems.
STPL merges and extends other practical logics such as TPTL (\cite{AlurH94acm}) and $S4_u$ (\cite{AielloPHB2007handbook}) with data (object) quantification, functions and relations that enable topological reasoning over time.
Our new logic can be used for specifying correctness requirements on perception systems, as well as to compare different machine learning stacks on their performance beyond the standard metrics (e.g., see \cite{MallickGBD2023rv}). 
\revisionTwo{
We have identified fragments of STPL which are efficiently monitorable for perception applications, and we have demonstrated that practically relevant requirements which do not fall within these fragments can still be efficiently monitorable in practice.
}
An open source publicly available toolbox has been developed \cite{stpl_tools} which can be used for offline perception system analysis.
An online monitor for the past fragment of STPL is also available (\cite{BalakrishnanEtAl2021rv}).
Using STPL, we have been able to discover inconsistencies in publicly available training datasets for AV/ADAS.

\revision{
Since STPL formulas are rather complex even for experts, we have have been working toward developing a Domain Specific Language (DSL) called PyFoReL (\cite{AndersonHF2022re}) for easier elicitation and maintenance of STPL requirements.
PyFoReL provides a Pythonic language to compose requirements in a modular way while enforcing that they are valid STPL formulas.
The next step would be to interface PyFoReL and/or the STPL syntax with Large Language Models (LLM).
Similar work has been done in the past for LTL and STL with practically relevant success (e.g., see \cite{PanCB2023icra,FuggittiC2023aaai}).
In addition, verification and debugging tools for STPL formulas will be needed since LLMs cannot be trusted to always produce correct translations. 
In the past, we have done similar work for STL/MTL specifications in \cite{DokhanchiHF17tecs}. 
We expect to achieve further computational improvements on our monitoring algorithms by parallelization and by filtering relevant sequences of data through our new query language SpRE (\cite{AndersonEtAl2023rv}) before the STPL tools are used.
}
Finally, it would also be interesting to see if meaningful robust semantics (\cite{BartocciEtAl2018survey}) could be defined in order to support test case generation or even self-supervised training of neural networks.

\ifCLASSOPTIONcompsoc
  \section*{Acknowledgments}
\else
  \section*{Acknowledgment}
\fi
This work was partially supported by NSF under grants CNS-2039087, CNS-2038666, IIP-1361926, and the NSF I/UCRC Center for Embedded Systems.

\medskip
\noindent





\clearpage
\appendices
    
\section{Appendix: STPL Future Syntax}\label{lbl:apx:grammar}

The following definition introduces a future fragment of the introduced STPL syntax in Def. \ref{lbl:stql-brief-syntax}.
Here, we restrict the grammar by including rules that enforce a formula to be an \textit{Almost Arbitrarily Nesting Formula} as in Def. \ref{lbl-AAN-formula}.
\hlRev{Notice that in the following, the grammar rules force the expressions to be indexed to track the level of nesting in quantifier operators.}
  
\begin{defn}[STPL AAN Syntax for Discrete-Time Signal] \label{lbl:stql-syntax}
Let $V_x$ and $V_o$ be sets of time variables and object variables, respectively.
Let $x$ be a vector of time variables, i.e., $x = [x_0, \ldots, x_{n-1}]^T$, and 
$id$ be a vector of object variables, i.e., $id = [id_0, \ldots, id_{m-1}]^T$,
and $\Ic$ be any non-empty interval of  $\;\mathbb{R}_{\ge 0}$ over time.
The syntax for Spatio-Temporal Perception Logic (STPL) formulas is provided by the following grammar:
\begin{align*}
    \Phi_{i,j} &::= \exists{id_{i}}@x_i.\Phi^{f,q}_{i} \;|\; x_i.\Phi^{f}_{i} \;|\; \exists{id_{i}}.\Phi^{q}_{i,j} 
    \\
    & \;\;\;\;\;\; \top \;|\; 
    \neg \Phi_{i,j}\;|\; \Phi_{i,j} \vee \Phi_{i,j} \;|\; \bigcirc \Phi_{i,j} \;|\; \Phi_{i,j} \;\Un\; \Phi_{i,j} \;|\;
    \\
    &\;\;\;\;\;\; {\bigcirc_\Ic \Phi_{i,j} \;|\; \Phi_{i,j} \;\Un_\Ic\; \Phi_{i,j} \;|\; \tilde{\bigcirc}_\Ic \Phi_{i,j} \;|\; \Phi_{i,j} \;\tilde{\Un}_\Ic\; \Phi_{i,j}}
	\displaybreak[2] 
    \\
    \\
    \Phi^{f,q}_{i} &::= \Phi^{f}_{i} \;|\; \Phi^{q}_{i,i} \;|\;   
    \\
    & \;\;\;\;\;\; \top \;|\; 
    \neg \Phi^{f,q}_{i} \;|\; \Phi^{f,q}_{i} \vee \Phi^{f,q}_{i} \;|\; \bigcirc \Phi^{f,q}_{i} \;|\; \Phi^{f,q}_{i} \;\Un\; \Phi^{f,q}_{i} \;|\; 
    \\
    &\;\;\;\;\;\; {\bigcirc_\Ic \Phi^{f,q}_{i} \;|\; \Phi^{f,q}_{i} \;\Un_\Ic\; \Phi^{f,q}_{i} \;|\; \tilde{\bigcirc}_\Ic \Phi^{f,q}_{i} \;|\; \Phi^{f,q}_{i} \;\tilde{\Un}_\Ic\; \Phi^{f,q}_{i}}
	\displaybreak[2] 
    \\
    \\
    \Phi^{f}_{i} &::= \Ctime - x_i > t \;|\; \Cframe - x_i > n \;|\; \Phi_{i+1,i} \;|\; \\
    & \;\;\;\;\;\; \top \;|\; 
    \neg \Phi^{f}_{i} \;|\; \Phi^{f}_{i} \vee \Phi^{f}_{i} \;|\; \bigcirc \Phi^{f}_{i} \;|\; \Phi^{f}_{i} \;\Un\; \Phi^{f}_{i} \;|\;
    \\
    &\;\;\;\;\;\; {\bigcirc_\Ic \Phi^{f}_{i} \;|\; \Phi^{f}_{i} \;\Un_\Ic\; \Phi^{f}_{i} \;|\; \tilde{\bigcirc}_\Ic \Phi^{f}_{i} \;|\; \Phi^{f}_{i} \;\tilde{\Un}_\Ic\; \Phi^{f}_{i}}    
	\displaybreak[2] 
    \\
    \\
    \Phi^{q}_{i,j} &::= 
    C(\Lambda_{i,j}) = c \;|\; C(\Lambda_{i,j}) = C(\Lambda_{i,j}) \;|\; P(\Lambda_{i,j}) \ge r 
    \\
    & \;\;\;\;\;\; |\; P(\Lambda_{i,j}) \ge r \times P(\Lambda_{i,j}) \;|\;  {\Lambda_{i,j} = \Lambda_{i,j} \;|\; \Phi_{i+1,j}} \;|\;
    \\
    & \;\;\;\;\;\; \top \;|\; 
    \neg \Phi^{q}_{i,j} \;|\;
    \Phi^{q}_{i,j} \vee \Phi^{q}_{i,j} \;|\; \bigcirc \Phi^{q}_{i,j} \;|\; \Phi^{q}_{i,j} \;\Un\; \Phi^{q}_{i,j} \;|\; 
    \\
    &\;\;\;\;\;\; {\bigcirc_\Ic \Phi^{q}_{i,j} \;|\; \Phi^{q}_{i,j} \;\Un_\Ic\; \Phi^{q}_{i,j} \;|\; \tilde{\bigcirc}_\Ic \Phi^{q}_{i,j} \;|\; \Phi^{q}_{i,j} \;\tilde{\Un}_\Ic\; \Phi^{q}_{i,j}} \;|\; 
    \\
    & \;\;\;\;\;\; {\Sexists\Omega_{i,j}} \;|\; {\Theta_{i,j}} \;|\; {\Pi_{i,j}}
	\displaybreak[2] 
    \\
    \\
    {\Lambda_{i,j}}  &::= id_j \;|\;  id_{j+1} \;|\dots \;|\;  id_i
	\displaybreak[2] 
    \\
    \\
    {\Omega_{i,j}} &::= \BB(\Lambda_{i,j}) \;|\; \Sempty \;|\; \Suniverse \;|\; 
    \overline{\Omega}_{i,j} \;|\; \Omega_{i,j} \sqcap \Omega_{i,j} \;|\; \Itl{\Omega_{i,j}} \;|\;
    \\
    & \;\;\;\;\;\; \Omega_{i,j} \;\Un_s\; \Omega_{i,j} \;|\; 
    \Diamond_s \Omega_{i,j} \;|\; \Box_s \Omega_{i,j} \;|\; \bigcirc_s \Omega_{i,j} \;|\;
    \\
    & \;\;\;\;\;\; {\Omega_{i,j} \;\Un^{s}_{\Ic}\; \Omega_{i,j} \;|\; 
    \Diamond^{s}_{\Ic} \Omega_{i,j} \;|\; \Box^{s}_{\Ic} \Omega_{i,j} \;|\; \bigcirc^{s}_{\Ic} \Omega_{i,j}} \;|\;
    \\
    & \;\;\;\;\;\; {\Omega_{i,j} \;\tilde{\Un}^{s}_{\Ic}\; \Omega_{i,j} \;|\; 
    \tilde{\Diamond}^{s}_{\Ic} \Omega_{i,j} \;|\; \tilde{\Box}^{s}_{\Ic} \Omega_{i,j} \;|\; \tilde{\bigcirc}^{s}_{\Ic} \Omega_{i,j}}
	\displaybreak[2] 
    \\
    \\
    {\Pi_{i,j}} &::= Area(\Omega_{i,j}) \ge r \;|\; Area(\Omega_{i,j}) \ge r \times Area(\Omega_{i,j})
	\displaybreak[2] 
    \\
    \\
    {\Theta_{i,j}} &::=  Dist(\Lambda_{i,j},\textnormal{\scriptsize{CRT}},\Lambda_{i,j},\textnormal{\scriptsize{CRT}}) \ge r \;|\; 
    \\
    & \;\;\;\;\;\; Lat(\Lambda_{i,j},\textnormal{\scriptsize{CRT}}) \ge r \;|\; Lon(\Lambda_{i,j},\textnormal{\scriptsize{CRT}}) \ge r \;|\; 
    \\
    & \;\;\;\;\;\; Lat(\Lambda_{i,j},\textnormal{\scriptsize{CRT}}) \ge r \times Lat(\Lambda_{i,j},\textnormal{\scriptsize{CRT}}) \;|\;     
    \\
    & \;\;\;\;\;\; 
    Lon(\Lambda_{i,j},\textnormal{\scriptsize{CRT}}) \ge r \times Lon(\Lambda_{i,j},\textnormal{\scriptsize{CRT}}) \;|\;
    \\
    & \;\;\;\;\;\; Lat(\Lambda_{i,j},\textnormal{\scriptsize{CRT}}) \ge r \times Lon(\Lambda_{i,j},\textnormal{\scriptsize{CRT}}) \;|\;
    \\
    & \;\;\;\;\;\; Area(\Lambda_{i,j}) \ge r | Area(\Lambda_{i,j}) \ge r \times Area(\Lambda_{i,j})
	\displaybreak[2] 
    \\
    \\
    {\textnormal{CRT}} &::= \textnormal{\scriptsize{LM}} \;|\; \textnormal{\scriptsize{RM}} \;|\; \textnormal{\scriptsize{TM}} \;|\; \textnormal{\scriptsize{BM}} \;|\; \textnormal{\scriptsize{CT}}
\end{align*}
where $i \ge 0$, and the grammar starts from $\Phi_{0,0}$.
\end{defn}
The time and frame constraints of STPL are represented in the form of $\Ctime - x > r$ and $\Cframe - x > n$, respectively. 
The freeze time quantifier $x.\phi$ assigns the current frame $i$ to time variable $x$ before processing the subformula $\phi$. 
The \textit{Existential} quantifier is denoted as $\exists$, and  
the \textit{Universal} quantifier is \hlRev{denoted as $\forall$. 
The following syntactic equivalences hold for the STPL formulas $\psi$ and $\phi$ using syntactic manipulation.}
$ \forall\{id\}@x.\phi\equiv \neg(\exists\{ id\}@x.\neg\phi)$, 
$\psi\wedge\phi\equiv\neg(\neg\psi\vee\neg\phi)$, $\bot\equiv\neg\top$ (False), $\psi\rightarrow\phi\equiv\neg\psi\vee\phi$ ($\psi$ Implies $\phi$),
$\phi \;\Rc\; \psi \equiv \neg (\neg \phi \;\Un\; \neg \psi)$ ($\phi$ releases $\psi$),
$\phi \;\overline{\Rc}\; \psi \equiv \phi \;\Rc\; (\phi \vee \psi)$ ($\phi$ non-strictly releases $\psi$),
$\Diamond\psi \equiv \top \;\Un\; \psi$ (Eventually $\psi$),
$\Box\psi \equiv \neg \Diamond \neg\psi$ (Always $\psi$).
\revision{All the other operators with $\tilde{\cdot}$ on them are  with frame intervals, 
that is in $\tilde{\Box}^s_{\Ic}$, $\tilde{\Diamond}^s_{\Ic}$, $\tilde{\Un}^s_{\Ic}$, 
$\tilde{\Next}^s_{\Ic}$, $\tilde{\Un}_{\Ic}$, and $\tilde{\Next}_{\Ic}$ the interval $\Ic$ is over frame interval.}

\hlRev{For parsing a formula using the above grammar, there are two production rules $\Phi^{f}_{i}$ and $\Phi^{q}_{i,j}$ in which we can use the initial production rule after increasing the index $i$ (i.e., $\Phi_{i+1,j}$).
The index $i$ is to force scope for the use of freeze time variables.
For example, if in the scope of a variant-quantifier operator we use $x_0$, then the index will increases to $1$ to avoid use of $x_0$ in the scope of the next variant-quantifier operator. 
The index $j$ is used as a pointer to each quantifier operator to track the scope of object variables.
For example, in the formula $\varphi \equiv \exists id_0. \Box \big(\exists id_1. \exists id_2.(\phi_1) \vee \phi_2\big)$, we have $i=2$ and $j=0$ while parsing the subformula $\phi_1$, whereas, in $\phi_2$, we have $i=0$ and $j=0$.
Thus any function in $\phi_1$ with object variables in it will use the production rule $\Lambda_{2,0}$.
Thus, the allowed object variables in $\phi_1$ are $id_0, id_1$ and $id_2$.
However, while parsing the subformula $\phi_2$, we use $\Lambda_{0,0}$ in which the only allowed object variable is $id_0$.
}

\section{Appendix: Complexity Analysis of STE formulas}    \label{appendix:complexity}

\begin{table}[]
    \centering
    \begin{tabular}{|m{4mm}|m{4mm}|m{8mm}|m{9mm}|m{10mm}|m{13mm}|m{4mm}|}
         \hline
         $\boldsymbol{l}$ & $\boldsymbol{t_n}$  & $\boldsymbol{t_{n-1}}$  & $\boldsymbol{t_{n-2}}$ & $\boldsymbol{t_{n-3}}$ & $\boldsymbol{t_{n-4}}$ &  $\boldsymbol{\dots}$\\
         \hline
         \textbf{0} & 1 & 1 & 1 & 1 & 1 & $\dots$\\
         \hline
         \textbf{1} & 1 & 1(1+1) \textbf{= 2} & 1(1+ 1(1+1)) \textbf{= 3} & 1(1+ 1(1+ 1(1+1))) \textbf{= 4} & 1(1+ 1(1+ 1(1+ 1(1+1)))) \textbf{= 5}  & $\dots$\\
         \hline
         \textbf{2} & 1 & 2(1+1) \textbf{= 4}, $2 \le 4 \le 3!$ & 3(1+ 2(1+1)) \textbf{= 15}, $2^2 \le 15 \le 4!$ & 4(1+ 3(1+ 2(1+1))) \textbf{= 64}, $ 2^3 \le 64 \le 5!$ & 5(1+ 4(1+ 3(1+ 2(1+1)))) \textbf{= 325}, $2^4 \le 325 \le 6!$  & $\dots$\\
         \hline
         \textbf{3} & 1 & 2(1+1) (1+1) \textbf{= 8} & 3(1+ 2(1+1)) (1+ 2(1+1) (1+1)) \textbf{= 135} & 4(1+ 3(1+ 2(1+1))) (1+ \newline 3(1+ 2(1+1)) (1+ 2(1+1) (1+1))) \textbf{= 8,704} & 5(1+ 4(1+ 3(1+ 2(1+1)))) (1+ \newline 4(1+ 3(1+ 2(1+1))) (1+ \newline 3(1+ 2(1+1)) (1+ 2(1+1) (1+1)))) \textbf{= 2,829,125}  & $\dots$\\
         \hline
         $\dots$ & $\dots$ & $\dots$ & $\dots$ & $\dots$ & $\dots$ & $\dots$ \\
         \hline
    \end{tabular}
    \caption{DP-based complexity analysis for spatio-temporal until operator with different levels of nesting. At $l = 0$ we have $\tau = \BB(Id)$; At $l=1$: we have $\tau \;\Un^{s}_{\Ic}\; \tau$, and $\tau \;\Un^{s}_{\Ic}\; \tau$; At $l=2$: we have $(\tau \;\Un^{s}_{\Ic}\; \tau)$ $\;\Un^{s}_{\Ic}\; (\tau \;\Un^{s}_{\Ic}\; \tau)$; Finally, at $l=3$: we have $\big((\tau \;\Un^{s}_{\Ic}\; \tau)$ $\;\Un^{s}_{\Ic}\; (\tau \;\Un^{s}_{\Ic}\; \tau)\big)$ $\;\Un^{s}_{\Ic}\; \big((\tau \;\Un^{s}_{\Ic}\; \tau)$ $\;\Un^{s}_{\Ic}\; (\tau \;\Un^{s}_{\Ic}\; \tau)\big)$.
    }
    \label{tbl:complexity}
\end{table}

Here the assumption is that we use the linked-list data structure to represent a spatial term $\Stau$ as a union of a finite number of unique subsets.
\revision{We can compute $\Sval(\tau_1 \;\Un^{s}_{\Ic}\; \tau_2, \Data, t, \tau, \epsilon, \zeta)$ recursively as 
$\Sval(\tau_2,\Data, t, \tau, \epsilon, \zeta)$ $\cup$ $\big(\Sval(\tau_1, \Data, t, \tau, \epsilon, \zeta)$ $\cap$ $\Sval((\tau_1 \;\Un^{s}_{\Ic}\; \tau_2), \Data, t+1, \tau, \epsilon, \zeta) \big)$.}
For each until formula in each row of Table \ref{tbl:complexity} (starting from the row $l=1$), for each time step $t$, we use the recursive evaluation function to calculate the maximum number of bounding boxes as a result of computing the formula. 
The maximum number of bounding boxes that can be produced by $\tau_1 \cup \tau_2$ is equal to the total number of boxes in the two spatial terms (i.e., $|\tau_1| + |\tau_2|$).
Additionally, the maximum number of bounding boxes that can be produced by $\tau_1 \cap \tau_2$ is equal to the product of number of boxes in the two spatial terms (i.e., $|\tau_1| \times |\tau_2|$).
Consequently, the maximum number of bounding boxes that can be produced at each time step $t$ for the above until formula is \revision{$|\Sval(\tau_2, \Data, t, \tau, \epsilon, \zeta)| + |\Sval(\tau_1, \Data, t, \tau, \epsilon, \zeta)| \times | \Sval(\tau_1 \;\Un^{s}_{\Ic}\; \tau_2, \Data, t+1, \tau, \epsilon, \zeta)|$.}

\subsection{Formulas with Exponential Time/Space Complexities}
As it is stated in the first and second rows in Table \ref{tbl:complexity}, the number of needed operations grow as in arithmetic sequences.
Thus, the time complexity which is the summation of numbers in the rows, are linear and polynomial functions of the number of the time steps for the first and second rows, respectively. 
Moreover, the space complexity for the first and the second rows are constant and linear functions of number of the time steps, respectively.

In the following, we calculate an upper bound and a lower bound for the maximum number of bounding boxes that can be produced for the third row (level 2) of the until operator.
We use the function $f_{2}(t)$ to denote the maximum number of bounding boxes that are produced at the time step $t$ for the until formula $(\tau \;\Un^{s}_{\Ic}\; \tau)$ $\;\Un^{s}_{\Ic}\; (\tau \;\Un^{s}_{\Ic}\; \tau)$.
\begin{multline}\label{lbl:eq:f_2}
    \sum_{t=1}^{n} f_{2}(t) = 
    \\
    1 + 
    2(1+1) + 
    3(1+ 2(1+1)) + 
    \\
    4(1+ 3(1+ 2(1+1))) +
    5(1+  4(1+3(1+ 2(1+1)))) + \dots
    \\
    + n(1 + (n-1)(1 + (n-2)(\dots 1+2(1+1)))\dots)
\end{multline}
We repetitively use the inequality $(a+1)b > a(1+b)$ for $b > a$ to derive the following inequality from the above equation
\begin{gather}
    \sum_{t=1}^{n} f_{2}(t) <
    1 + 3! + 4! + 5! + 6! + \dots + (n+1)!
\end{gather}
where $n \ge 2$.
Therefore, we have
\begin{gather}\label{lbl:eq:f2simplified_ub}
    \sum_{t=1}^{n} f_{2}(t) < n \times (n+1)! < (n+2)!
\end{gather}

Next, we calculate a lower bound for the maximum number of bounding boxes that can be produced for the level 2 of the until operator.
We repetitively use the inequality $2^{(b+1)} < a(1+2^b)$ for $a \ge 2$ to derive the following inequality from Eq. (\ref{lbl:eq:f_2})
\begin{gather}
    \sum_{t=1}^{n} f_{2}(t) >
    1 + 2^1 + 2^2 + 2^3 + 2^4 + \dots + 2^{(n-1)}
\end{gather}
where $n \ge 2$.
Therefore, we have
\begin{gather}\label{lbl:eq:f2simplified_lb}
    \sum_{t=1}^{n} f_{2}(t) > 2^n 
\end{gather}

This time inequality suggests that the time/space complexity for any formulas with more than one level of nesting can be exponential.

\subsection{Best Complexity for the Worst Formulas}
We can repeat the above method to calculate a lower bound for each row of the table by using the inequality $a^{r+1} < a^r(1+b)$ for $b>a$ in each summation of the elements of rows to derive the below inequality 
\begin{multline*}
    \sum_{t=1}^{n} f_{0}(t) +\sum_{t=1}^{n} f_{1}(t) +\dots+ \sum_{t=1}^{n} f_{l}(t) >  
    \\
    n + \frac{n(n+1)}{2} + \frac{2^{n}-1}{2-1} + \frac{3^{n}-1}{3-1} + \dots 
    + \frac{l^{n}-1}{l-1} >
    \\
    n + \frac{n(n+1)}{2} + 
    2^{(n-1)} + 3^{(n-1)} + \dots + l^{(n-1)} 
    \\
    - (1+\frac{1}{2} +\frac{1}{3} +\dots +\frac{1}{l-1} ) 
    \\
    >
    2^{(n-1)} + 3^{(n-1)} + \dots + l^{(n-1)} 
\end{multline*}
where $n \ge 2$.
Therefore, we have
\begin{gather}\label{lbl:eq:f2simplified_lc}
    \sum_{t=1}^{n} f_{0}(t) +\sum_{t=1}^{n} f_{1}(t) +\dots+ \sum_{t=1}^{n} f_{l}(t) >  l^{(n-1)}
\end{gather}

This concludes the complexity of the algorithm to be $\Omega(|\varphi_s|^{(|\hat{\rho}|-1)})$.

\end{document}